\pdfoutput=1
\def\year{2021}\relax
\documentclass[letterpaper]{article} 
\usepackage{aaai21}  
\usepackage{times}  
\usepackage{helvet} 
\usepackage{courier}  
\usepackage[hyphens]{url}  
\usepackage{graphicx} 
\usepackage{natbib}  
\usepackage{caption} 
\usepackage{amsmath}
\usepackage{amsthm}
\usepackage{algorithm}
\usepackage{algpseudocode}
\usepackage{algorithmicx}
\usepackage{mathtools,amssymb}
\usepackage{bm}
\usepackage{subfig}
\usepackage{verbatimbox}
\usepackage{makecell}
\usepackage{booktabs}

\newtheorem{thm}{\bf Theorem}[section]
\newtheorem{cor}{\bf Corollary}[section]

\newtheorem{Assumption}{Assumption}[section]

\title{Stabilizing Q Learning Via Soft Mellowmax Operator}

\author{Yaozhong Gan, Zhe Zhang, Xiaoyang Tan}

\affiliations{

    College of Computer Science and Technology,Nanjing University of Aeronautics and Astronautics
    
    MIIT Key Laboratory of Pattern Analysis and Machine Intelligence
    
    Collaborative Innovation Center of Novel Software Technology and Industrialization
    
    \{yzgancn, zhangzhe, x.tan\}@nuaa.edu.cn
}

\begin{document}

\maketitle

\begin{abstract}
  Learning complicated value functions in high dimensional state space by function approximation is a challenging task, partially due to that the max-operator used in temporal difference updates can theoretically cause instability for most linear or non-linear approximation schemes. Mellowmax is a recently proposed differentiable and non-expansion softmax operator that allows a convergent behavior in learning and planning. Unfortunately, the performance bound for the fixed point it converges to remains unclear, and in practice, its parameter is sensitive to various domains and has to be tuned case by case. Finally, the Mellowmax operator may suffer from oversmoothing as it ignores the probability being taken for each action when aggregating them. In this paper, we address all the above issues with an enhanced Mellowmax operator, named SM2 (Soft Mellowmax). Particularly, the proposed operator is reliable, easy to implement, and has provable performance guarantee,  while preserving all the advantages of Mellowmax. Furthermore, we show that our SM2 operator can be applied to the challenging multi-agent reinforcement learning scenarios, leading to stable value function approximation and state of the art performance.

\end{abstract}

\section{Introduction}\label{Introduction}
Reinforcement Learning (RL) aims to learn an optimal mapping from states to actions, through experience obtained by interacting with the underlying environment, in terms of some accumulated performance criteria. One major challenge of RL is that in many real-world scenarios (e.g., Atari games \cite{Mni}, continuous control task \cite{baselines}, and robot control \cite{Lil}), one has to explore the optimal control in high dimensional state and/or action space, leading to the so-called 'Curse of Dimensionality' problem. Among others, one popular and successful approach to address this tractability issue is to rely on suitable approximate representations. For example, in value function-based RL, it is required that a target value function can be sufficiently precisely captured by some general-purpose function approximators, such as a deep neural network \cite{Mni}. However, training such a value function approximator is in general difficult and can diverge even in some simple cases \cite{sutton1996generalization}. It is found that in practice the deep RL is sensitive to both the initial condition and the underlying training dynamics \cite{henderson2017deep}.

Theoretically the max-operator in Bellman equation and temporal difference (TD) can  keep most linear or non-linear approximation schemes from converging \cite{tsitsiklis1997analysis}. One possible reason behind this is that the non-differentiable property of max-operator has a discretization effect during the TD target estimation - that is, small changes in value function may cause large policy variation, making a suboptimal policy being incorrectly reinforced. This issue is termed \emph{overestimation phenomenon} in literatures \cite{Thr}, which turns out to be a major source of instability of the original deep Q-network (DQN) \cite{Mni}. To address this issue, double DQN (DDQN) adopts a heuristic approach by decoupling the action selection and the TD target estimation \cite{Van}.

Recently there has been increasing interest in stablizing the training procedure of non-linear function approximator by replacing the max function with softmax operators with better theoretical properties. In \cite{Ans}, a method named Averaged-DQN is proposed by basing their max-operator on a running average function over previously learned Q-values estimates. They show that this is beneficial to alleviate the overestimation issue by controlling its upper bound. To encourage exploration and improve robustness, several authors investigate using the technique of entropy regularization to smooth policy \cite{Fox, Mei, BoD, Kyun}, this leads to softmax operators. \cite{Song} revisit the softmax Bellman operator and show that it can consistently outperform its max counterpart on several Atari games.

Along this line of research, an alternative mellowmax operator is proposed in \cite{Asa}. This is a differentiable and non-expansion softmax operator that allows a convergent behavior in learning and planning. Unfortunately, the performance bound for the fixed point it converges to remains unclear, and in practice, its parameter is sensitive to various domains and has to be tuned case by case \cite{Kim}. Finally, the Mellowmax operator may suffer from oversmoothing as it ignores the probability being taken for each action when aggregating them.

In this paper we address all the above issues of Mellowmax with an enhanced operator, named SM2 (Soft Mellowmax). Particularly, the SM2 operator can be thought of as a weight quasi-arithmetic mean \cite{Bel}, taking into account all action values in each state and the corresponding policy. We show that the proposed operator not only has provable performance guarantee, but is reliable, easy to implement, and capable of stablizing the TD training by alleviating the overestimation error.

As another contribution of this work, we extend our study of soft operator into the field of multi-agent reinforcement learning (MARL) \cite{Kra, Oli},  showing that \emph{overestimation phenomenon} could become even worse with the increasing number of agents. This is an important issue but unfortunately often ignored by the community, and our result highlights the need to pay more attention on this. To this end, we prove that our SM2 operator is helpful in tackling the overestimation problem in MARL, and provide the lower and upper bound of the overestimation reduction that could be achieved. Extensive experimental results on the StarCraft Multi-agent Challenge (SMAC) \cite{SMAC} benchmark show that our proposed method significantly improves the performance and sample efficiency compared to several state of the art MARL algorithms.

In what follows, we first provide the background in Section \ref{Ba}.
The proposed SM2 algorithm and its theoretical analysis are described in Section \ref{Me}.
In Section \ref{Om}, the overestimation problem is discussed.
The main experimental results are given in Section \ref{Ex}, and we conclude the paper in Section \ref{Co}.

\section{Background}\label{Ba}
Commonly, the reinforcement learning problem can be modeled as a Markov Decision Processes (MDP) which described by a tuple $\left\langle\mathcal{S},\mathcal{A},P, R, \gamma\right\rangle$.
$\mathcal{S}$ and $\mathcal{A}$ are the state space and action space respectively.
The function $P(s'|s,a): \mathcal{S}\times \mathcal{A}\times \mathcal{S}\longmapsto[0,1]$ outputs the transition probability from state $s$ to state $s'$ under action $a$.
The reward on each transition is given by the function $R(s,a) : \mathcal{S} \times \mathcal{A} \longmapsto \mathbb{R}$, whose maximum absolute value is $R_{max}$.
$\gamma \in [0,1)$ is the discount factor for long-horizon returns.
The goal is to learn a policy $\pi: \mathcal{S} \times \mathcal{A}\longmapsto [0,1]$ for interacting with the environment.

\subsection{The Bellman and Mellowmax operator}
Based on the MDP model, the action value function is defined as $Q^{\pi}(s_t, a_t)=\mathbb{E}_{s_{t+1}:\infty,a_{t+1}:\infty}[G_t|s_t,a_t]$, where $G_t=\sum_{i=0}^{\infty}\gamma^i R_{t+i}$ is the discount return.
The standard Bellman operator $\mathcal{T}$ is defined as,
\begin{equation}\label{b-1}
\mathcal{T}Q(s,a) = R(s,a)+ \gamma \sum_{s'}P(s'|s,a)\max_{a'}Q(s',a')
\end{equation}

It is well known \cite{Mar,Ric, Ric2} that the optimal Q function $Q^*(s,a)=\max_{\pi}Q^{\pi}(s,a)$ is the unique solution $Q^*$ to the standard Bellman equation and satisfies:
\begin{equation*}\label{b-2}
\begin{split}
Q^*(s,a)&=\mathcal{T}Q^*(s,a)\\
&=R(s,a)+\gamma \sum_{s'}P(s'|s,a)\max_{a'}Q^*(s',a')
\end{split}
\end{equation*}
The corresponding optimal policy is defined as $\pi^*=\arg\max_{a}Q^*(s,a)$.

In DQN \cite{Mni}, the action value function $Q_{\theta}(s,a)$ is implemented with a deep neural network parametrized by $\theta$, which takes the state $s$ as input and outputs the $ Q $ value of every action $a$.
In order to learn the parameters $\theta$, batches of transitions are sampled from the \textit{replay buffer}, which stores the transition tuples $\langle s,a,R,s' \rangle$.
By minimizing the squared TD error, the training object can be represented as:
\begin{equation}\label{b-3}
\min_{\theta}\frac{1}{2}\|Q_{\theta}(s,a)-y_{T}^{DQN}\|^2
\end{equation}
where $y_{T}^{DQN}=R(s,a)+\gamma \max_{a'}Q_{\theta^-}(s',a')$ denotes the target value, $\theta^-$ represents the parameters of a target network that are periodically copied from $\theta$.

To alleviate the overestimation problem and make the training process more stable, double-DQN \cite{Van} uses two separate neural networks for action selection and evaluation and replaces the target $y_{T}^{DQN}$ in Eq.(\ref{b-3}) with
\begin{equation*}\label{b-4}
y_{T}^{DDQN}=R(s,a)+\gamma Q_{\theta^-}(s', \arg\max_{a'}Q_{\theta}(s',a'))
\end{equation*}
That is, it selects the action based on the current network $\theta$ and evaluates the $ Q $ value using the target network $\theta^-$.

Mellowmax operator \cite{Asa, Kim} is an alternative way to reduce the overestimation bias, and is defined as:

\begin{equation}\label{b-5}
mm_{\omega}Q(s',\cdot)=\frac{1}{\omega}\log[\sum_{i=1}^{n}\frac{1}{n}\exp(\omega Q(s',a_{i}'))]
\end{equation}
where $\omega$ $>0$, and by adjusting it, the $mm_{\omega}Q(s,\cdot)$ value can change between the mean and maximum of all action values $Q(s,\cdot)$ under state $s$.
Analogous to the Bellman operator Eq.(\ref{b-1}), we can also define the Mellomax Bellman operator as:
\begin{equation}\label{b-55}
\mathcal{T}_m Q(s,a) = R(s,a)+ \gamma \sum_{s'}P(s'|s,a)mm_{\omega}Q(s',\cdot)
\end{equation}
Similarly, mellow-DQN algorithm modifies the target $y_{T}^{DQN}$ in Eq.(\ref{b-3}) with $y_{T}^{mm_{\omega}}=R(s,a)+\gamma mm_{\omega}Q_{\theta^-}(s',\cdot)$.

It can be shown that $mm_{\omega}$ is non-expansion under the infinity norm.
Therefore, $Q(s, a)$ is guaranteed to converge to a unique fixed point by recursively iterating $\mathcal{T}_m$ .
However, the relationship between the stationary $Q$ value and the optimal $ Q $ value is still unknown.
And in practice, the parameter $\omega$ is sensitive to different tasks and has to be tuned case by case \cite{Kim}.
According to Eq.(\ref{b-5}), Mellowmax can be thought of as a controlled smoothing operator that aggregates Q values over the whole action space to give an estimation of future return. 
Under some complicated environments with sparse rewards or large action space, the estimated action values calculated by Mellowmax may tend to be much smaller than the ones by max operator with an improper $ \omega $ value. We call this oversmoothing issue.

\section{Soft Mellowmax}\label{Me}
In this section, we first introduce a new operator called soft mellowmax (SM2) operator, which aims to avoid the shortcomings of mellow operator mentioned before, and then give its theoretic properties.

\subsection{The Soft Mellowmax Operator}
The soft mellowmax operator is defined as
\begin{equation}\label{m-1}
sm_{\omega}Q(s,\cdot)=\frac{1}{\omega}\log[\sum_{i=1}^{n}soft_{\alpha}(Q(s',a_i'))\exp(\omega Q(s',a_i'))]
\end{equation}
where $soft_{\alpha}(Q(s',a_i'))=\frac{\exp(\alpha Q(s',a_i'))}{\sum_{j=1}^{n}\exp(\alpha Q(s',a_j'))}$, $\omega>0$ and 
$ \alpha \in \mathbb{R} $,
which can be viewed as a particular instantiation of the weighted quasi-arithmetic mean \cite{Bel}.
The $soft_{\alpha}Q$ term can be thought of as a policy probability, and can be justified from the perspective of  entropy regularization framework with KL divergence \cite{Fox, Mei}.

Analogous to the Bellman operator Eq.(\ref{b-1}), we define the Soft Mellowmax Bellman operator as:
\begin{equation}\label{sm-1}
\mathcal{T}_{sm} Q(s,a) = R(s,a)+ \gamma \sum_{s'}P(s'|s,a)sm_{\omega}Q(s',\cdot)
\end{equation}

Note that if $\alpha=0$, the $\mathcal{T}_{sm}$ operator is reduced to the Mellomax Bellman operator $\mathcal{T}_{m}$. As will shown below, $\mathcal{T}_{sm}$ extends the capability of the $\mathcal{T}_{m}$ in several aspects, in terms of provable performance bound, overestimation bias reduction and sensitiveness to parameters settings.



\subsection{Contraction}
The following theorem shows that that $\mathcal{T}_{sm}$ is a contraction. That is,
it is guaranteed to converge to a unique fixed point.

\begin{thm}\label{m-th1}
	(\textbf{Contraction})
	Let $Q_1$, $Q_2$ are two different Q-value functions. If $\omega>0$ and 
	$ -\frac{\omega}{1-e^{-c\omega}} \leq \alpha \leq \frac{\omega}{e^{c\omega}-1}$
	where $c=\frac{2R_{max}}{1-\gamma}$. Then
	\begin{equation*}\label{m-3}
	\|\mathcal{T}_{sm}Q_1-\mathcal{T}_{sm}Q_2\|\leq \gamma\|Q_1-Q_2\|
	\end{equation*}
\end{thm}
The proof of the theorem is given in Appendix.

Here we give an example, showing that if there is no restriction on $\alpha$, the $\mathcal{T}_{sm}$ operator will not be a contraction.
\textbf{Example 3.1}.\\
Let $\mathcal{S}=\{0\}$, $\mathcal{A}=\{1,2\}$, $\alpha=1$, $\omega=1$, $Q_1(0,1)=50$, $Q_1(0,2)=1$, $Q_2(0,1)=5$, and $Q_2(0,2)=1$, then
\begin{equation*}
|sm_{\omega}Q_1-sm_{\omega}Q_2|\approx 45.018 \geq \max_{a}|Q_1-Q_2|=45
\end{equation*}

Therefore, we have shown that $\mathcal{T}_{sm}$ is a contraction under certain conditions, but how about the quality of the converged fixed point then? we answer this next.



\subsection{Performance Bound}

In this section, we derive the bound of the performance gap between the optimal action-value function $Q^*$ and $Q_k$ during $k$-th Q-iterations with $\mathcal{T}_{sm}$ Eq.(\ref{sm-1}). 
And we only discuss the case of $ \alpha\geq 0 $ in the following sections.

\begin{thm}\label{m-th}
	Let the optimal Q-value function $Q^*$ be a fixed point during Q-iteration with $\mathcal{T}$, i.e. $Q^*(s,a)=\mathcal{T}Q^*(s,a)$, and $Q^k(s,a)\triangleq \mathcal{T}_{sm}^kQ^0(s,a)$ for arbitrary initial function $Q^0(s,a)$ during $k$-th Q-iteration.
	$\forall (s,a)$
	\\
	\\
	(\uppercase\expandafter{\romannumeral1}) If $\alpha\geq\omega$, then
	\begin{align*}
	\limsup\limits_{k\rightarrow\infty}\|Q^*(s,a)-Q^k(s,a)\|\leq\frac{\gamma}{\omega(1-\gamma)}\log(\frac{1+n}{2}) \quad\quad
	\end{align*}
	(\uppercase\expandafter{\romannumeral2}) If $\alpha<\omega$, then
	\begin{align*}
	\!\limsup\limits_{k\rightarrow\infty}\!\|Q^*(s,a)\!-\!Q^k(s,a)\|\!
	\!\leq\! \frac{\gamma}{\omega(1\!-\!\gamma)}\!\log(n\!-\!\frac{\alpha(n\!-\!1)}{\alpha\!+\!\omega})
	\end{align*}
\end{thm}

\begin{cor}
	For the Mellowmax Bellman operator $ \mathcal{T}_{m}$ Eq.(\ref{b-55}), defined $ Q^{k}(s,a)\triangleq \mathcal{T}_{m}^k Q^0(s,a)$, we have
	\begin{equation*}
	\lim\limits_{k \rightarrow \infty}\|Q^{\star}(s,a)-Q^k(s,a)\|\leq \frac{\gamma}{\omega(1-\gamma)}\log(n) \quad\quad\quad
	\end{equation*}
\end{cor}

The proof of the theorem is given in Appendix.

This result shows that the optimal action-value function $Q^*$ and $Q^k$ can be limited to a reasonable range in either cases, and the bound is independent of the dimension of state space, but depends on the size $n$ of action space. 

Furthermore, it shows that $\mathcal{T}_{sm}$ converges to $\mathcal{T}$ with a linear rate with respect to $\omega$ when $\alpha\geq\omega$. If $\alpha<\omega$, it converges to $\mathcal{T}$ with respect to $\omega$ and $\alpha$. 
Compared with Mellowmax Bellman operator, our method can converge to the optimal $ Q^* $ more quickly.
Note that the softmax operator \cite{Song} has a better convergence rate, that is, exponential with respect to the parameters instead of linear, and its scalar term is ($n-1$).
In contrast, the scalar term of our method and some other methods within entropy-regularized MDP framework \cite{BoD, Kyun} are $\log n$, as shown in theorem \ref{m-th}.




\subsection{Alleviating Overestimation}
The overestimation issue of Q-learning due to the max operator may lead to the suboptimal policy and cause the failure of Q-learning \cite{Thr}, and it is may be one of the main reasons of the poor performance of DQN in 
many real-world scenarios
 \cite{Van}. Recently this issue has received increasing attentions \cite{Van,Ans, Song}. In this section, following the same assumptions \cite{Van}, we show that our SM2 operator is able to reduce the overestimation in Q-learning.

\begin{thm}\label{m-over1}
	Assume all the true optimal action value are equal at $Q^*(s, a)=V^*(s)$ and the estimation errors $Z^{s,a}=Q(s, a)-Q^*(s, a)$ are independently distributed uniformly random in $[-\epsilon, \epsilon]$, defined the overestimation error
	$\varTheta\triangleq\mathbb{E}[\max_{a}Q(s,a)-Q^*(s,a)]$,
	and
	$\varTheta_{sm}\triangleq\mathbb{E}[sm_{\omega}Q(s,a)-Q^*(s,a)]$,
	\\
	\\
	(\uppercase\expandafter{\romannumeral1}) If $\alpha\geq\omega$, then
	\begin{equation*}
	\varTheta-\varTheta_{sm}\in (0,\frac{1}{\omega} \log(\frac{1+n}{2})]  \quad \quad \quad \
	\end{equation*}
	(\uppercase\expandafter{\romannumeral2}) If $\alpha<\omega$, then
	\begin{equation*}
	\varTheta-\varTheta_{sm}\in (0,\frac{1}{\omega}\log(n-\frac{\alpha(n-1)}{\alpha+\omega})]
	\end{equation*}
	(\uppercase\expandafter{\romannumeral3}) The overestimation error for $\mathcal{T}_{sm}$ is increasing monotonically w.r.t. $\alpha\in[0,\infty)$ and $\omega\in(0,\infty)$.
\end{thm}

\begin{cor}
	Under the above assumption, defined $ \varTheta_{m}\triangleq\mathbb{E}[mm_{\omega}Q(s,a)-Q^*(s,a)] $, we have
	\begin{equation*}
	\varTheta-\varTheta_{m}\in (0, \frac{1}{\omega}\log(n)]
	\end{equation*}
\end{cor}

The proof of the theorem is given in Appendix.

This theorem implies that our SM2 algorithm can reduce the overestimation of max-operator. Particularly, when $\omega$ is fixed, we can adjust the scalar of $\alpha$ to control the degree of overestimation in part (\uppercase\expandafter{\romannumeral2}).
Similarly, we also analyze the overestimation of Mellowmax operator.
If the $ \omega $ of SM2 and Mellowmax are same, the overestimation bias of Mellowmax is smaller, which may also cause the oversmoothing issue. 
SM2 considers the weights of difference $ Q $ values, which take the importance of each $ Q $ value into account. 
Compared with \cite{Song}, the scalar term of our method is at most $\log(n)$, but independent of $ \gamma $ and $ R_{max} $.

\section{Soft Mellowmax for Mulit-Agent Reinforcement Learning}\label{Om}
In this section, we investigate how to apply the proposed SM2 operator for multi-agent reinforcement learning (MARL). Particularly, we consider a fully cooperative multi-agent scenario under the paradigm of centralized training with decentralized execution (CTDE). The goal is to find a separate policy for each agent $i$, such that when following the joint policy $\bm{\pi}$ in an predefined environment, the expected centralized value $ Q_{tot} $ is maximized, under certain coordinative constraints between centralized value function $ Q_{tot} $ and decentralized value functions $ Q_i $. Formally, we consider the following optimization problem.

\begin{equation*}\label{1.1}
\begin{split}
&\mathop{{\rm maximize}}_{\bm{\pi}} \mathop{\mathbb{E}}_{s_0\sim \rho_0(s_0), \bm{a}\sim \bm{\pi}}[Q_{tot}(s_0,\bm{a})]\\
&\ \mbox{subject to}\ Q_{tot}(s, \bm{a})=f(\bm{Q})(s, \bm{a})
\end{split}
\end{equation*}
where $ \rho_0: \mathcal{S}\longmapsto [0,1]$ is the distribution of the initial state $ s_0 $. The value functions $Q_i$ of each agent $i$ are combined to give the centralized value function $ Q_{tot} $, through a state-dependent continuous function $ f $, where $ f(\bm{Q})(s, \bm{a})= f(Q_1(s,a_1), \cdots, Q_N(s,a_N)) $, with $\textbf{Q}=(Q_1, Q_2, \cdots, Q_N)$ being the joint value vector and $\textbf{a}=(a_1, a_2, \cdots, a_N)$ being the joint action vector of the $N$ agents. In what follows, for simplicity, we ignore $(s, \textbf{a})$.


\begin{Assumption}\label{b-ssu}
	For any $\textbf{Q}=[Q_{i}]_{i=1}^{N}$, and exits $l\geq0$,  $L>0$, the following holds\\
	$$  l\leq\frac{\partial Q_{tot}}{\partial Q_i}\leq L,\ i=1,2, \cdots,N.  $$
\end{Assumption}

Assumption \ref{b-ssu} suggests that $Q_{tot}$ is a monotonic function and a Lipschitz continuous function about $\textbf{Q}$ (every function that has bounded first derivatives is Lipschitz continuous). Among others, both VDN \cite{Sun} and QMIX \cite{Ras} satisfy this assumption in practice.
We take this assumption as a prior to capture the coordination information required to solve cooperative tasks. That is, each agent has a chance to influence the team reward positively instead of canceling out with each other in fully cooperative scenarios.

Following the same assumptions of theorem \ref{m-over1}, we first quantify the overestimation problem in the situation of MARL, as follows,

\begin{thm}\label{b-oth}
	Under the assumptions of Theorem \ref{m-over1}
	Defined the overestimation error $\varTheta^{1}\triangleq\mathbb{E}[Q_{tot}(\max_{\textbf{a}}\textbf{Q})(s,\textbf{a})$
	$-Q_{tot}(\textbf{Q})^*(s,\textbf{a})]$, then
	\begin{equation*}\label{b-10}
	\begin{split}
	\varTheta^{1}
	\in [\epsilon l N \frac{n-1}{n+1}, \epsilon L N \frac{n-1}{n+1}]
	\end{split}
	\end{equation*}
\end{thm}

The proof of the theorem is given in Appendix.

This result implies that the amount of overestimation is proportional to the number of agents, which means that in MARL, the overestimation problem is more serious than that in the single agent case as more than one agents are involved in learning. As shown in the experimental section (c.f., Figure \ref{fig:overestimation}), the QMIX does suffer from this issue.

Next, we show that our SM2 operator is beneficial to address this, and gives both lower and upper bounds of the range of the overestimation reduction.

\begin{thm}\label{m-over}
	Under the assumptions of Theorem \ref{m-over1}.
	Defined the overestimation error $\varTheta_{sm}^{1}\triangleq\mathbb{E}[Q_{tot}(sm_{\omega}\textbf{Q})(s,\textbf{a})$
	$-Q_{tot}(\textbf{Q})^*(s,\textbf{a})]$,
	\\
	\\
	(\uppercase\expandafter{\romannumeral1}) If $\alpha\geq\omega$, then
	\begin{equation*}
	\varTheta^{1}-\varTheta_{sm}^{1}\in (0,\frac{1}{\omega}L N \log(\frac{1+n}{2})]  \quad \quad \quad\
	\end{equation*}
	(\uppercase\expandafter{\romannumeral2}) If $\alpha<\omega$, then
	\begin{equation*}
	\varTheta^{1}-\varTheta_{sm}^{1}\in (0,\frac{1}{\omega}LN\log(n-\frac{\alpha(n-1)}{\alpha+\omega})]
	\end{equation*}
\end{thm}
The proof of the theorem is given in Appendix.

This theorem implies that our SM2 algorithm can reduce overestimation in MARL.
The scale of the overestimation error is related to the selection of hyperparameters and the number of agents. 

\section{Experiment}\label{Ex}
In this section, we present our experimental results  to verify the effectiveness of the proposed SM2 method.

\begin{figure}[th!]
	\centering
	\subfloat[Catcher-PLE]{\includegraphics[width=0.23\textwidth]{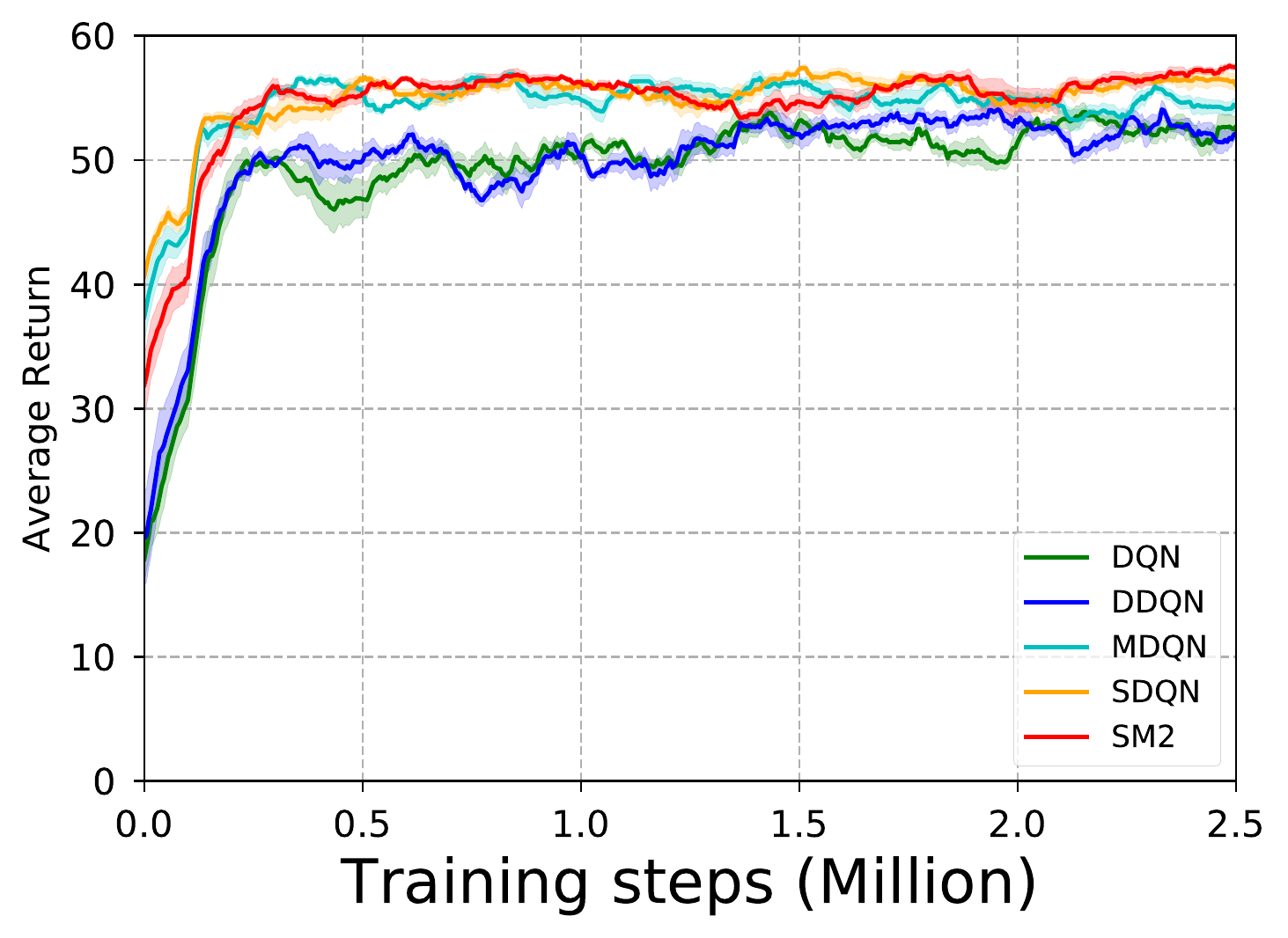}}\hfill
	\subfloat[Pixelcopter-PLE]{\includegraphics[width=0.23\textwidth]{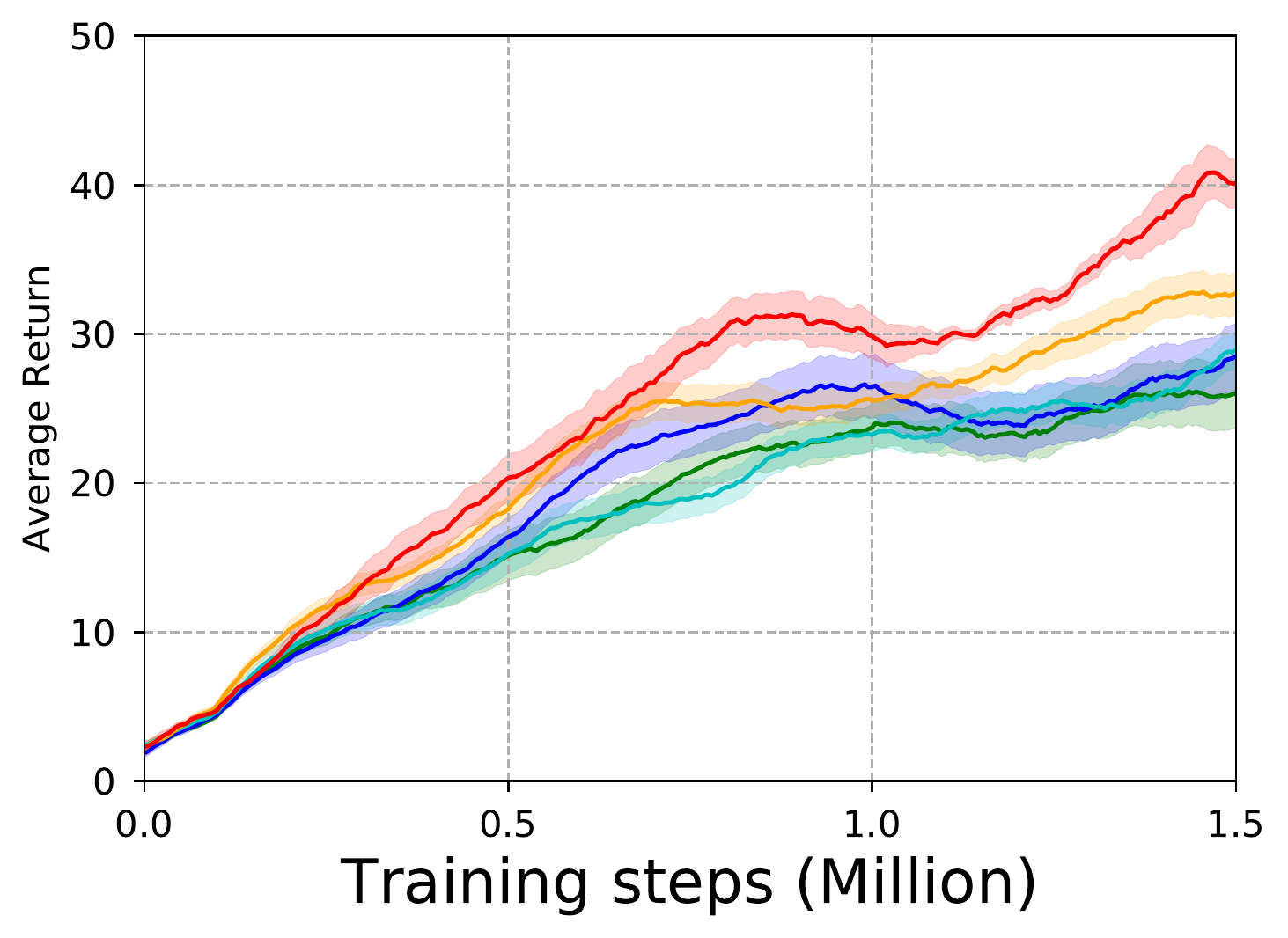}}\hfill\\
	\subfloat[Pong-PLE]{\includegraphics[width=0.23\textwidth]{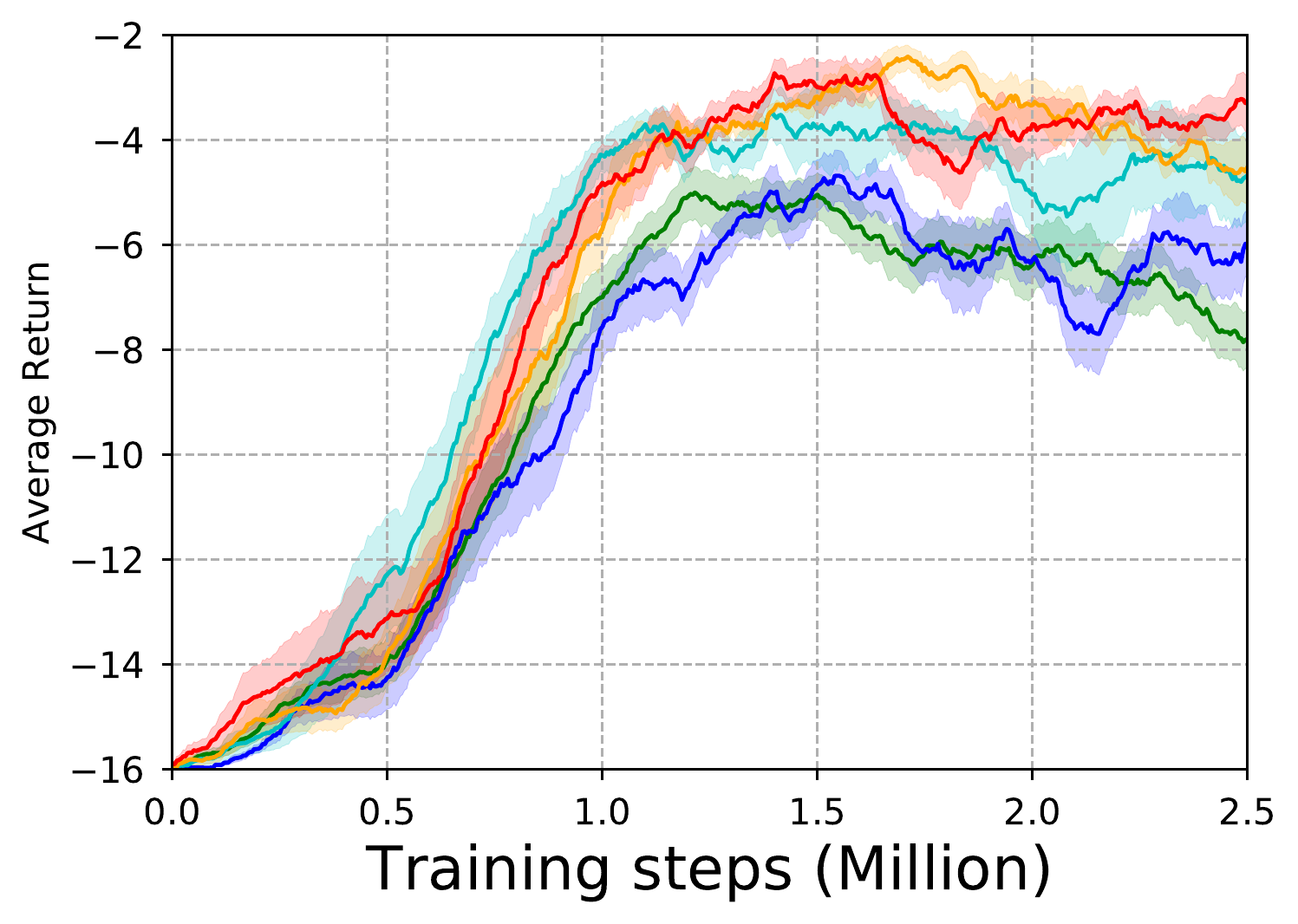}}
	\subfloat[Asterix-MinAtar-PLE]{\includegraphics[width=0.23\textwidth]{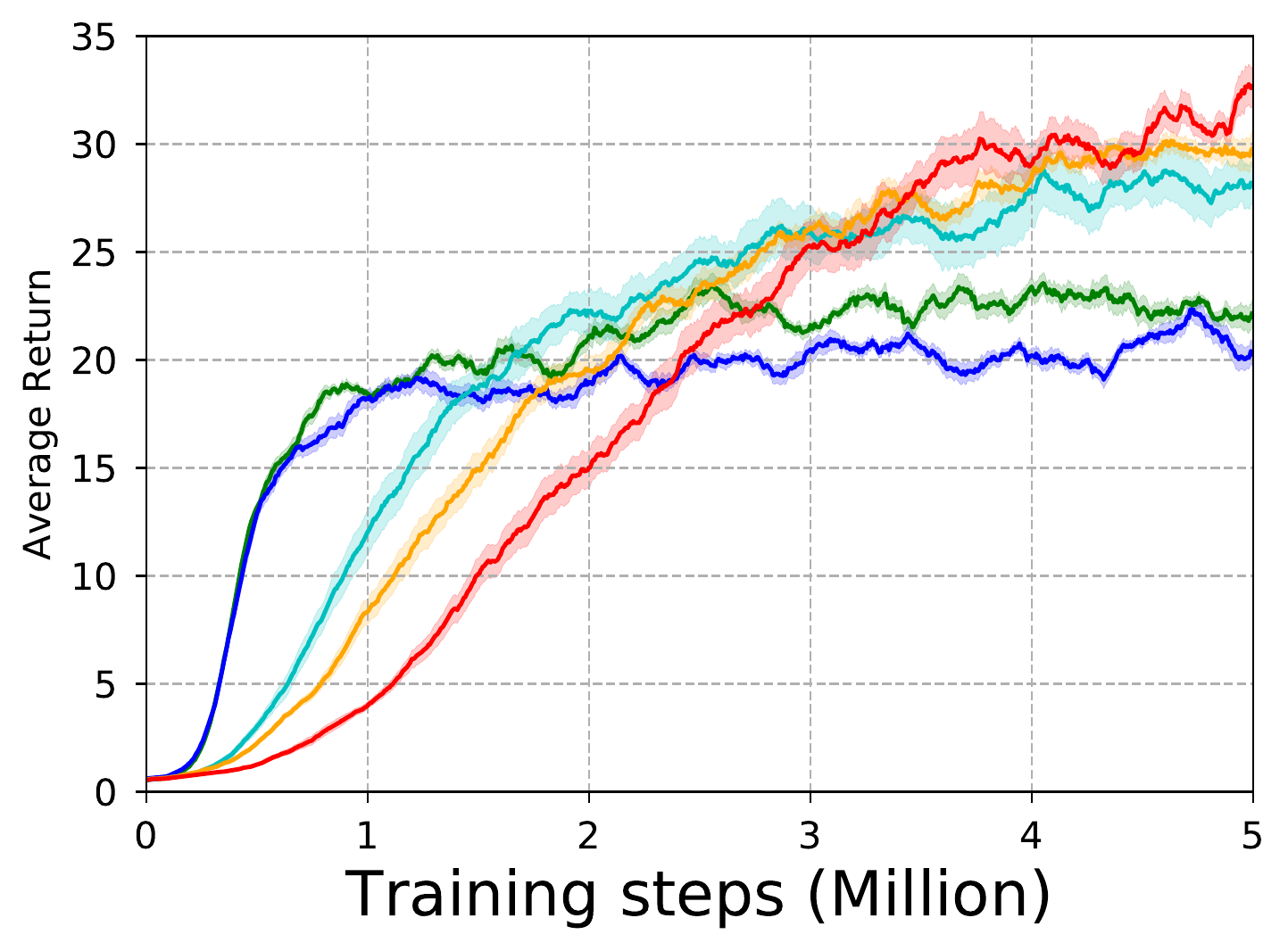}}\hfill\\
	\caption{Learning curves on the PLE and MinAtar environments. The depicted return is averaged over 10 test episodes every 5000 steps across 5 independent runs.}
	\label{fig:single}
\end{figure}

\begin{table}[t!]
	\centering
	\setlength{\tabcolsep}{1.6mm}{
		\begin{tabular}{cccccc}
			\toprule
			\multicolumn{1}{c}{Algorithm} &
			\multicolumn{1}{c}{\begin{tabular}[c]{@{}c@{}} DQN \end{tabular}} &
			\multicolumn{1}{c}{\begin{tabular}[c]{@{}c@{}} DDQN\end{tabular}} &
			\multicolumn{1}{c}{\begin{tabular}[c]{@{}c@{}} SDQN \end{tabular}} &
			\multicolumn{1}{c}{\begin{tabular}[c]{@{}c@{}} MDQN \end{tabular}} &
			\multicolumn{1}{c}{\begin{tabular}[c]{@{}c@{}} SM2 \end{tabular}}  \\
			\midrule
			Catcher & 
			\begin{tabular}[c]{@{}c@{}}52.47\\(3.48)\end{tabular} & \begin{tabular}[c]{@{}c@{}}52.06\\(1.30)\end{tabular} & \begin{tabular}[c]{@{}c@{}}56.36\\(2.21)\end{tabular} & \begin{tabular}[c]{@{}c@{}}55.53\\(1.32)\end{tabular} & \begin{tabular}[c]{@{}c@{}}\textbf{57.49}\\(0.84)\end{tabular}  \\
			\midrule
			Pixelcopter & \begin{tabular}[c]{@{}c@{}}25.95\\(11.02)\end{tabular} & \begin{tabular}[c]{@{}c@{}}28.37\\(10.61)\end{tabular} & \begin{tabular}[c]{@{}c@{}}31.79\\(8.69)\end{tabular} & \begin{tabular}[c]{@{}c@{}}28.27\\(3.11)\end{tabular} & \begin{tabular}[c]{@{}c@{}}\textbf{40.11}\\(5.57)\end{tabular} \\
			\midrule
			Pong & 
			\begin{tabular}[c]{@{}c@{}}-8.67\\(2.63)\end{tabular} & \begin{tabular}[c]{@{}c@{}}-6.15\\(2.26)\end{tabular} & \begin{tabular}[c]{@{}c@{}}-4.59\\(2.06)\end{tabular} & \begin{tabular}[c]{@{}c@{}}-4.70\\(2.88)\end{tabular} & \begin{tabular}[c]{@{}c@{}}\textbf{-3.29}\\(1.81)\end{tabular}  \\
			\midrule
			Asterix & 
			\begin{tabular}[c]{@{}c@{}}22.16\\(2.97)\end{tabular} & \begin{tabular}[c]{@{}c@{}}20.29\\(2.65)\end{tabular} & \begin{tabular}[c]{@{}c@{}}29.77\\(3.90)\end{tabular} & \begin{tabular}[c]{@{}c@{}}28.15\\(3.24)\end{tabular} & \begin{tabular}[c]{@{}c@{}}\textbf{32.66}\\(3.08)\end{tabular}  \\
			\bottomrule
	\end{tabular}}
	\caption{Mean of average return for different methods in Catcher (Catcher-PLE), Pixelcopter (Pixelcopter-PLE), Pong (Pong-PLE) and Asterix (Asterix-MinAtar) games (standard deviation in parenthesis).}
	\label{tab:results}
\end{table}

\subsection{Overview}
We performed two types of experiments: one is about single agent learning and the other is about multi-agent learning. For single agent RL, we used several games from PyGame Learning Environment (PLE) \cite{tas} and MinAtar \cite{Kyo}; while for MARL, we adopted StarCraft Multi-agent Challenge (SMAC) \cite{Sam}. For each type of experiments, we used as baselines several related soft operators or approaches aiming to reduce overestimation for Q-learning, including DDQN \cite{Van}, SoftMax \cite{Song} and MellowMax \cite{Asa}.
We choose the candidates $ \alpha $ and $ \omega $ from the set of \{1, 2, 5, 10, 15\} for SM2, and choose $ \omega $ among \{1, 5, 10, 15, 20\} for Softmax \cite{Song}. As for Mellowmax \cite{Asa}, we choose $ \omega $ among \{5, 10,30, 50, 100, 200\} . We select the best hyperparameters for these methods on each environment separately.

\subsection{PLE and MinAtar}

\subsubsection{Experimental Setup.}
For PLE game environments, the neural network was a multi-layer perceptron with hidden layer fixed to [64, 64].
The discount factor was 0.99.
The size of the replay buffer was 10000.
The weights of neural networks were optimized by RMSprop with gradient clip 5.
The batch size was 32.
The target network was updated every 200 frames.
$\epsilon$-greedy was applied as the exploration policy with $\epsilon$ decreasing linearly from 1.0 to 0.01 in 1, 000 steps.
After 1, 000 steps, $\epsilon$ was fixed to 0.01.
The rest of the experimental setting are detailed in the appendix.
\begin{figure}[t!]  
	\centering
	\subfloat[Catcher-PLE]{\includegraphics[width=0.23\textwidth]{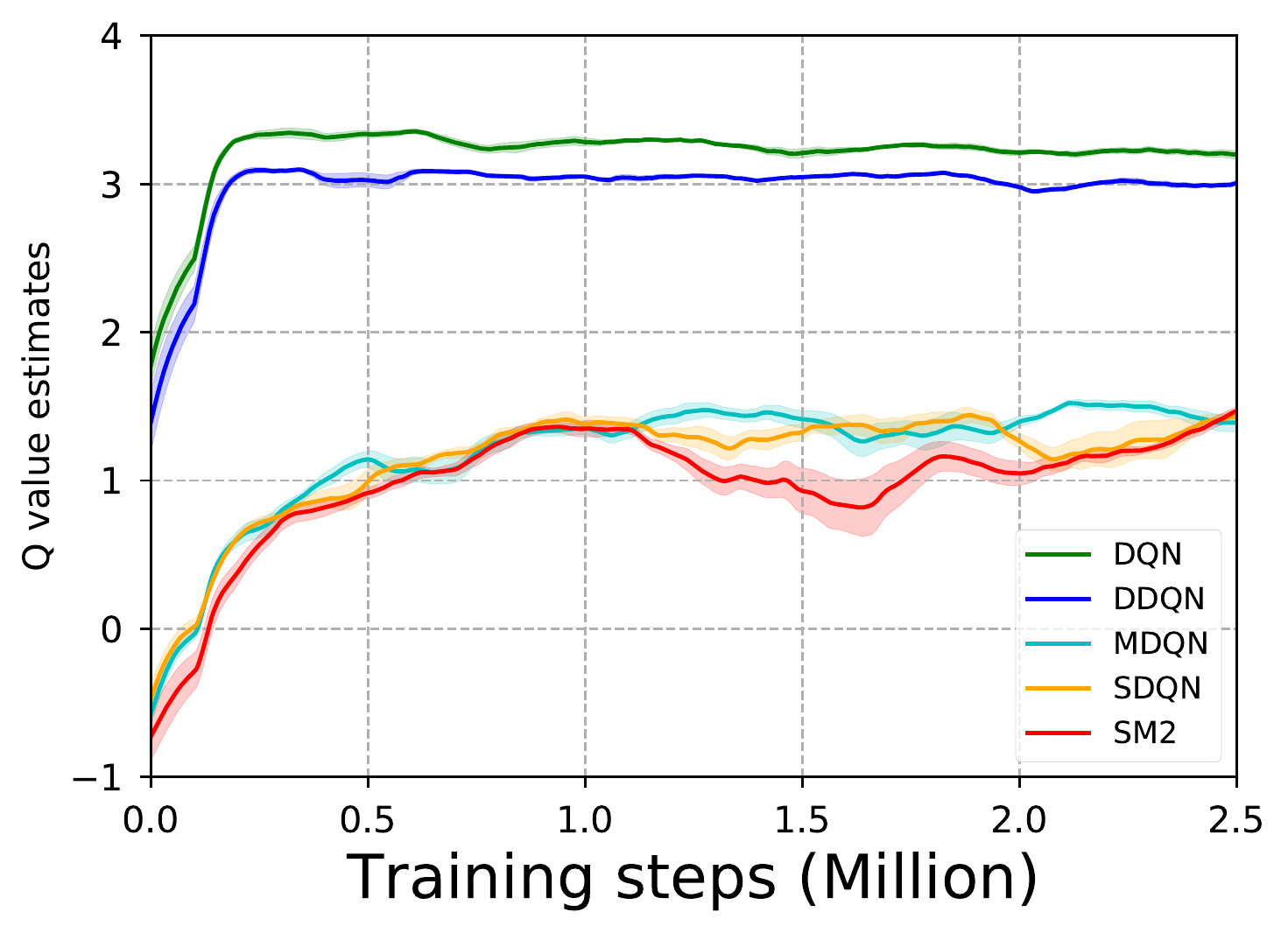}}\hfill
	\subfloat[Asterix-MinAtar]{\includegraphics[width=0.23\textwidth]{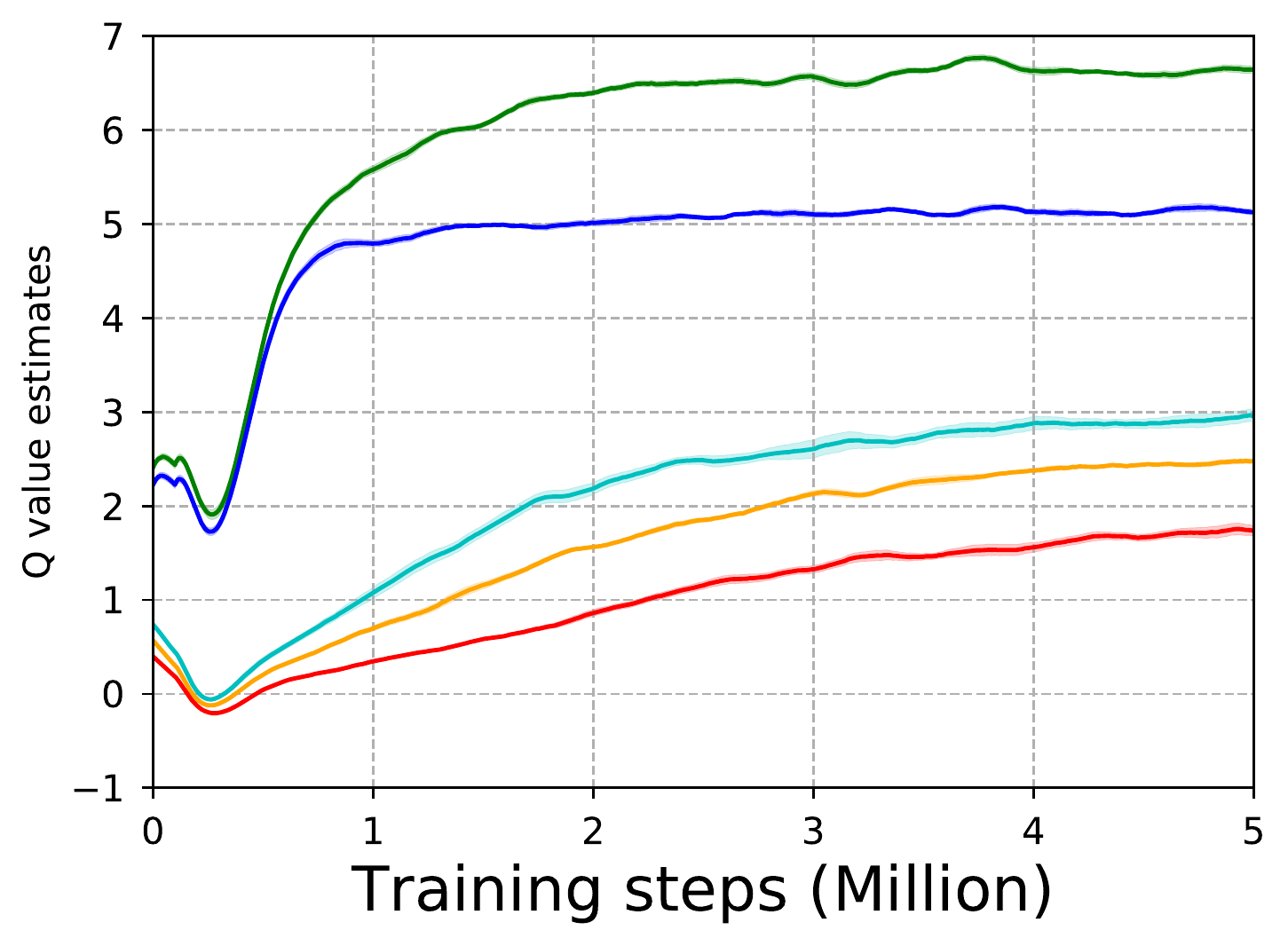}}
	\caption{The estimated $ Q $ values for our methods SM2, and comparison methods. The mean and a standard deviation are shown across 5 independent runs.}
	\label{fig:singleQvalue}
\end{figure}

\begin{figure*}[t!]
	\centering
	\subfloat[3m]{\includegraphics[width=0.25\textwidth]{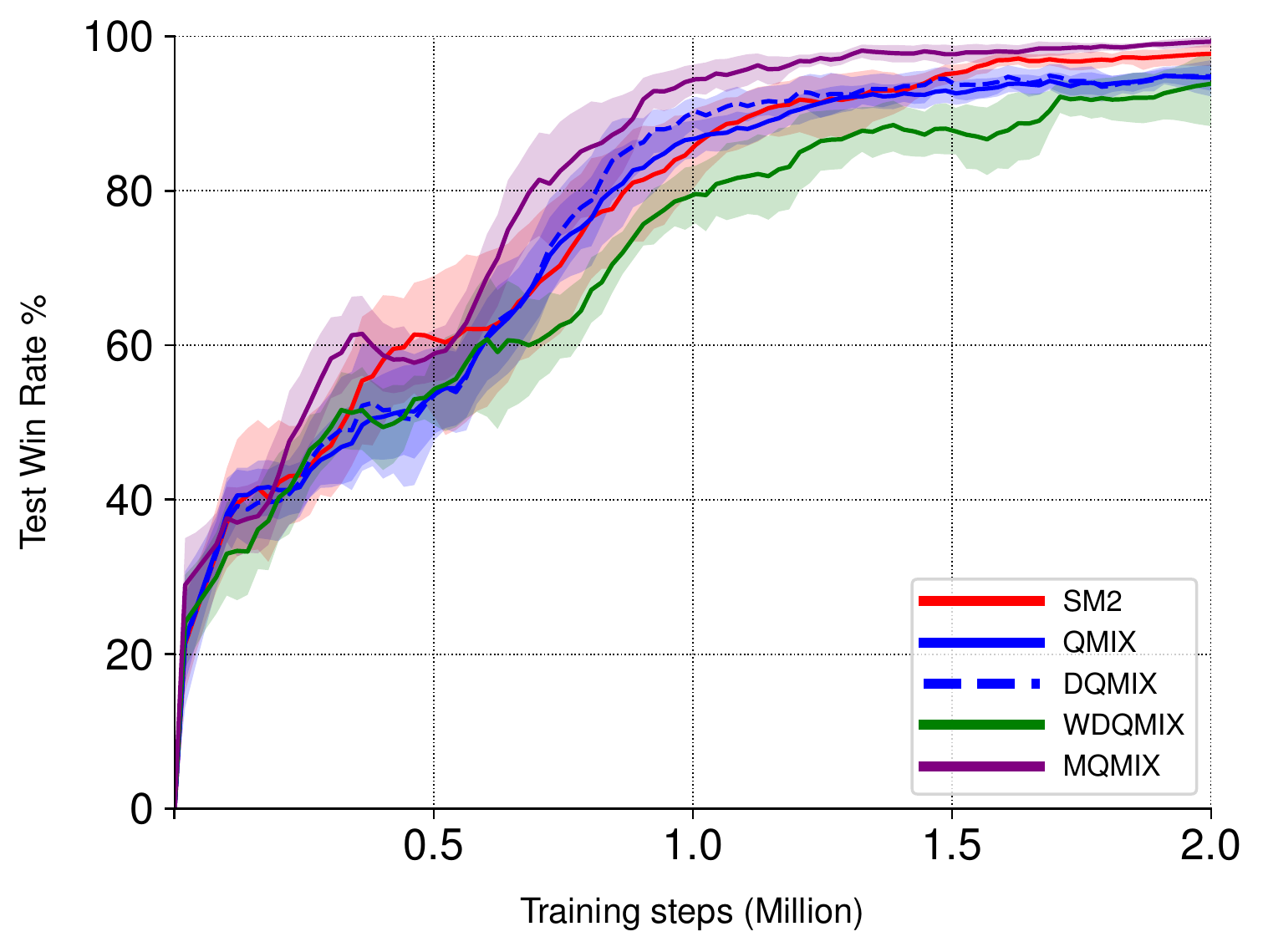}}
	\subfloat[8m]{\includegraphics[width=0.25\textwidth]{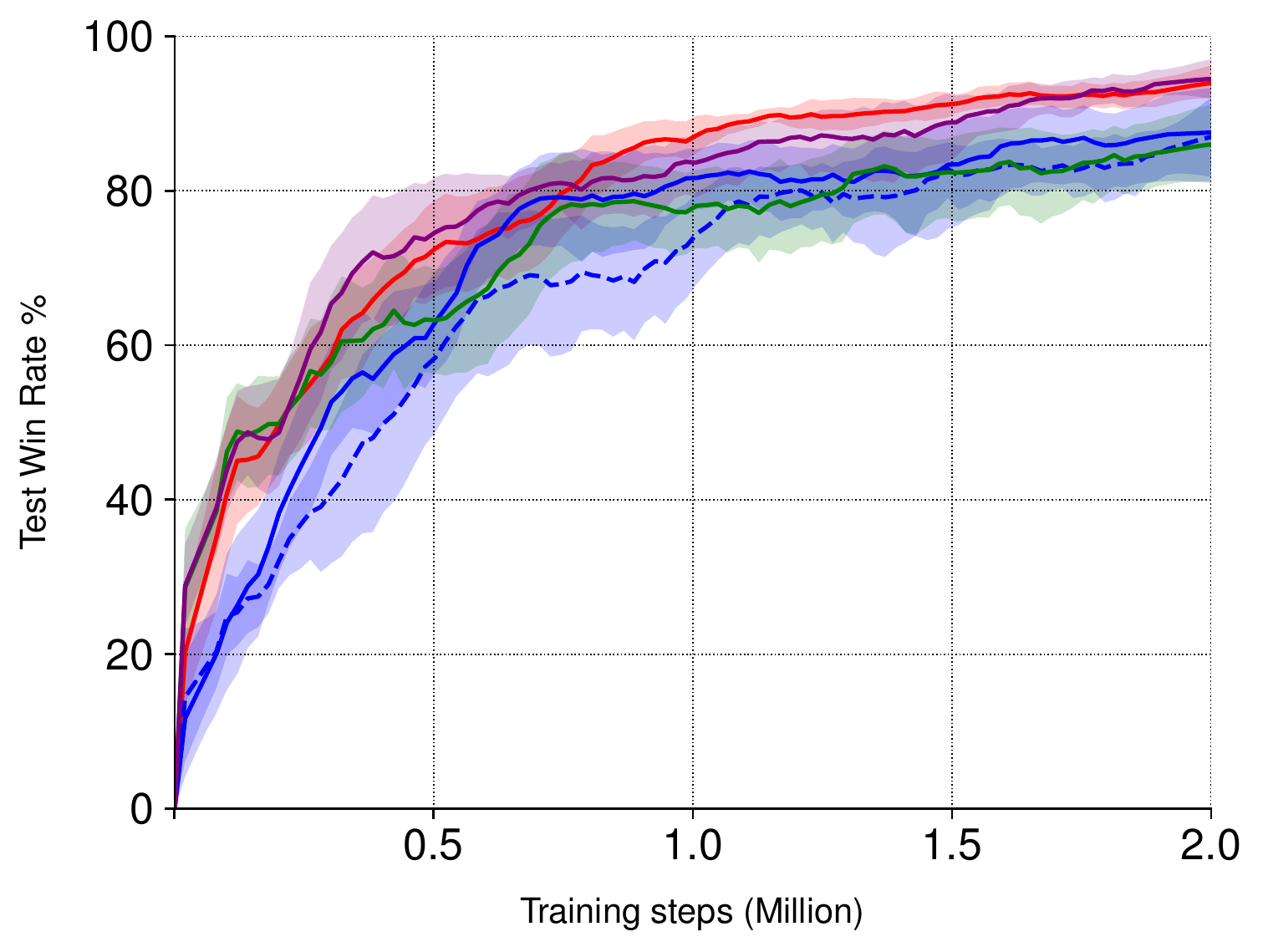}}
	\subfloat[3s5z]{\includegraphics[width=0.25\textwidth]{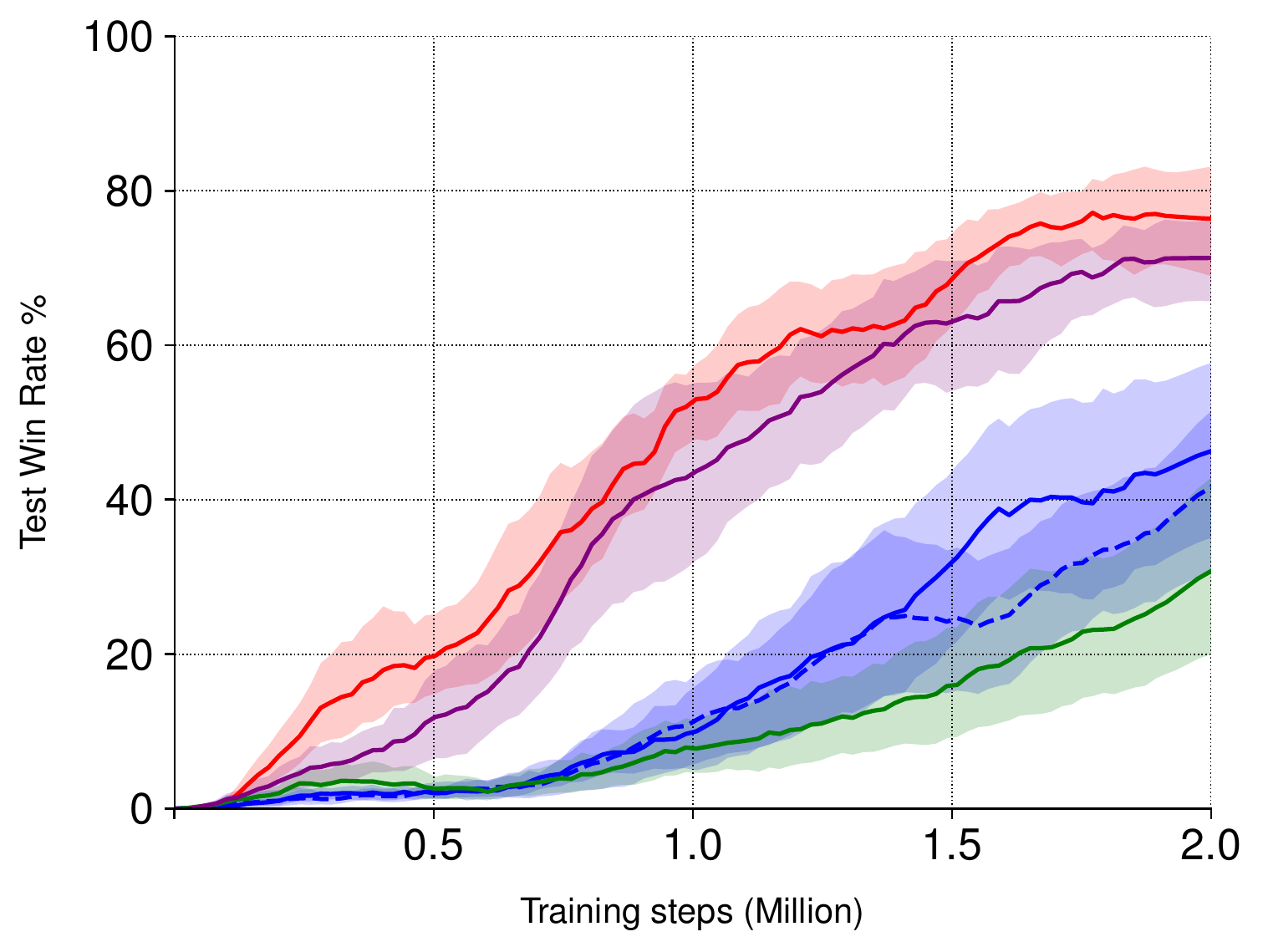}}
	\subfloat[2s\_vs\_1sc]{\includegraphics[width=0.25\textwidth]{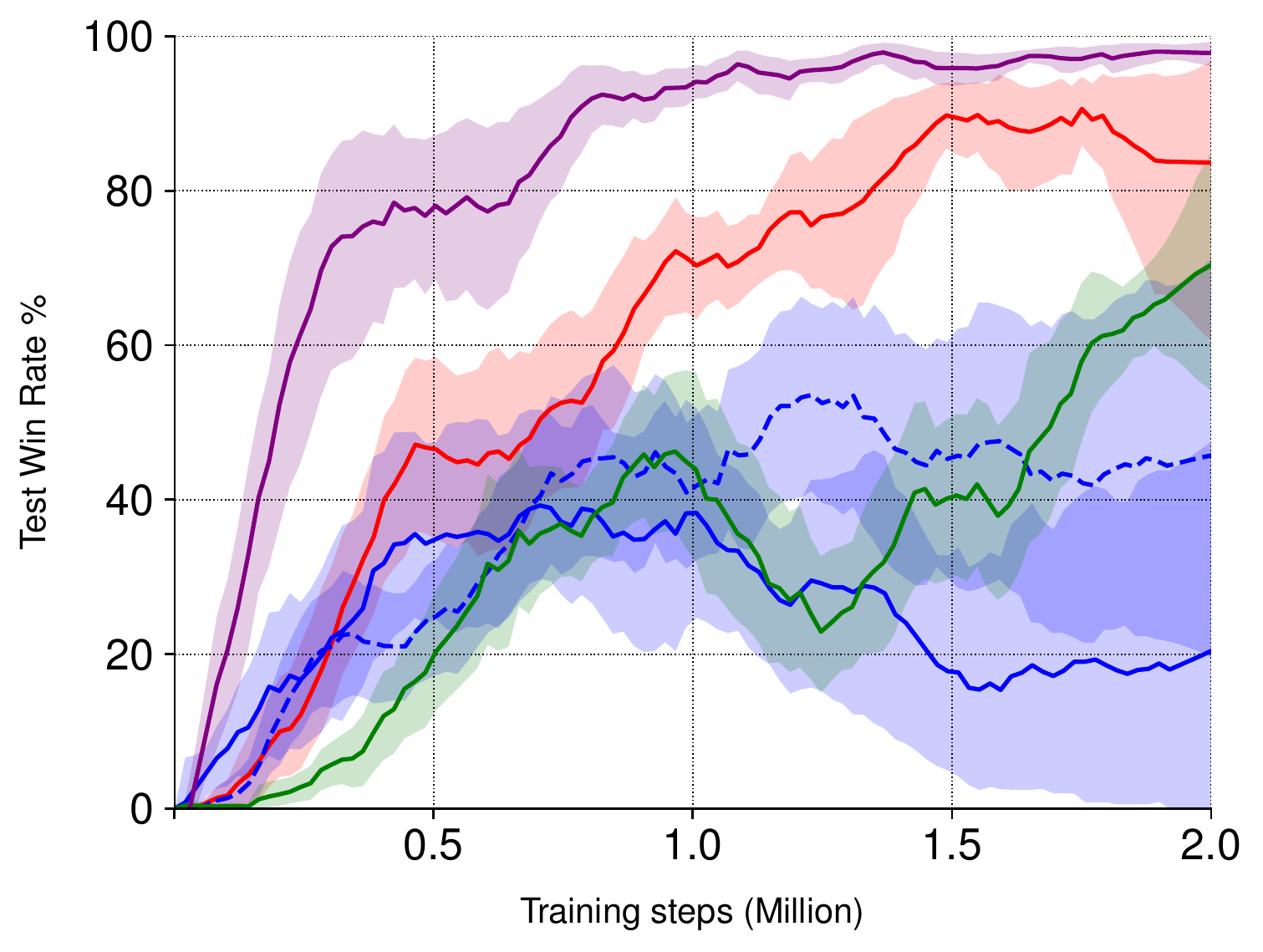}}\\
	\subfloat[3s6z]{\includegraphics[width=0.25\textwidth]{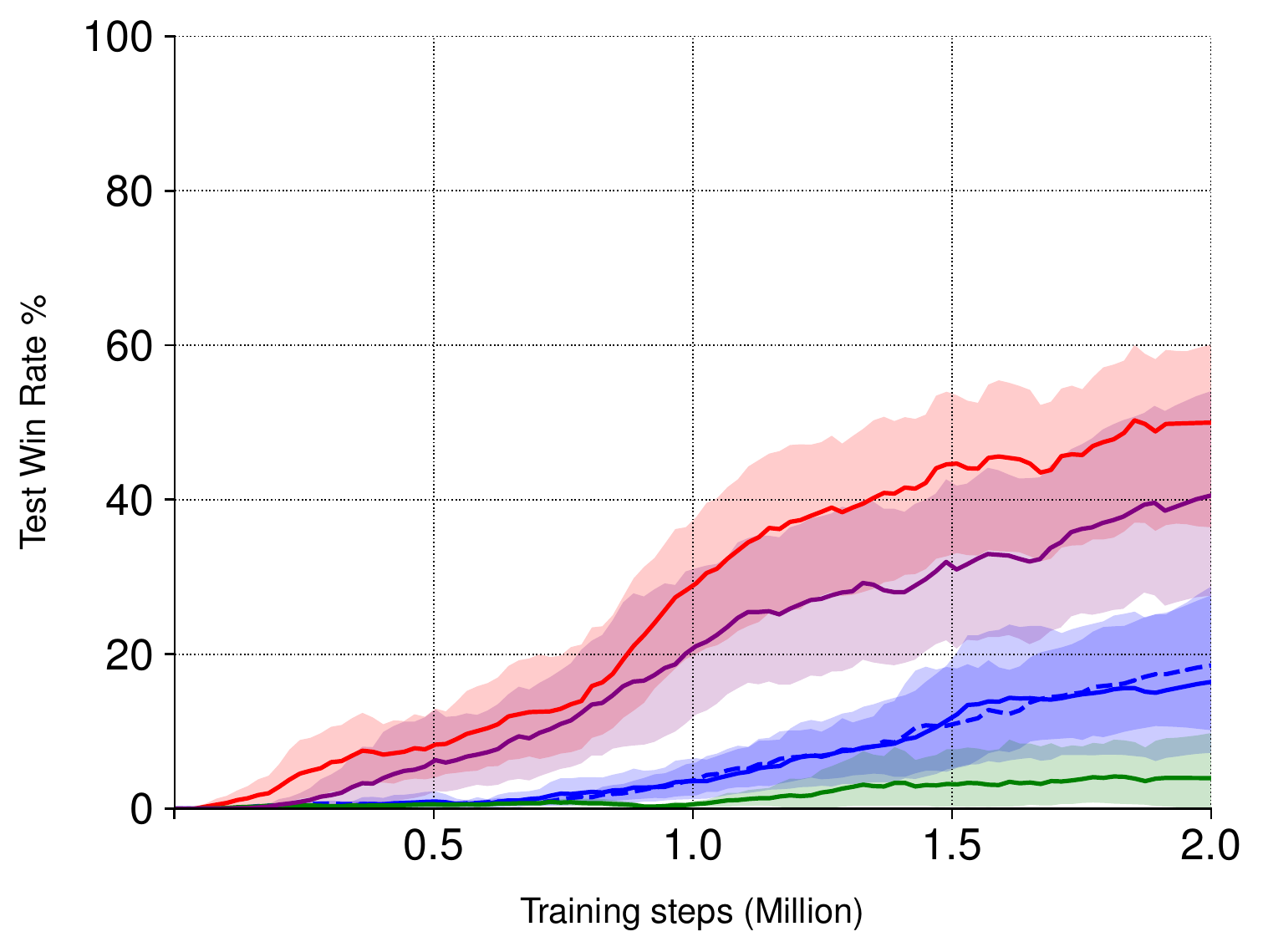}}
	\subfloat[1c3s5z]{\includegraphics[width=0.25\textwidth]{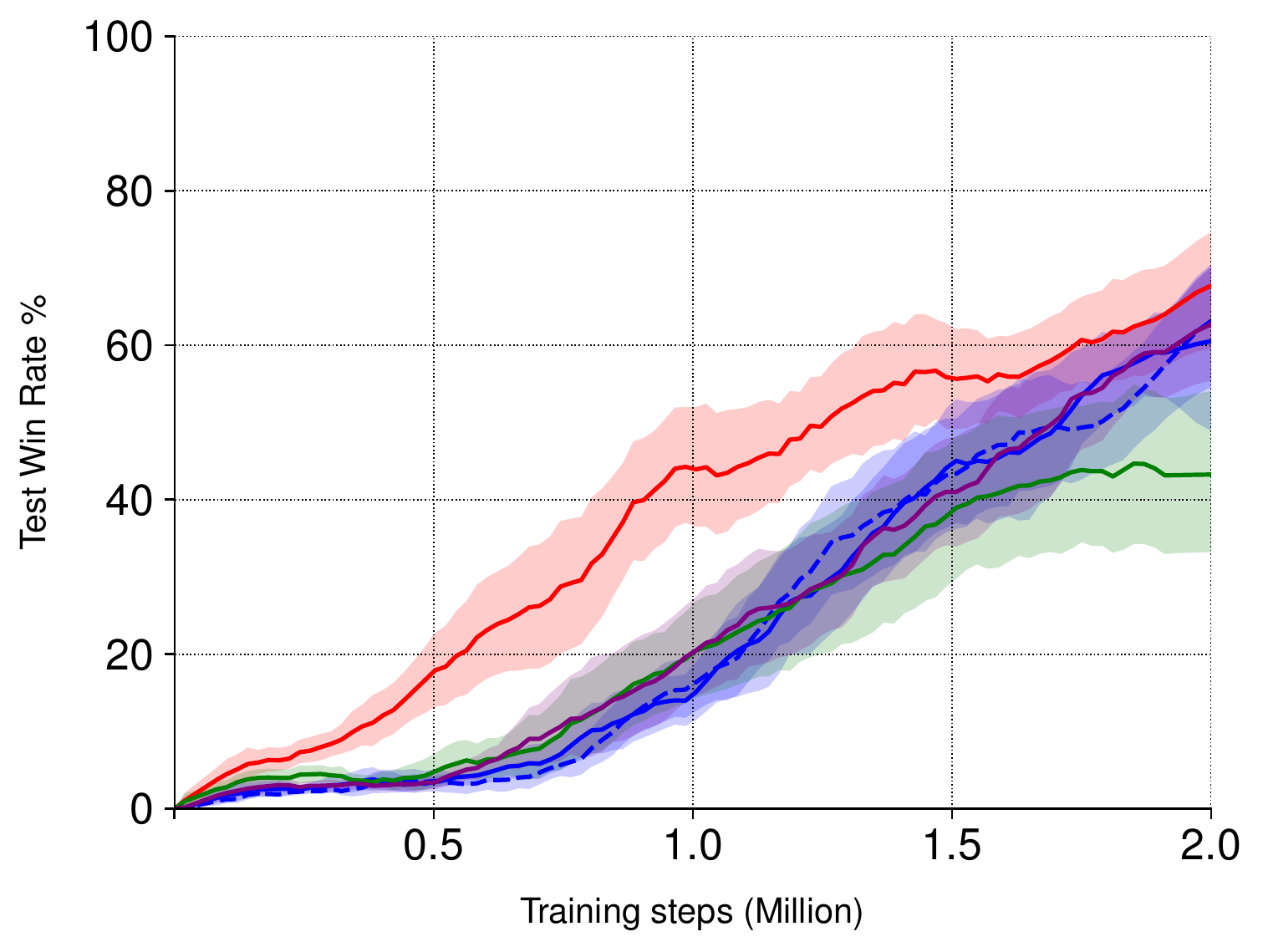}}
	\subfloat[3s5z]{\includegraphics[width=0.25\textwidth]{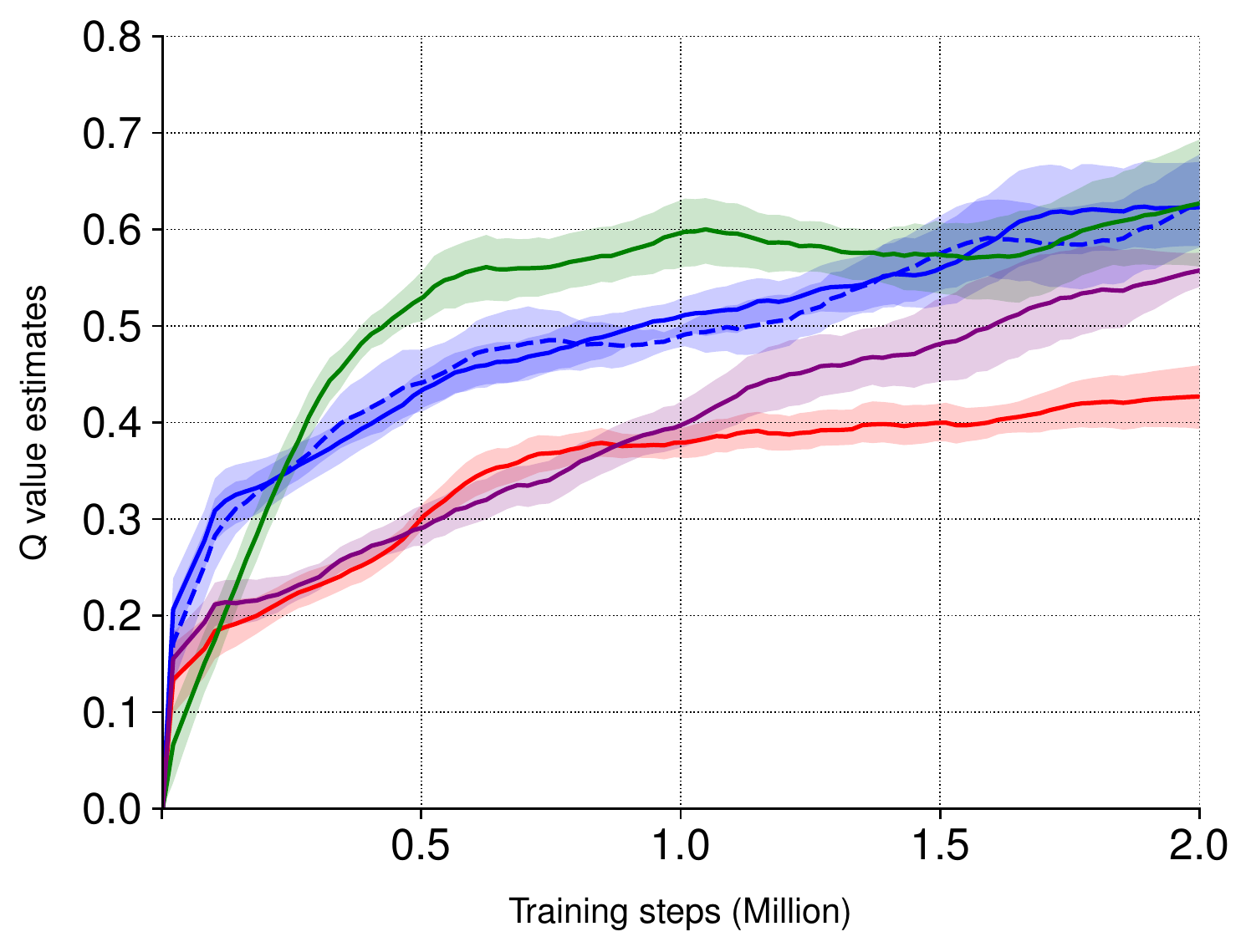}}\hfill
	\subfloat[1c3s5z]{\includegraphics[width=0.25\textwidth]{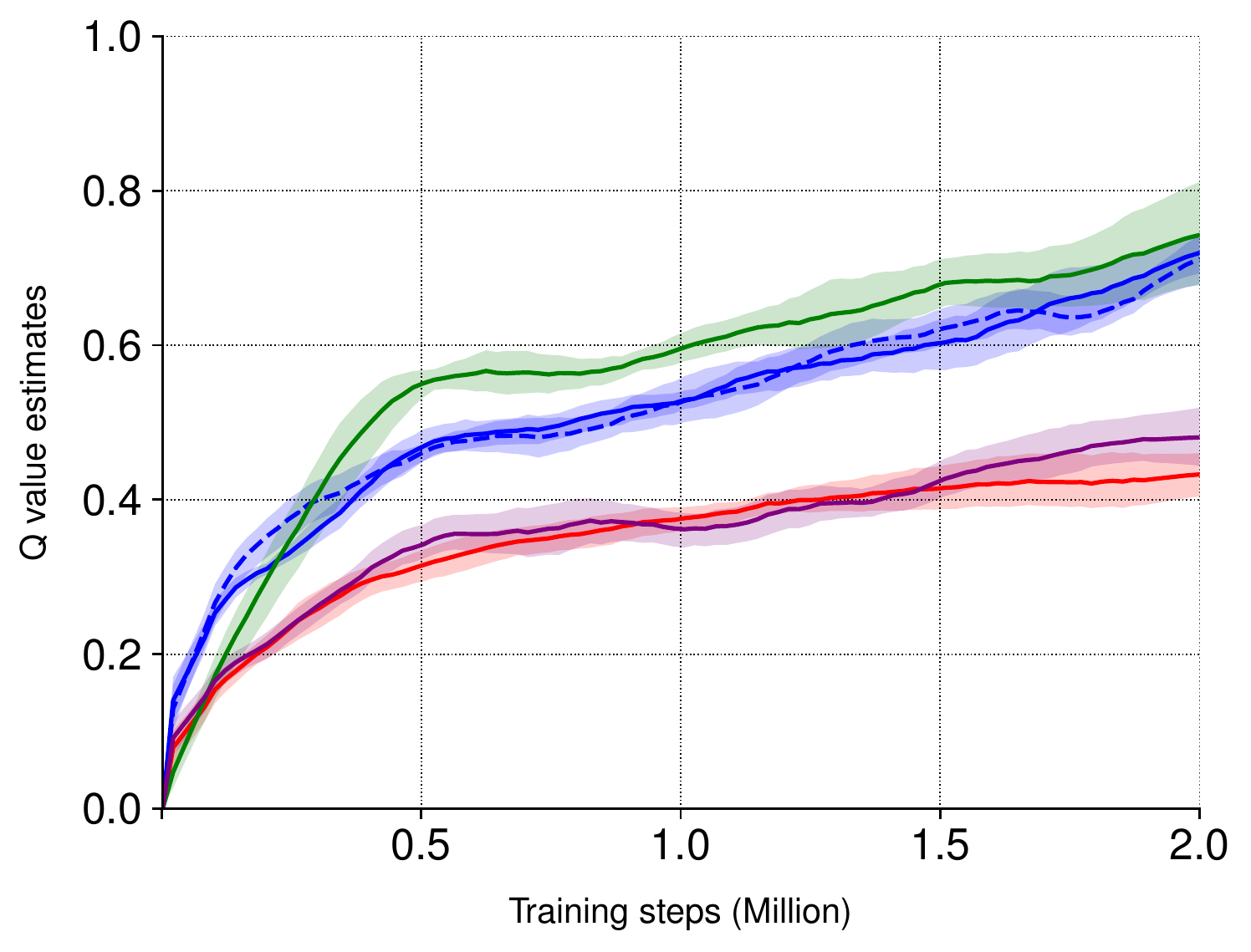}}\\
	\caption{Test win rates for our methods SM2 with $\alpha=10$ and $\omega=5$, and comparison methods QMIX, double-QMIX (DQMIX) and Mellowmax-QMIX (MQMIX) in six different scenarios (a)-(f).
	The estimated Q-values for our methods SM2 and other algorithms in (g)-(h).
	The mean and 95\% confidence interval are shown across 10 independent runs.
	}
	\label{fig:overestimation}
\end{figure*}

\begin{figure}[t!]
	\centering
	\subfloat[3s5z]{\includegraphics[width=0.22\textwidth]{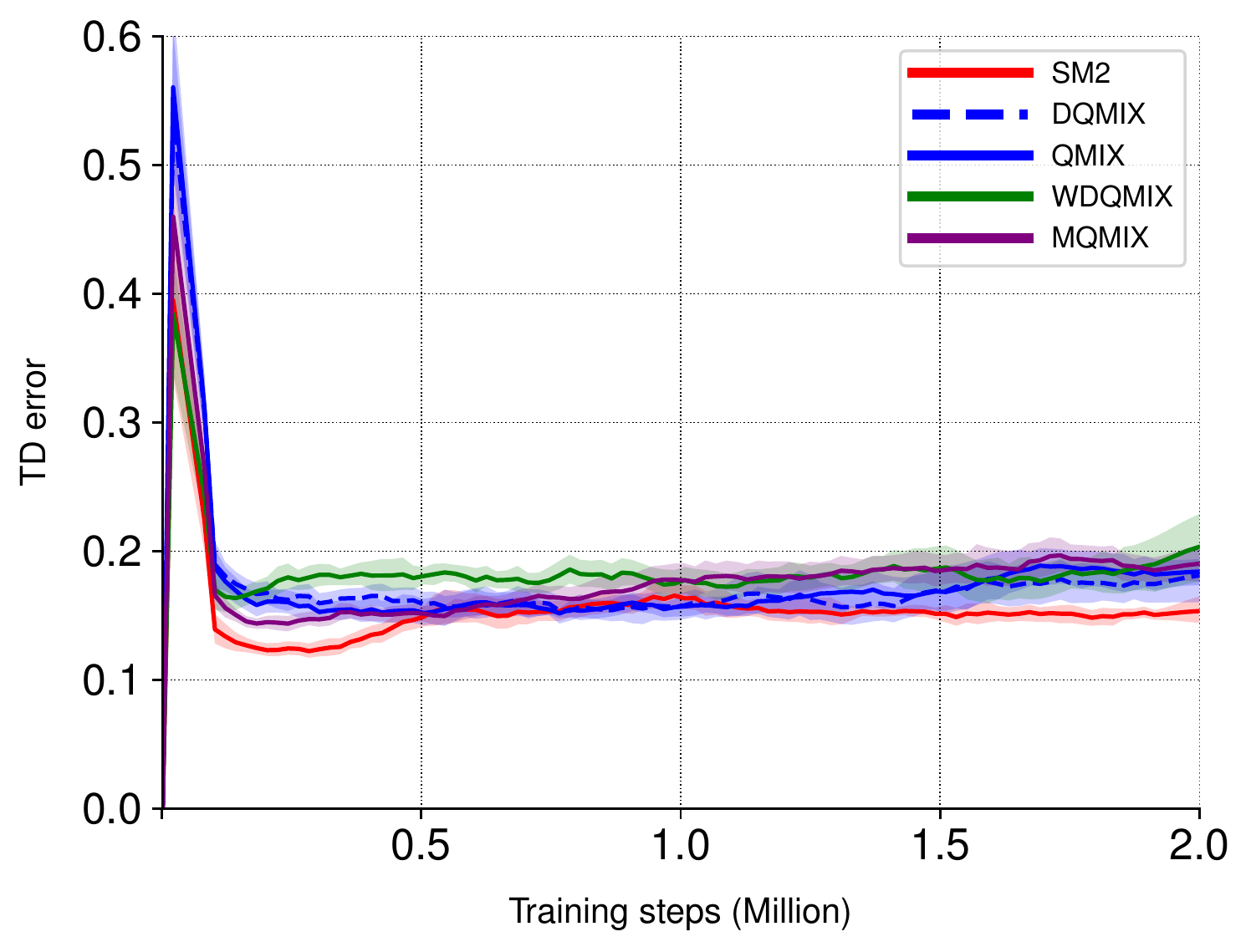}}
	\subfloat[1c3s5z]{\includegraphics[width=0.22\textwidth]{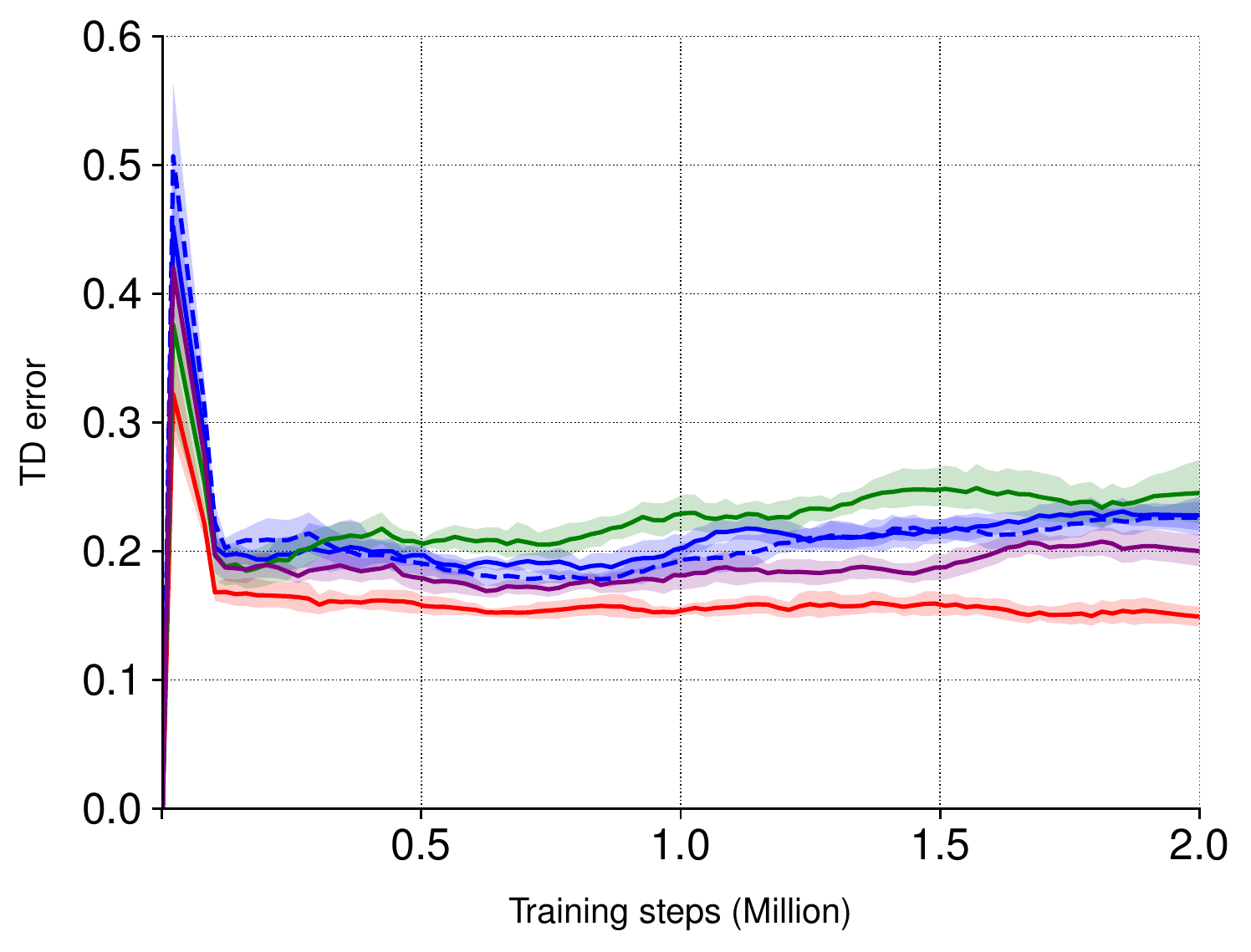}}
	\caption{The TD error of different algorithms are evaluated in six different scenarios.
	The mean and 95\% confidence interval are shown across 10 independent runs.
	}
	\label{fig:td_error_abs}
\end{figure}

\subsubsection{Performance Comparison.}
As mentioned before, we used the classical DQN algorithm \cite{Mni} and its several variants: DDQN \cite{Van}, Soft DQN (SDQN) \cite{Song} and Mellow DQN (MDQN) \cite{Asa} as baselines for performance comparison. We evaluated the average return of 10 test episodes every 5000 training steps.

Figure \ref{fig:single} shows all the results of each algorithm.
From the figure, one can see that the proposed SM2 performs as well as or better than other algorithms in all the domains. In addition, except the early stages in Asterix, SM2 maintains a faster learning in most of the time. 
Table \ref{tab:results} gives the mean and standard deviation of average returns for all algorithms across the four games at the end of training.
According to these quantitative results, one can see that SM2 achieves the best final performance with the lower variance.

\subsubsection{Overestimation Reduction.}
To verify the capability of the proposed SM2 method to make the training process of DQN more stable, we compared the estimated Q-values (and the corresponding mean and one standard deviation) from DQN, DDQN, MDQN, SDQN and SM2 in Catcher and Asterix tasks.  Figure \ref{fig:singleQvalue} gives the results.
It shows that SM2 can reduce the overestimation bias, which is consistent with the analysis made in Section 3.4. To some extent, this overestimation reduction also helps to explain the higher average return and more stable learning for SM2 in Figure \ref{fig:single}.

\subsection{The StarCraft Multi-agent Challenge}

\subsubsection{Comparison with Other Overestimation Reduction Methods.}
In the first set of experiments, we used the QMIX \cite{Ras} as our basic test bed and adopted several improved algorithms based on it for the overestimation reduction: DQMIX which used double network \cite{Van} in QMIX, WDQMIX  which used weighted double network \cite{Zheng} in QMIX, MQMIX which used mellowmax in QMIX.

Figure \ref{fig:overestimation} shows the mean and 95\% confidence interval of the test win rate over the different algorithms in six different scenarios.
The results suggest that our method SM2 achieves a higher win rate and faster learning than QMIX, DQMIX and WDQMIX.
As for MQMIX, our SM2 is better or comparable in five of the six tasks.
From these performance curves, we can also see that SM2 not only achieves a better performance, but also improves the sample efficiency in most scenarios.
Note that DQMIX doesn't obviously improve the performance compared with QMIX except in 2s\_vs\_1sc map, and sometimes its performance may be worse.
This suggests that the double Q networks used in MARL can not have the same effect as in the single-agent setting.
In 2s\_vs\_1sc, it can be seen that QMIX has fallen into a suboptimal policy, and cannot jump out.
MQMIX can achieve better performance - however, its performance is sensitive to the parameter $\omega$, which has to be tuned for each task. For example, in 2s\_vs\_1sc, $\omega$ = 5 of mellowmax can get better performance, while $\omega$ = 200 in 1c3s5z map.

\begin{table*}[t!]
	\centering
	\setlength{\tabcolsep}{1mm}{
		\begin{tabular}{ccccccccccccc}
			\toprule
			Algorithm &
			\begin{tabular}[c]{@{}c@{}} QMIX \end{tabular} &
			\begin{tabular}[c]{@{}c@{}} MQMIX\\$\omega=5$\end{tabular} &
			\begin{tabular}[c]{@{}c@{}} MQMIX\\$\omega=10$ \end{tabular} &
			\begin{tabular}[c]{@{}c@{}} MQMIX\\$\omega=30$ \end{tabular} &
			\begin{tabular}[c]{@{}c@{}} MQMIX\\$\omega=50$ \end{tabular} &
			\begin{tabular}[c]{@{}c@{}} MQMIX\\$\omega=100$\end{tabular} &
			\begin{tabular}[c]{@{}c@{}} MQMIX\\$\omega=200$\end{tabular} &
			\begin{tabular}[c]{@{}c@{}} SM2\\$\alpha=1$,\\ $\omega=10$\end{tabular} &
			\begin{tabular}[c]{@{}c@{}} SM2\\$\alpha=5$,\\ $\omega=5$\end{tabular} &
			\begin{tabular}[c]{@{}c@{}} SM2\\$\alpha=5$,\\ $\omega=10$\end{tabular} &
		    \begin{tabular}[c]{@{}c@{}} SM2\\$\alpha=10$,\\ $\omega=5$\end{tabular} &	
			\begin{tabular}[c]{@{}c@{}} SM2\\$\alpha=10$,\\ $\omega=10$ \end{tabular} \\
			\hline
			3m & 
			\begin{tabular}[c]{@{}c@{}}94.57\\(3.30)\end{tabular} & \begin{tabular}[c]{@{}c@{}}65.08\\(23.54)\end{tabular} & \begin{tabular}[c]{@{}c@{}}97.56\\(3.55)\end{tabular} & \begin{tabular}[c]{@{}c@{}}99.23\\(1.01)\end{tabular} &
			\begin{tabular}[c]{@{}c@{}}98.51\\(1.39)\end{tabular} & \begin{tabular}[c]{@{}c@{}}97.36\\(2.31)\end{tabular} & \begin{tabular}[c]{@{}c@{}}96.30\\(3.91)\end{tabular} &
			\begin{tabular}[c]{@{}c@{}}97.03\\(2.92)\end{tabular} &
			\begin{tabular}[c]{@{}c@{}}98.80\\(1.62)\end{tabular} & \begin{tabular}[c]{@{}c@{}}95.51\\(3.87)\end{tabular} & \begin{tabular}[c]{@{}c@{}}97.65\\(2.03)\end{tabular} & \begin{tabular}[c]{@{}c@{}}97.30\\(3.41)\end{tabular} \\
			\hline
			8m & 
			\begin{tabular}[c]{@{}c@{}}87.30\\(9.18)\end{tabular} & \begin{tabular}[c]{@{}c@{}}1.24\\(2.26)\end{tabular} & \begin{tabular}[c]{@{}c@{}}7.42\\(7.95)\end{tabular} & \begin{tabular}[c]{@{}c@{}}90.87\\(6.10)\end{tabular} &
			\begin{tabular}[c]{@{}c@{}}94.35\\(3.65)\end{tabular} &
			\begin{tabular}[c]{@{}c@{}}94.80\\(3.45)\end{tabular} & \begin{tabular}[c]{@{}c@{}}94.19\\(3.98)\end{tabular} &
			\begin{tabular}[c]{@{}c@{}}90.27\\(4.12)\end{tabular} &
			\begin{tabular}[c]{@{}c@{}}94.73\\(3.16)\end{tabular} & \begin{tabular}[c]{@{}c@{}}93.81\\(3.76)\end{tabular} & \begin{tabular}[c]{@{}c@{}}93.72\\(3.06)\end{tabular} & \begin{tabular}[c]{@{}c@{}}92.78\\(3.65)\end{tabular} \\
			\hline
			3s5z & 
			\begin{tabular}[c]{@{}c@{}}45.67\\(18.59)\end{tabular} & \begin{tabular}[c]{@{}c@{}}3.49\\(6.68)\end{tabular}   & \begin{tabular}[c]{@{}c@{}}3.35\\(3.11)\end{tabular}   & \begin{tabular}[c]{@{}c@{}}22.97\\(13.62)\end{tabular} &
			\begin{tabular}[c]{@{}c@{}}51.85\\(12.56)\end{tabular} &
			\begin{tabular}[c]{@{}c@{}}71.29\\(8.07)\end{tabular}  & \begin{tabular}[c]{@{}c@{}}63.32\\(12.78)\end{tabular} &
		    \begin{tabular}[c]{@{}c@{}}74.16\\(4.59)\end{tabular}  &
			\begin{tabular}[c]{@{}c@{}}73.74\\(13.43)\end{tabular} & \begin{tabular}[c]{@{}c@{}}76.85\\(9.69)\end{tabular}  & \begin{tabular}[c]{@{}c@{}}76.47\\(10.10)\end{tabular} & \begin{tabular}[c]{@{}c@{}}75.42\\(3.65)\end{tabular}  \\
			\hline
			2s\_vs\_1sc & \begin{tabular}[c]{@{}c@{}}19.82\\(36.56)\end{tabular} & \begin{tabular}[c]{@{}c@{}}97.89\\(2.13)\end{tabular}  & \begin{tabular}[c]{@{}c@{}}95.44\\(4.83)\end{tabular}  &
			\begin{tabular}[c]{@{}c@{}}83.06\\(28.27)\end{tabular} & \begin{tabular}[c]{@{}c@{}}85.55\\(19.16)\end{tabular} &
			\begin{tabular}[c]{@{}c@{}}66.66\\(37.79)\end{tabular} &
			\begin{tabular}[c]{@{}c@{}}50.19\\(40.67)\end{tabular} &
		    \begin{tabular}[c]{@{}c@{}}95.34\\(7.89)\end{tabular}  & \begin{tabular}[c]{@{}c@{}}88.86\\(20.23)\end{tabular} &
			\begin{tabular}[c]{@{}c@{}}92.42\\(9.53)\end{tabular}  & \begin{tabular}[c]{@{}c@{}}83.69\\(31.08)\end{tabular} & \begin{tabular}[c]{@{}c@{}}92.61\\(4.25)\end{tabular}  \\
			\hline
			3s6z & 
			\begin{tabular}[c]{@{}c@{}}16.14\\(16.06)\end{tabular} & \begin{tabular}[c]{@{}c@{}}3.83\\(5.73)\end{tabular}   & \begin{tabular}[c]{@{}c@{}}4.48\\(6.56)\end{tabular}   & \begin{tabular}[c]{@{}c@{}}18.06\\(10.70)\end{tabular} &
			\begin{tabular}[c]{@{}c@{}}26.75\\(15.98)\end{tabular} &
			\begin{tabular}[c]{@{}c@{}}40.06\\(19.99)\end{tabular} & \begin{tabular}[c]{@{}c@{}}32.21\\(21.27)\end{tabular} &
			\begin{tabular}[c]{@{}c@{}}55.51\\(6.03)\end{tabular}  &
			\begin{tabular}[c]{@{}c@{}}58.69\\(15.48)\end{tabular} & \begin{tabular}[c]{@{}c@{}}50.41\\(18.77)\end{tabular} & \begin{tabular}[c]{@{}c@{}}49.94\\(18.88)\end{tabular} & \begin{tabular}[c]{@{}c@{}}44.91\\(23.29)\end{tabular} \\
			\hline
			1c3s5z & 
			\begin{tabular}[c]{@{}c@{}}60.15\\(15.00)\end{tabular} & \begin{tabular}[c]{@{}c@{}}8.10\\(8.10)\end{tabular}   & \begin{tabular}[c]{@{}c@{}}1.42\\(3.48)\end{tabular}   & \begin{tabular}[c]{@{}c@{}}16.68\\(11.34)\end{tabular} &
			\begin{tabular}[c]{@{}c@{}}27.07\\(17.90)\end{tabular} & \begin{tabular}[c]{@{}c@{}}55.69\\(15.65)\end{tabular} &
			\begin{tabular}[c]{@{}c@{}}61.83\\(10.81)\end{tabular} &
			\begin{tabular}[c]{@{}c@{}}54.69\\(13.38)\end{tabular} &
			\begin{tabular}[c]{@{}c@{}}59.02\\(11.48)\end{tabular} & \begin{tabular}[c]{@{}c@{}}61.61\\(8.03)\end{tabular}  & \begin{tabular}[c]{@{}c@{}}66.80\\(11.72)\end{tabular} & \begin{tabular}[c]{@{}c@{}}63.84\\(8.80)\end{tabular}  \\
			\bottomrule
	\end{tabular}}
	\caption{Mean of average return for different hyperparameters of the algorithms in six different scenarios (standard deviation in parenthesis). 
}
	\label{tab:multi_table}
\end{table*}

\begin{figure}[t!]
	\centering
	\subfloat[3s5z]{\includegraphics[width=0.22\textwidth]{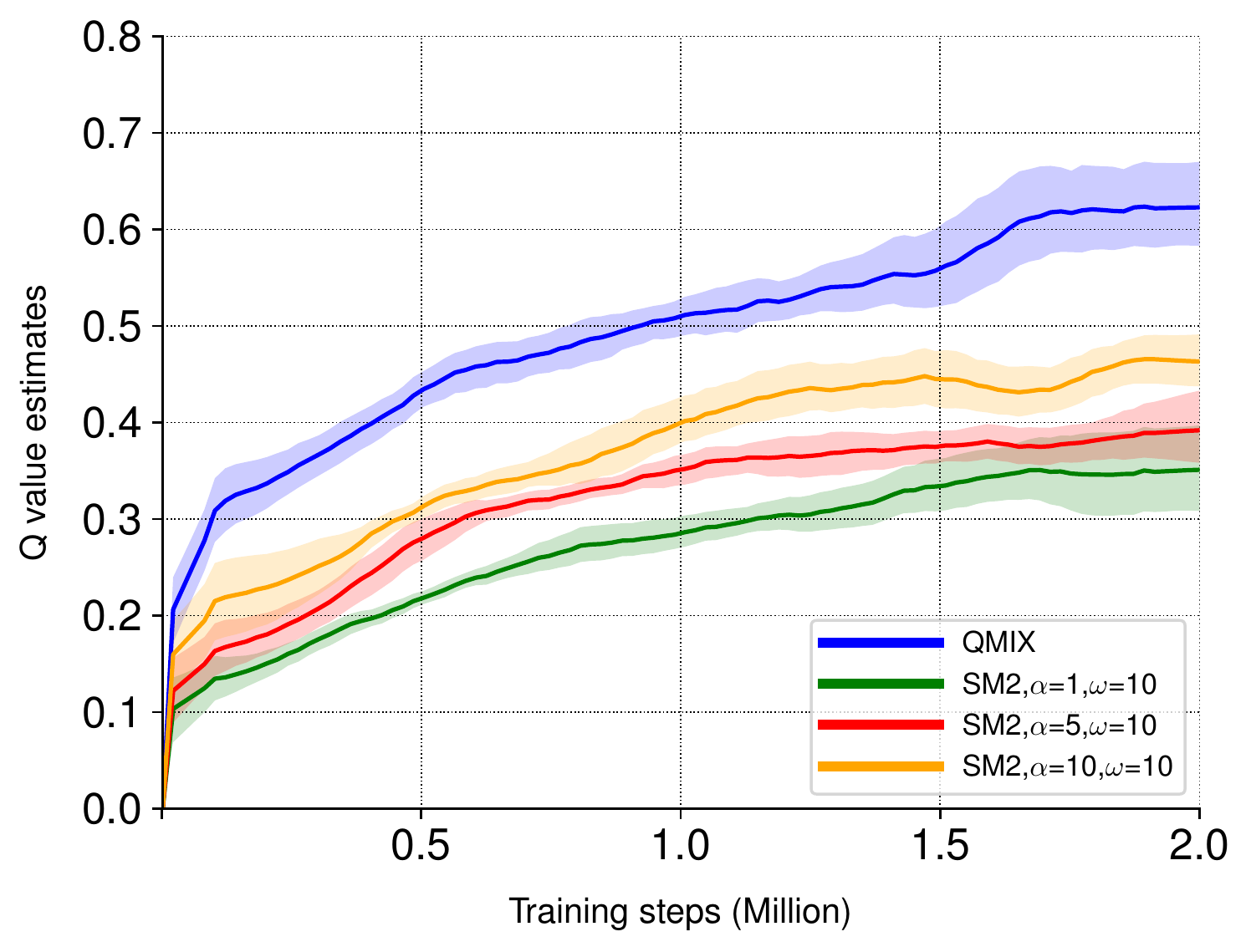}}\hfill
	\subfloat[1c3s5z]{\includegraphics[width=0.22\textwidth]{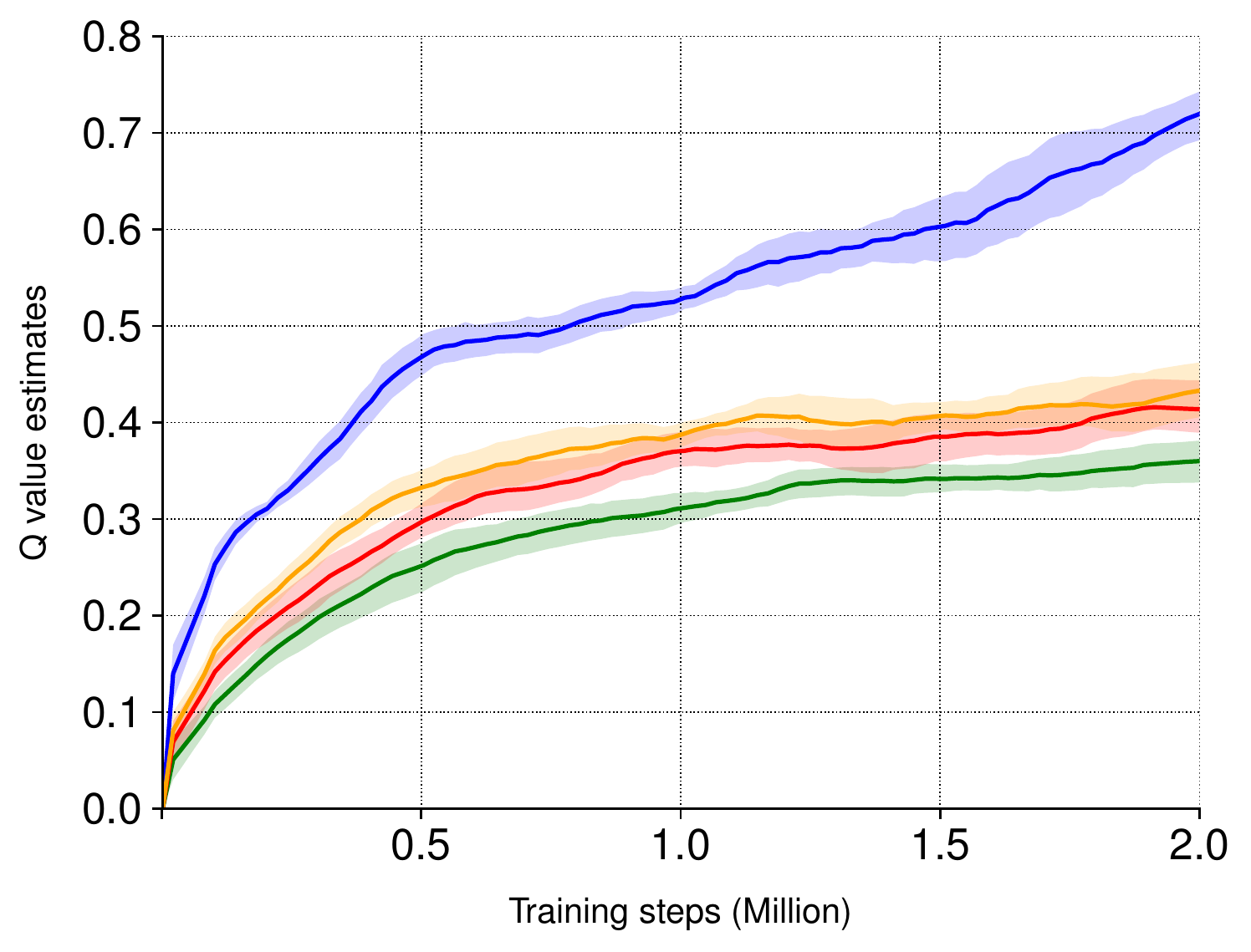}}
	\caption{The estimated $ Q $ values of SM2 under condition $\alpha<\omega$. The mean and 95\% confidence interval are shown across 10 independent runs. }
	\label{fig:alpha}
\end{figure}

By contrast, SM2 can quickly learn a good policy, and doesn't fall into the suboptimal policy using the same set of hyperparameters $\alpha=10$ and $\omega=5$ across all the tasks.


\begin{figure}[t!]
	\centering
	\subfloat[3s5z]{\includegraphics[width=0.22\textwidth]{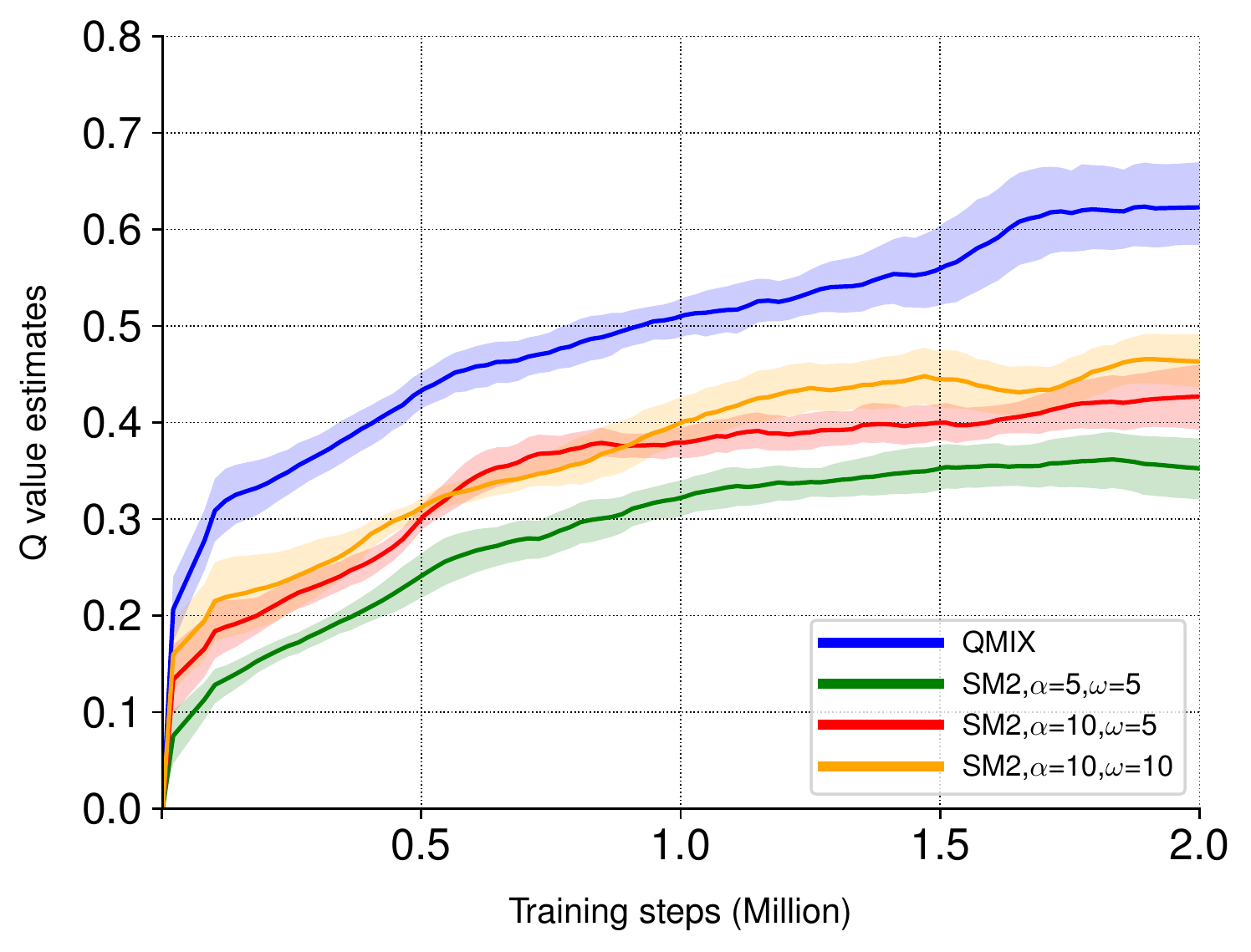}}\hfill
	\subfloat[1c3s5z]{\includegraphics[width=0.22\textwidth]{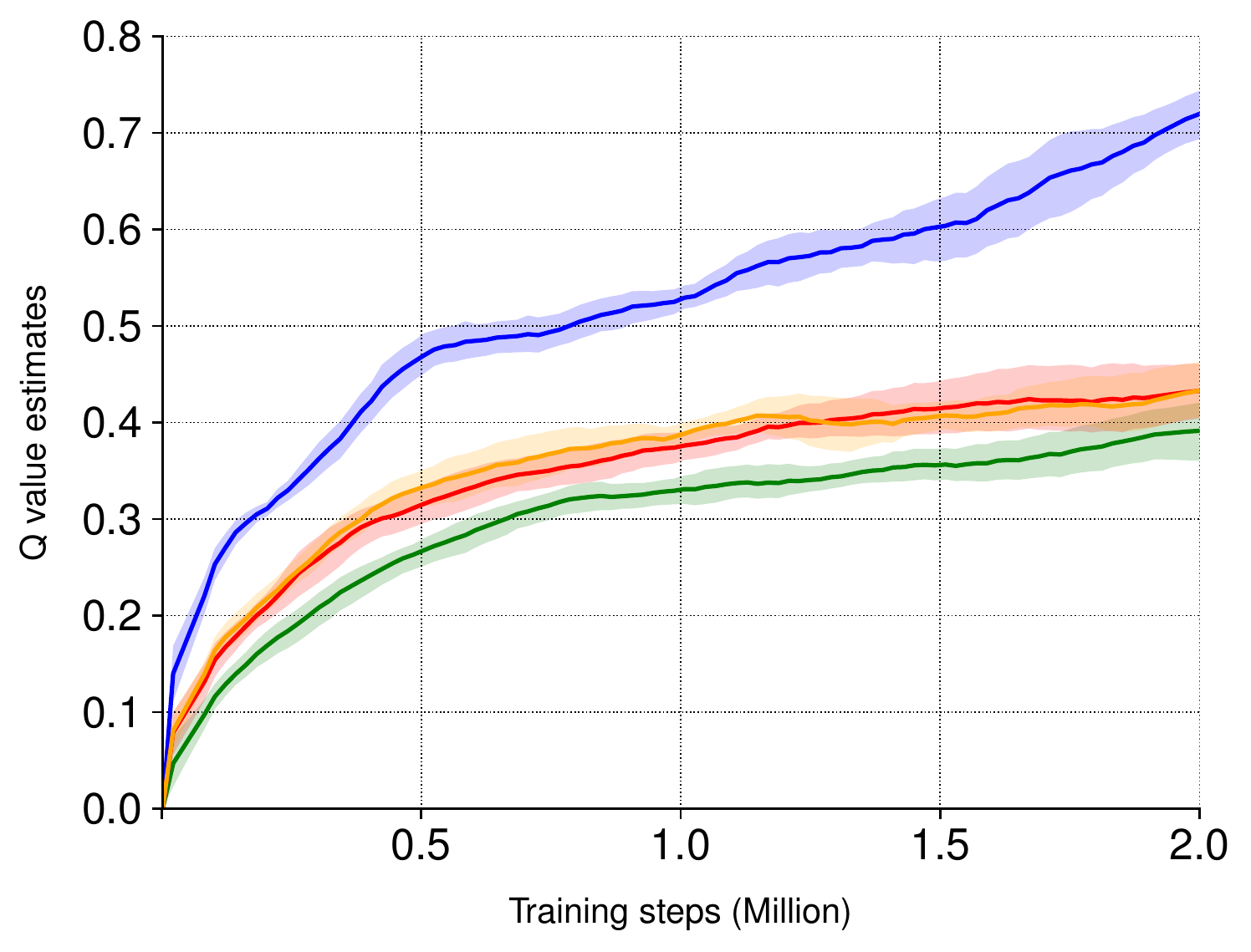}}
	\caption{The estimated $ Q $ values of SM2 under condition $\alpha\geq\omega$. The mean and 95\% confidence interval are shown across 10 independent runs. }
	\label{fig:Qomega}
\end{figure}

\subsubsection{TD error Comparison}
Figure \ref{fig:td_error_abs} shows the TD error of different algorithms evaluated over different scenarios.
The TD error merely at the current state is less representative, so we first sample some batches of state-action pair from the replay buffer, and calculate the absolute value of the TD error for each pair. And finally we choose the mean of them as the estimate of TD error.
It can be seen that the TD error of SM2 is very smaller than other methods, which implies that the target estimates of SM2 are more accurate.
Furthermore, the smaller TD error also makes the training of SM2 more stable.



\subsubsection{Overestimation Reduction.}

Figure \ref{fig:overestimation} gives the estimated $ Q $ values of different algorithms.
We can see that the estimated $ Q $ values of SM2 are much smaller than others, indicating that SM2 reduces the possibility of overestimation.
In Figure \ref{fig:overestimation}, one can see that the reduction of the overestimation makes SM2 perform better, especially on difficult scenarios.
The estimated $ Q $ values of DQMIX are almost the same as QMIX, indicating that it does not reduce the overestimation of $ Q $ values significantly - actually, its performance is also similar to QMIX.
Overall, our algorithm performs more conservatively in the estimation of $ Q $ value, which makes the training more stable and reduces the possibility of falling into the suboptimal policy.

\subsubsection{Sensitiveness to Hyperparameters Setting.}
In this section, we analyze the influence of $\alpha$ and $\omega$ on the estimated $ Q $ value in SM2.
From Figure \ref{fig:alpha} and \ref{fig:Qomega}, two sets of comparative experiments are implemented under two conditions about $\alpha\geq\omega$ and $\alpha<\omega$.
We find that the gap between the $ Q $ values of SM2 and QMIX becomes smaller as the scales of $\alpha$ and $\omega$ increase.
In other words, if $ \alpha $ and $ \omega $ increase, the estimated $ Q $ values will also increase.
This result is consistent with previous theoretical analyses.
Table \ref{tab:multi_table} shows the effect of different hyperparameters of Mellowmax and SM2. This suggests that Mellowmax would have the bad performance with an improper $ \omega $. In contrast, SM2 doesn't require a lot of hyperparameter adjustments across different tasks. When $ \omega=10 $, we find that a small $ \alpha $ can greatly improve the algorithm performance. And different choices of $ \alpha $ and $ \omega $ have little difference on performance, which shows the insensitivity of SM2.

\section{Conclusion and Future Work}\label{Co}

In our work, we propose a new method SM2, which is an enhanced version of the Mellowmax operator.
Our theoretical analysis and extensive experimental results show that SM2 can effectively reduce the overestimation bias, and own the capability to converge to the optimal $ Q $ function in a stable way.
We also show that the overestimation is proportional to the number of agents under some mild conditions in MARL, and our proposed method is beneficial to address this issue and achieves better multi-agent cooperations as well.

In this paper, we only discussed the case where the hyperparameters of SM2 are constant. It's an interesting direction to investigate how to tune them automatically in the future, just like \cite{LinP, Seung}.


\section*{Acknowledgements}
This work is partially supported by National Science Foundation of China (61976115,61672280,61732006), AI+ Project of NUAA(XZA20005, 56XZA18009), research project(315025305).
We would also like to thank the anonymous reviewers, for offering thoughtful comments and helpful advice on earlier versions of this work.



\bibliography{SM2_all}

\begin{thebibliography}{31}
\providecommand{\natexlab}[1]{#1}
\providecommand{\url}[1]{\texttt{#1}}
\providecommand{\urlprefix}{URL }
\expandafter\ifx\csname urlstyle\endcsname\relax
  \providecommand{\doi}[1]{doi:\discretionary{}{}{}#1}\else
  \providecommand{\doi}{doi:\discretionary{}{}{}\begingroup
  \urlstyle{rm}\Url}\fi

\bibitem[{Anschel, Baram, and Shimkin(2017)}]{Ans}
Anschel, O.; Baram, N.; and Shimkin, N. 2017.
\newblock Averaged-DQN: Variance Reduction and Stabilization for Deep
  Reinforcement Learning.
\newblock In \emph{Proceedings of the 34th International Conference on Machine
  Learning, {ICML}}, volume~70, 176--185.

\bibitem[{Asadi and Littman(2017)}]{Asa}
Asadi, K.; and Littman, M.~L. 2017.
\newblock An Alternative Softmax Operator for Reinforcement Learning.
\newblock In \emph{Proceedings of the 34th International Conference on Machine
  Learning, {ICML}}, volume~70, 243--252.

\bibitem[{Beliakov, Sola, and Calvo(2016)}]{Bel}
Beliakov, G.; Sola, H.~B.; and Calvo, T. 2016.
\newblock \emph{A Practical Guide to Averaging Functions}, volume 329.
\newblock Springer.

\bibitem[{Bellman(1958)}]{Ric2}
Bellman, R. 1958.
\newblock Dynamic Programming and Stochastic Control Processes.
\newblock \emph{Inf. Control.} 1(3): 228--239.

\bibitem[{Bellman and Dreyfus(1959)}]{Ric}
Bellman, R.; and Dreyfus, S. 1959.
\newblock Functional approximations and dynamic programming.
\newblock \emph{Mathematics of Computation} 13(68): 247--247.

\bibitem[{Dai et~al.(2018)Dai, Shaw, Li, Xiao, He, Liu, Chen, and Song}]{BoD}
Dai, B.; Shaw, A.; Li, L.; Xiao, L.; He, N.; Liu, Z.; Chen, J.; and Song, L.
  2018.
\newblock {SBEED:} Convergent Reinforcement Learning with Nonlinear Function
  Approximation.
\newblock In \emph{Proceedings of the 35th International Conference on Machine
  Learning, {ICML} 2018, Stockholmsm{\"{a}}ssan, Stockholm, Sweden, July 10-15,
  2018}, volume~80 of \emph{Proceedings of Machine Learning Research},
  1133--1142. {PMLR}.

\bibitem[{Dhariwal et~al.(2017)Dhariwal, Hesse, Klimov, Nichol, Plappert,
  Radford, Schulman, Sidor, Wu, and Zhokhov}]{baselines}
Dhariwal, P.; Hesse, C.; Klimov, O.; Nichol, A.; Plappert, M.; Radford, A.;
  Schulman, J.; Sidor, S.; Wu, Y.; and Zhokhov, P. 2017.
\newblock OpenAI Baselines.
\newblock \url{https://github.com/openai/baselines}.

\bibitem[{Feinberg(1996)}]{Mar}
Feinberg, A. 1996.
\newblock Markov Decision Processes: Discrete Stochastic Dynamic Programming
  (Martin L. Puterman).
\newblock \emph{{SIAM}} 38(4): 689.

\bibitem[{Fox, Pakman, and Tishby(2016)}]{Fox}
Fox, R.; Pakman, A.; and Tishby, N. 2016.
\newblock Taming the Noise in Reinforcement Learning via Soft Updates.
\newblock In \emph{Proceedings of the Thirty-Second Conference on Uncertainty
  in Artificial Intelligence, {USA}}.

\bibitem[{Henderson et~al.(2017)Henderson, Islam, Bachman, Pineau, Precup, and
  Meger}]{henderson2017deep}
Henderson, P.; Islam, R.; Bachman, P.; Pineau, J.; Precup, D.; and Meger, D.
  2017.
\newblock Deep reinforcement learning that matters.
\newblock \emph{arXiv preprint arXiv:1709.06560} .

\bibitem[{Kim et~al.(2019)Kim, Asadi, Littman, and Konidaris}]{Kim}
Kim, S.; Asadi, K.; Littman, M.~L.; and Konidaris, G.~D. 2019.
\newblock DeepMellow: Removing the Need for a Target Network in Deep
  Q-Learning.
\newblock In \emph{Proceedings of the Twenty-Eighth International Joint
  Conference on Artificial Intelligence, {IJCAI}}, 2733--2739.

\bibitem[{Kim and Konidaris(2019)}]{Seung}
Kim, S.; and Konidaris, G. 2019.
\newblock Adaptive Temperature Tuning for Mellowmax in Deep Reinforcement
  Learning .

\bibitem[{Kraemer and Banerjee(2016)}]{Kra}
Kraemer, L.; and Banerjee, B. 2016.
\newblock Multi-agent reinforcement learning as a rehearsal for decentralized
  planning.
\newblock \emph{Neurocomputing} 190: 82--94.

\bibitem[{Lee, Choi, and Oh(2017)}]{Kyun}
Lee, K.; Choi, S.; and Oh, S. 2017.
\newblock Sparse Markov Decision Processes with Causal Sparse Tsallis Entropy
  Regularization for Reinforcement Learning.
\newblock \emph{CoRR} abs/1709.06293.

\bibitem[{Lillicrap et~al.(2016)Lillicrap, Hunt, Pritzel, Heess, Erez, Tassa,
  Silver, and Wierstra}]{Lil}
Lillicrap, T.~P.; Hunt, J.~J.; Pritzel, A.; Heess, N.; Erez, T.; Tassa, Y.;
  Silver, D.; and Wierstra, D. 2016.
\newblock Continuous control with deep reinforcement learning.
\newblock In \emph{4th International Conference on Learning Representations,
  {ICLR}}.

\bibitem[{Mei et~al.(2019)Mei, Xiao, Huang, Schuurmans, and M{\"{u}}ller}]{Mei}
Mei, J.; Xiao, C.; Huang, R.; Schuurmans, D.; and M{\"{u}}ller, M. 2019.
\newblock On Principled Entropy Exploration in Policy Optimization.
\newblock In \emph{Proceedings of the Twenty-Eighth International Joint
  Conference on Artificial Intelligence, {IJCAI}}, 3130--3136.

\bibitem[{Mnih et~al.(2015)Mnih, Kavukcuoglu, Silver, Rusu, Veness, Bellemare,
  Graves, Riedmiller, Fidjeland, Ostrovski, Petersen, Beattie, Sadik,
  Antonoglou, King, Kumaran, Wierstra, Legg, and Hassabis}]{Mni}
Mnih, V.; Kavukcuoglu, K.; Silver, D.; Rusu, A.~A.; Veness, J.; Bellemare,
  M.~G.; Graves, A.; Riedmiller, M.~A.; Fidjeland, A.; Ostrovski, G.; Petersen,
  S.; Beattie, C.; Sadik, A.; Antonoglou, I.; King, H.; Kumaran, D.; Wierstra,
  D.; Legg, S.; and Hassabis, D. 2015.
\newblock Human-level control through deep reinforcement learning.
\newblock \emph{Nature} 518(7540): 529--533.

\bibitem[{Oliehoek, Spaan, and Vlassis(2011)}]{Oli}
Oliehoek, F.~A.; Spaan, M. T.~J.; and Vlassis, N.~A. 2011.
\newblock Optimal and Approximate Q-value Functions for Decentralized POMDPs
  \urlprefix\url{http://arxiv.org/abs/1111.0062}.

\bibitem[{Pan et~al.(2020)Pan, Cai, Meng, Chen, and Huang}]{LinP}
Pan, L.; Cai, Q.; Meng, Q.; Chen, W.; and Huang, L. 2020.
\newblock Reinforcement Learning with Dynamic Boltzmann Softmax Updates
  1992--1998.

\bibitem[{Rashid et~al.(2018)Rashid, Samvelyan, de~Witt, Farquhar, Foerster,
  and Whiteson}]{Ras}
Rashid, T.; Samvelyan, M.; de~Witt, C.~S.; Farquhar, G.; Foerster, J.~N.; and
  Whiteson, S. 2018.
\newblock {QMIX:} Monotonic Value Function Factorisation for Deep Multi-Agent
  Reinforcement Learning.
\newblock In \emph{Proceedings of the 35th International Conference on Machine
  Learning, {ICML}}, volume~80, 4292--4301.

\bibitem[{Samvelyan et~al.(2019)Samvelyan, Rashid, de~Witt, Farquhar, Nardelli,
  Rudner, Hung, Torr, Foerster, and Whiteson}]{Sam}
Samvelyan, M.; Rashid, T.; de~Witt, C.~S.; Farquhar, G.; Nardelli, N.; Rudner,
  T. G.~J.; Hung, C.; Torr, P. H.~S.; Foerster, J.~N.; and Whiteson, S. 2019.
\newblock The StarCraft Multi-Agent Challenge.
\newblock In \emph{Proceedings of the 18th International Conference on
  Autonomous Agents and MultiAgent Systems, {AAMAS}}, 2186--2188.

\bibitem[{Song, Parr, and Carin(2019)}]{Song}
Song, Z.; Parr, R.; and Carin, L. 2019.
\newblock Revisiting the Softmax Bellman Operator: New Benefits and New
  Perspective.
\newblock In \emph{Proceedings of the 36th International Conference on Machine
  Learning, {ICML}}, volume~97, 5916--5925.

\bibitem[{Sunehag et~al.(2018)Sunehag, Lever, Gruslys, Czarnecki, Zambaldi,
  Jaderberg, Lanctot, Sonnerat, Leibo, Tuyls, and Graepel}]{Sun}
Sunehag, P.; Lever, G.; Gruslys, A.; Czarnecki, W.~M.; Zambaldi, V.~F.;
  Jaderberg, M.; Lanctot, M.; Sonnerat, N.; Leibo, J.~Z.; Tuyls, K.; and
  Graepel, T. 2018.
\newblock Value-Decomposition Networks For Cooperative Multi-Agent Learning
  Based On Team Reward.
\newblock In \emph{Proceedings of the 17th International Conference on
  Autonomous Agents and MultiAgent Systems, {AAMAS}}, 2085--2087.

\bibitem[{Sutton(1996)}]{sutton1996generalization}
Sutton, R.~S. 1996.
\newblock Generalization in reinforcement learning: Successful examples using
  sparse coarse coding.
\newblock In \emph{Advances in neural information processing systems},
  1038--1044.

\bibitem[{Tasfi(2016)}]{tas}
Tasfi, N. 2016.
\newblock PyGame Learning Environment.
\newblock \url{https://github.com/ntasfi/PyGame-Learning-Environment}.

\bibitem[{Thrun and Schwartz(1993)}]{Thr}
Thrun, S.; and Schwartz, A. 1993.
\newblock Issues in Using Function Approximation for Reinforcement Learning.
\newblock \emph{Proceedings of the 4th Connectionist Models Summer School
  Hillsdale, NJ. Lawrence Erlbaum} 1--9.

\bibitem[{Tsitsiklis and Van~Roy(1997)}]{tsitsiklis1997analysis}
Tsitsiklis, J.~N.; and Van~Roy, B. 1997.
\newblock Analysis of temporal-diffference learning with function
  approximation.
\newblock In \emph{Advances in neural information processing systems},
  1075--1081.

\bibitem[{van Hasselt, Guez, and Silver(2016)}]{Van}
van Hasselt, H.; Guez, A.; and Silver, D. 2016.
\newblock Deep Reinforcement Learning with Double Q-Learning.
\newblock In \emph{Proceedings of the Thirtieth {AAAI} Conference on Artificial
  Intelligence}, 2094--2100.

\bibitem[{Vinyals et~al.(2017)Vinyals, Ewalds, Bartunov, Georgiev, Vezhnevets,
  Yeo, Makhzani, K{\"{u}}ttler, Agapiou, Schrittwieser, Quan, Gaffney,
  Petersen, Simonyan, Schaul, van Hasselt, Silver, Lillicrap, Calderone, Keet,
  Brunasso, Lawrence, Ekermo, Repp, and Tsing}]{SMAC}
Vinyals, O.; Ewalds, T.; Bartunov, S.; Georgiev, P.; Vezhnevets, A.~S.; Yeo,
  M.; Makhzani, A.; K{\"{u}}ttler, H.; Agapiou, J.~P.; Schrittwieser, J.; Quan,
  J.; Gaffney, S.; Petersen, S.; Simonyan, K.; Schaul, T.; van Hasselt, H.;
  Silver, D.; Lillicrap, T.~P.; Calderone, K.; Keet, P.; Brunasso, A.;
  Lawrence, D.; Ekermo, A.; Repp, J.; and Tsing, R. 2017.
\newblock StarCraft {II:} {A} New Challenge for Reinforcement Learning.
\newblock \emph{CoRR} abs/1708.04782.

\bibitem[{Young and Tian(2019)}]{Kyo}
Young, K.; and Tian, T. 2019.
\newblock MinAtar: An Atari-inspired Testbed for More Efficient Reinforcement
  Learning Experiments \urlprefix\url{http://arxiv.org/abs/1903.03176}.

\bibitem[{Zheng et~al.(2018)Zheng, Meng, Hao, and Zhang}]{Zheng}
Zheng, Y.; Meng, Z.; Hao, J.; and Zhang, Z. 2018.
\newblock Weighted Double Deep Multiagent Reinforcement Learning in Stochastic
  Cooperative Environments.
\newblock 421--429. Springer.

\end{thebibliography}

\clearpage

\section*{Appendix}

\subsection*{A.1 Contraction of SM2}
\textbf{Lemma 6.1.} (Song, Parr, and Carin 2019) \label{a-2-1}
Assuming $\forall (s,a)$, $|R(s,a)|\leq R_{max}$ and the initial Q-value function $Q^0(s,a)\in [-R_{max}, R_{max}]$, then $Q^k(s,a) \triangleq\mathcal{T}_{sm}^kQ^0(s,a)\in[-\frac{R_{max}}{1-\gamma},\frac{R_{max}}{1-\gamma}]$ for $k$ iterations.
\begin{proof}
	We use the mathematical induction to prove it.
	Assuming $Q^{k}(s,a)\leq \sum_{i=0}^{k}\gamma^i R_{max}$.\\
	(1) When $k=1$, we can get
	\begin{equation*}
	\begin{split}
	Q^1(s,a)
	&=\mathcal{T}_{sm}Q^0(s,a)\\
	&\leq \mathcal{T}Q^0(s,a)\\
	&\leq R_{max}+\gamma R_{max}
	\end{split}
	\end{equation*}
	(2)  Assuming when $k=j$, it has $Q^j\leq \sum_{i=0}^{j}\gamma^j R_{max}$.
	When $k=j+1$, then
	\begin{equation*}
	\begin{split}
	Q^{j+1}(s,a)
	&= \mathcal{T}_{sm}Q^{j}(s,a)\\
	&\leq \mathcal{T}Q^{j}(s,a)\\
	&\leq \sum_{i=1}^{j+1}\gamma^i R_{max}
	\end{split}
	\end{equation*}
\end{proof}

\textbf{Theorem 3.1}\label{A.2}
(\textbf{Contraction})
Let $Q_1$, $Q_2$ are two different Q-value function. If $\omega>0$ and 
$ -\frac{\omega}{1-e^{-c\omega}}\leq \alpha \leq \frac{\omega}{e^{c\omega}-1} $
where $c=\frac{2R_{max}}{1-\gamma}$. Then
\begin{equation*}\label{a-2-2}
\|\mathcal{T}_{sm}Q_1-\mathcal{T}_{sm}Q_2\|\leq \gamma\|Q_1-Q_2\|
\end{equation*}

\begin{proof}
	We first prove $sm_{\omega}Q$ is a non-expansion.
	Let $\Delta_i=Q_1(s,a_i)-Q_2(s,a_i)$ for $i \in \{1,2,\cdots,n\}$, and $i^*=\arg\max_{i}\Delta_i$.
	Without loss of generality, we assume that $Q_1(s,a_{i^*})-Q_2(s,a_{i^*})\geq0$, then
	\begin{equation*}
	\begin{split}
	&|sm_{\omega}Q_1(s,\cdot)_1-sm_{\omega}Q_2(s,\cdot)|\\
	=& \frac{1}{\omega}|\log\frac{\sum_{i=1}^{n}soft_{\alpha}(Q_1(s,a_i)e^ {\omega Q_1(s,a_i)}}{\sum_{i=1}^{n}soft_{\alpha}(Q_2(s,a_i)e^ {\omega Q_2(s,a_i)}}|\\
	=& \frac{1}{\omega}|\log[\frac{\sum_{i=1}^{n}e^{(\alpha+\omega)Q_1(s,a_{i})}}{\sum_{i=1}^{n}e^{\alpha Q_1(s,a_{i})}}
	\frac{\sum_{i=1}^{n}e^{\alpha Q_2(s,a_{i})}}{\sum_{i=1}^{n}e^{(\alpha+\omega)Q_2(s,a_{i})}}]|\\
	=& \frac{1}{\omega}|\log[\frac{\sum_{i=1}^{n}e^{(\alpha+\omega)(\Delta_i+Q_2(s,a_{i}))}}{\sum_{i=1}^{n}e^{\alpha (\Delta_i +Q_2(s,a_{i}))}}
	\frac{\sum_{i=1}^{n}e^{\alpha Q_2(s,a_{i})}}{\sum_{i=1}^{n}e^{(\alpha+\omega)Q_2(s,a_{i})}}]|
	\end{split}
	\end{equation*}
	Let $F(\Delta)=\frac{\sum_{i=1}^{n}e^{(\alpha+\omega)(\Delta_i+Q_2(s,a_{i}))}}{\sum_{i=1}^{n}e^{\alpha (\Delta_i +Q_2(s,a_{i}))}}$, then we will prove that $F(\Delta)$ is monotonic function about $\Delta_i$, $i\in \{1,2,$ $\cdots,n\}$.
	\begin{equation*}
	\begin{split}
	&\frac{\partial F(\Delta)}{\partial \Delta_i}\\
	= &\sum_{j=1}^{n}\!\frac{\alpha e^{(\alpha\!+\!\omega) (\Delta_j+Q_2(s,a_j))}
		[(1\!+\!\frac{\omega}{\alpha})e^{\omega (Q_1(s,a_i)-Q_1(s,a_j))}\!-\!1]}
	{(\sum_{i=1}^{n}e^{\alpha (\Delta_i +Q_2(s,a_{i}))})^2}
	\end{split}
	\end{equation*}
	We apply Lemma 6.1, then $|Q(s,a_i)-Q(s,a_j)|\leq \frac{2R_{max}}{1-\gamma}$.
	Using $\omega>0$ and $ -\frac{\omega}{1-e^{-c\omega}} \leq \alpha \leq \frac{\omega}{e^{c\omega}-1}$
	where $c=\frac{2R_{max}}{1-\gamma}$, then $\frac{\partial F(\Delta)}{\partial \Delta_i}\geq 0$.
	We have
	\begin{equation*}
	\begin{split}
	&|sm_{\omega}Q_1(s,\cdot)_1-sm_{\omega}Q_2(s,\cdot)|\\
	=& \frac{1}{\omega}|\log[F(\Delta)
	\frac{\sum_{i=1}^{n}e^{\alpha Q_2(s,a_{i})}}{\sum_{i=1}^{n}e^{(\alpha+\omega)Q_2(s,a_{i})}}]|\\
	\leq & \frac{1}{\omega}|\log[F(\Delta^*)
	\frac{\sum_{i=1}^{n}e^{\alpha Q_2(s,a_{i})}}{\sum_{i=1}^{n}e^{(\alpha+\omega)Q_2(s,a_{i})}}]|\\
	=& |\Delta_{i^*}|\\
	=& \max_{i}|Q_1(s,a_i)-Q_2(s,a_i)|
	\end{split}
	\end{equation*}
	Therefor, we can get
	\begin{equation*}
	\|\mathcal{T}_{sm}Q_{1}-\mathcal{T}_{sm}Q_2\|\leq \gamma\|sm_{\omega}Q_1-sm_{\omega}Q_2\|\leq \gamma\|Q_1-Q_2\|
	\end{equation*}
\end{proof}



\subsection*{A.2 Performance Bound}

\textbf{Lemma 6.2} \label{lemma6_2}
For $ \forall\ x\geq 0 $, defined the function $$ f(x)=\frac{e^{\omega x}}{e^{(\omega+\alpha)x}+1}, $$
then, we have
\begin{equation*}
\begin{split}
\max_{x}f(x)=
\left\{
\begin{array}{lll}
\frac{\omega}{\alpha+\omega}, && {\rm if}\ \alpha-\omega<0\\
&&\\
\frac{1}{2}, && {\rm if}\ \alpha-\omega\geq0
\end{array}
\right.
\end{split}
\end{equation*}

\begin{proof}
	The derivative of $ f(x) $ is 
	$$ f'(x)= e^{\omega x}\frac{\omega-\alpha e^{(\omega+\alpha)x}}{(e^{(\omega+\alpha)x}+1)^2}$$
	When $ \alpha-\omega<0 $, the zero point of $ f'(x) $ is $ x^{\star}=\frac{1}{\omega+\alpha}\log\frac{\omega}{\alpha} $.
	Then,
	\begin{equation*}
	\begin{aligned}
	&\max_{x}f(x)=f(x^{\star})\\
	=&\frac{(\frac{\omega}{\alpha})^{\frac{\omega}{\omega+\alpha}}}{\frac{\omega}{\alpha}+1}\\
	\leq& \frac{\frac{\omega}{\alpha}}{\frac{\omega}{\alpha}+1}\\
	=& \frac{\omega}{\alpha+\omega}
	\end{aligned}
	\end{equation*} 
	
	When $ \alpha-\omega\geq0 $, we have $ f'(x)\leq0 $. Then
	$$ \max_{x}f(x)=f(0)= \frac{1}{2}$$
\end{proof}

\textbf{Lemma 6.3} \label{a-3-1}
For arbitrary Q-value function $Q(s,a)$, defined $\xi=\sup_{Q}(\max_{a}Q-sm_{\omega}Q)$, then\\
\begin{equation*}
\begin{split}
0< \xi
\leq
\left\{
\begin{array}{lll}
\frac{1}{\omega}\log\left(n-\frac{\alpha(n-1)}{\alpha+\omega}\right)
, && {\rm if}\ \alpha-\omega<0\\
&&\\
\frac{1}{\omega}\log(\frac{1+n}{2}), && {\rm if}\ \alpha-\omega\geq0
\end{array}
\right.
\end{split}
\end{equation*}

\begin{proof}
	We prove this Lemma in two steps.
	Without loss of generality, we assume that $Q(s,a_1)\geq Q(s,a_2)\geq\cdots\geq Q(s,a_n)$. Let $\Delta_i\triangleq Q(s,a_i)-Q(s,a_1)$, then $\Delta_{i}\leq0$, $i\in \{1,2,\cdots,n\}$.\\
	(1) Upper bound:
	\begin{equation*}
	\begin{split}
	&\max_a Q(s,a)-sm_{\omega}Q(s,\cdot)\\
	=& \frac{1}{\omega}\log[\frac{e^{\omega Q(s,a_1)} \sum_{i=1}^{n}e^{\alpha Q(s,a_i)}}{\sum_{i=1}^{n}e^{(\alpha+\omega)Q(s,a_i)}}]\\
	=& \frac{1}{\omega}\log[\frac{1+\sum_{i=2}^{n}e^{\alpha \Delta_i}}{1+\sum_{i=2}^{n}e^{(\alpha+\omega)\Delta_i}}]\\
	\leq&  \frac{1}{\omega}\log[1+\frac{\sum_{i=2}^{n}e^{\alpha \Delta_i}}{1+\sum_{i=2}^{n}e^{(\alpha+\omega)\Delta_i}}]\\
	\leq& \frac{1}{\omega}\log[1+\sum_{i=2}^{n}\frac{e^{\alpha \Delta_i}}{1+e^{(\alpha+\omega)\Delta_i}}]\\
	=& \frac{1}{\omega}\log[1+\sum_{i=2}^{n}\frac{e^{\omega(-\Delta_i)}}{1+e^{(\omega+\alpha) (-\Delta_i)}}]\\
	\leq&
	\left\{
	\begin{array}{lll}
	\frac{1}{\omega}\log\left(n-\frac{\alpha(n-1)}{\alpha+\omega}\right),
	&&     {\rm if}\ \alpha-\omega<0\\
	&&\\
	\frac{1}{\omega}\log(\frac{1+n}{2}),&& {\rm if}\ \alpha-\omega\geq0
	\end{array}
	\right.
	\\
	\end{split}
	\end{equation*}
	The last inequality, it is easy to get it accroding to the above Lamma 6.2 and $ \log(1+x)\leq x, \ (x\geq0) $.\\
	(2) Lower bound:
	\begin{equation*}
	\begin{split}
	&sm_{\omega}Q(s,\cdot)\\
	=&\frac{1}{\omega}\log[\sum_{i=1}^{n}soft_{\alpha}Q(s,a_i)e^{\omega Q(s,a_i)}]\\
	<&\frac{1}{\omega}\log[\sum_{i=1}^{n}soft_{\alpha}Q(s,a_i)e^{\omega Q(s,a_1)}]\\
	=&Q(s,a_1)
	\end{split}
	\end{equation*}
	Note that if the inequality can be equal, so $sm_{\omega}Q(s,\cdot)=\max_{a}Q(s,a)$
\end{proof}

\textbf{Corollary 6.4} \label{corollary}
For the Mellowmax operator $ mm_{\omega}Q $, defined $ \xi'=\sup_{Q}(\max_{a}Q-mm_{\omega}Q) $, we have
\begin{equation*}
0<\xi'\leq \frac{1}{\omega}\log(n)
\end{equation*}
\begin{proof}
	When $ \alpha=0 $, the $ sm_{\omega}Q $ has became the $ mm_{\omega}Q $.
\end{proof}

\textbf{Theorem 3.2}\label{A.3}
Let the optimal Q-value function $Q^*$ is a fixed point during Q-iteration with $\mathcal{T}$, i.e. $\mathcal{T}Q^*(s,a)=Q^*(s,a)$, and $Q^k(s,a)\triangleq \mathcal{T}_{sm}^kQ^0(s,a)$ for arbitrary initial function $Q^0(s,a)$ during $k$ Q-iteration.
$\forall (s,a)$
\\
\\
(\uppercase\expandafter{\romannumeral1}) If $\alpha\geq\omega$, then
\begin{align*}
\limsup\limits_{k\rightarrow\infty}\|Q^*(s,a)-Q^k(s,a)\|\leq
\frac{\gamma}{\omega(1-\gamma)} \log(\frac{1+n}{2}) \quad
\end{align*}
(\uppercase\expandafter{\romannumeral2}) If $\alpha<\omega$, then
\begin{align*}
\!\limsup\limits_{k\rightarrow\infty}\!\|Q^*(s,a)\!-\!Q^k(s,a)\|\!
\!\leq\! \frac{\gamma}{\omega(1\!-\!\gamma)}\!\log(n\!-\!\frac{\alpha(n\!-\!1)}{\alpha\!+\!\omega})
\end{align*}



\begin{proof}
	Defined $\xi=\sup_{Q}\max_{a}Q-sm_{\omega}Q$ , which is the supremum of the different between the max and soft mellowmax for each agent.
	For simplicity, we omit $(s,a)$, then
	\begin{equation*}
	\begin{split}
	&\limsup\limits_{k\rightarrow\infty}\|Q^*(s,a)-Q^k(s,a)\|\\
	=&\limsup\limits_{k\rightarrow\infty}\|\mathcal{T}^{k+1}Q^0-\mathcal{T}_{sm}^{k+1}Q^0\|\\
	=& \limsup\limits_{k\rightarrow\infty}\|\mathcal{T}\mathcal{T}^{k}Q^0 \!-\! \mathcal{T}\mathcal{T}_{sm}^kQ^0 \!+\! \mathcal{T}\mathcal{T}_{sm}^kQ^0 \!-\! \mathcal{T}_{sm}\mathcal{T}_{sm}^{k}Q^0\|\\
	\leq&
	\gamma
	\limsup\limits_{k\rightarrow\infty}\|(\mathcal{T}^{k}Q^0-\mathcal{T}_{sm}^{k}Q^0)\|+
	\gamma \xi\\
	\leq&
	\gamma
	\limsup\limits_{k\rightarrow\infty}\|\max_{s',a'}(\mathcal{T}^{k}Q^0-\mathcal{T}_{sm}^{k}Q^0)\|+\gamma \xi\\
	&\cdots\\
	\leq & \gamma \xi+\gamma^2 \xi+\cdots+\gamma^{k+1} \xi\\
	\leq& \frac{\gamma \xi}{1-\gamma}
	\end{split}
	\end{equation*}
	Applying Lemma 6.3 is easy to get it.
	The other side, using induction can get it.
\end{proof}

\textbf{Corollary 3.1}
For the Mellowmax Bellman operator $ \mathcal{T}_{m}$, defined $ Q^{k}(s,a)\triangleq \mathcal{T}_{m}^k Q^0(s,a)$, we have
\begin{equation*}
\lim\limits_{k \rightarrow \infty}\|Q^{\star}(s,a)-Q^k(s,a)\|\leq \frac{\gamma}{\omega(1-\gamma)}\log(n)
\end{equation*}

\subsection*{A.3 Alleviate overestimation}

\textbf{Theorem 3.3}\label{A.31}
Assume all the true optimal action value are equal at $Q^*(s, a)=V^*(s)$ and the estimation errors $Z^{s,a}=Q(s, a)-Q^*(s, a)$ are independently distributed uniformly random in $[-\epsilon, \epsilon]$, defined the overestimation error
$\varTheta\triangleq\mathbb{E}[\max_{a}Q(s,a)-Q^*(s,a)]$,
and
$\varTheta_{sm}\triangleq\mathbb{E}[sm_{\omega}Q(s,a)-Q^*(s,a)]$,
\\
\\
(\uppercase\expandafter{\romannumeral1}) If $\alpha\geq\omega$, then
\begin{equation*}
\varTheta-\varTheta_{sm}\in (0,\frac{1}{\omega} \log(\frac{1+n}{2})] \quad \quad \quad \
\end{equation*}
(\uppercase\expandafter{\romannumeral2}) If $\alpha<\omega$, then
\begin{equation*}
\varTheta-\varTheta_{sm}\in (0,\frac{1}{\omega}\log(n-\frac{\alpha(n-1)}{\alpha+\omega})] 
\end{equation*}
(\uppercase\expandafter{\romannumeral3}) The overestimation error for $\mathcal{T}_{sm}$ is increasing monotonically w.r.t. $\alpha\in[0,\infty)$ and $\omega\in(0,\infty)$.

\begin{proof}
	(\uppercase\expandafter{\romannumeral1} and \uppercase\expandafter{\romannumeral2})
	\begin{equation*}
	\begin{split}
	&\varTheta-\varTheta_{sm}\\
	=& \mathbb{E}[\max_{a} Q(s,a)-sm_{\omega}Q(s,a)]\\
	\leq&
	\left\{
	\begin{array}{lll}
	\frac{1}{\omega}\log(n-\frac{\alpha(n-1)}{\alpha+\omega}),
	&&     {\rm if}\ \alpha-\omega<0\\
	&&\\
	\frac{1}{\omega}\log(\frac{1+n}{2}),&& {\rm if}\ \alpha-\omega\geq0
	\end{array}
	\right.\\
	\end{split}
	\end{equation*}
	Applying Lemma 6.3 is easy to get it.\\
	Similarly, we can get that $\varTheta-\varTheta_{sm}>0$.
	\\
	\\
	(\uppercase\expandafter{\romannumeral3})
	Below, we prove that $sm_{\omega}Q$ is a monotonically increasing for $\alpha$ and $\omega$. Defined $Q_i\triangleq Q(s',a_{i}')$.\\
	Let $\alpha_1>\alpha_2\geq0$, and
	$f(\alpha)\triangleq\sum_{i=1}^{n}soft_{\alpha}(Q_i)e^{\omega Q_i}$.
	Duo to the monotonicity of $\log(\cdot)$ function. We want to show that $f(\alpha_1)\geq f(\alpha_2)$.
	\begin{equation*}
	\begin{split}
	\frac{\mathrm{d} f(\alpha) }{\mathrm{d} \alpha}= \sum_{i,j}^{n}
	\frac{e^{\alpha(Q_i+Q_j)}e^{\omega Q_i}}
	{\sum_{i,j}^{n}e^{\alpha(Q_i+Q_j)}}(Q_i-Q_j)
	\end{split}
	\end{equation*}
	We just need to prove $\frac{\mathrm{d} f(\alpha) }{\mathrm{d} \alpha}\geq 0$.\\
	Let $\Delta(i,j)=e^{\alpha(Q_i+Q_j)}e^{\omega Q_i}(Q_i-Q_j)$, so $\frac{\mathrm{d} f(\alpha) }{\mathrm{d} \alpha} = \frac{1}{\sum_{i,j}^{n}e^{\alpha(Q_i+Q_j)}}\sum_{i,j}^{n}\Delta(i,j)$.\\
	When $i=j$, 
	\begin{equation*}
	\Delta(i,i)=e^{\alpha(Q_i+Q_j)}e^{\omega Q_i}(Q_i-Q_i)=0
	\end{equation*}
	When $i\neq j$,
	\begin{equation*}
	\Delta(i,j) \!+\! \Delta(j,i)\!=\!e^{\alpha(Q_i+Q_j)}(e^{\omega Q_i}\!-\!e^{\omega Q_j})(Q_i\!-\!Q_j)\!\geq\! 0
	\end{equation*}
	Therefore,
	\begin{equation*}
	\begin{split}
	\frac{\mathrm{d} f(\alpha) }{\mathrm{d} \alpha}
	=& \frac{1}{\sum_{i,j}^{n}e^{\alpha(Q_i+Q_j)}}\sum_{i,j}^{n}\Delta(i,j)\\
	=&\frac{1}{\sum_{i,j}^{n}e^{\alpha(Q_i+Q_j)}}
	\frac{1}{2}\sum_{i,j}^{n}(\Delta(i,j)+\Delta(j,i))\geq 0.
	\end{split}
	\end{equation*}
	Let $\omega_1>\omega_2>0$, we want to show that $sm_{\omega_1}Q\geq sm_{\omega_2}Q$.
	\begin{equation*}
	\begin{split}
	sm_{\omega_1}Q(s,\cdot)
	&=\frac{1}{\omega_1}\log[\sum_{i=1}^{n}
	soft_{\alpha}(Q_i)e^{\omega_1 Q_i}]\\
	&=\frac{1}{\omega_1}\log[\sum_{i=1}^{n}
	soft_{\alpha}(Q_i)e^{(\omega_2 Q_i)\frac{\omega_1}{\omega_2}}]\\
	&\geq\frac{1}{\omega_1}\log[\sum_{i=1}^{n}
	soft_{\alpha}(Q_i)e^{(\omega_2 Q_i)}]^{\frac{\omega_1}{\omega_2}}\\
	&=sm_{\omega_2}Q(s,\cdot)
	\end{split}
	\end{equation*}
	where the inequality utilize the convexity of the function $f(x)=x^{\frac{\omega_1}{\omega_2}}$.
\end{proof}

\begin{cor}
	Under the above assumption, defined $ \varTheta_{m}\triangleq\mathbb{E}[mm_{\omega}Q(s,a)-Q^*(s,a)] $, we have
	\begin{equation*}
	\varTheta-\varTheta_{m}\in (0, \frac{1}{\omega}\log(n)]
	\end{equation*}
\end{cor}

\subsection*{A.4 Overestimation of MARL}

\textbf{Theorem 4.1}\label{A.1-1}
Under the assumptions of Theorem 3.3,
defined $\varTheta^{1}\triangleq\mathbb{E}[Q_{tot}(\max_{\textbf{a}}\textbf{Q})(s,\textbf{a})-Q_{tot}(\textbf{Q})^*(s,\textbf{a})]$, then
\begin{equation*}\label{b-101}
\begin{split}
\varTheta^{1}
\in [\epsilon l N \frac{n-1}{n+1}, \epsilon L N \frac{n-1}{n+1}]
\end{split}
\end{equation*}
\begin{proof}
	According to the definition of $Z_{i}^{s,a_{i}}$, due to the independent of error, we can derive
	\begin{align*}
	&P(\max_{a_i}Z_{i}^{s,a_{i}}<x)\\
	=& P(Z_{i}^{s,a_{i}^1}<x \wedge Z_{i}^{s,a_{i}^2}<x \wedge \cdots
	\wedge Z_{i}^{s,a_{i}^n}<x ) \\
	=& \prod _{j=1}^{n}P(Z_{i}^{s,a_{i}^j}<x)
	\end{align*}
	By cumulative distribution function (CDF) of $Z_{i}^{s,a_{i}}$, then
	\begin{equation*}
	\begin{split}
	P(\max_{a_i}Z_{i}^{s,a_{i}}<x)
	& = \prod _{j=1}^{n}P(Z_{i}^{s,a_{i}^j}<x)\\
	&= \left\{
	\begin{array}{lll}
	0, &&     {\rm if}\ x\leq -\epsilon\\
	(\frac{\epsilon+x}{2\epsilon})^n, && {\rm if}\ x\in (-\epsilon,\epsilon)\\
	1,&& {\rm if}\ x\geq \epsilon
	\end{array}
	\right.
	\end{split}
	\end{equation*}
	Then
	\begin{equation*}
	\begin{split}
	\varTheta^{1}
	&=\mathbb{E}[\max_{\textbf{a}}Q_{tot}(\textbf{Q})(s,\textbf{a})-Q_{tot}(\textbf{Q})^*(s,\textbf{a})] \\
	&=\mathbb{E}[\max_{\textbf{a}}Q_{tot}(Q_1,\cdots,Q_N)(s,\textbf{a})
	-Q_{tot}(Q_{1}^*,\\
	&\quad\quad \cdots,Q_{N}^*)(s,\textbf{a})] \\
	& \leq \mathbb{E}[\sum_{i=1}^{N}L(\max_{a_{i}}Q_{i}(s,a_i)-Q_{i}^*(s,a_i))]\\
	& = LN\mathbb{E}[\max_{a_{i}}Q_{i}(s,a_i)-Q_{i}^*(s,a_i))]\\
	& = LN \mathbb{E}(\max_{a_{i}} Z_{i}^{s,a_{i}})\\
	& = \epsilon LN \frac{n-1}{n+1}
	\end{split}
	\end{equation*}
	where the inequality follows by Assumption 4.1.\\
	Similarly, we can get that $\mathbb{E}(\max_{\textbf{a}}Z^{\textbf{a}})\geq \epsilon lN \frac{n-1}{n+1}$.
\end{proof}

\textbf{Theorem 4.2}\label{A.4}
Under the assumptions of Theorem 3.3. Defined $\varTheta_{sm}^{1}\triangleq\mathbb{E}[Q_{tot}(sm_{\omega}\textbf{Q})(s,\textbf{a})-Q_{tot}(\textbf{Q})^*(s,\textbf{a})]$,
\\
\\
(\uppercase\expandafter{\romannumeral1}) If $\alpha\geq\omega$, then
\begin{equation*}
\varTheta^{1}-\varTheta_{sm}^{1}\in (0,\frac{1}{\omega}L N \log(\frac{1+n}{2})]
\end{equation*}
(\uppercase\expandafter{\romannumeral2}) If $\alpha<\omega$, then
\begin{equation*}
\varTheta^{1}-\varTheta_{sm}^{1}\in (0,\frac{1}{\omega}LN\log(n-\frac{\alpha(n-1)}{\alpha+\omega})] \quad \quad \quad
\end{equation*}

\begin{proof}
	(\uppercase\expandafter{\romannumeral1} and \uppercase\expandafter{\romannumeral2})
	
	\begin{equation*}
	\begin{split}
	&\varTheta^{1}-\varTheta_{sm}^{1}\\
	=& \mathbb{E}[\max_{\textbf{a}}Q_{tot}(\textbf{Q})(s,\textbf{a})-Q_{tot}(sm_{\omega}\textbf{Q})(s,\textbf{a})]\\
	\leq& \mathbb{E}\{L\sum_{i=1}^{N}[\max_{a_{i}}Q_{i}(s,a_{i})-sm_{\omega}Q_i(s,\cdot)]\}\\
	\leq&
	\left\{
	\begin{array}{lll}
	\frac{1}{\omega}LN \log(n-\frac{\alpha(n-1)}{\alpha+\omega}),
	&&     {\rm if}\ \alpha-\omega<0\\
	&&\\
	\frac{1}{\omega}LN\log(\frac{1+n}{2}),&& {\rm if}\ \alpha-\omega\geq0
	\end{array}
	\right.\\
	\end{split}
	\end{equation*}
	Applying Assumption 4.1 and Lemma 6.3 is easy to get it.\\
	Similarly, we can get that $\varTheta^{1}-\varTheta_{sm}^{1}>0$.
\end{proof}

\begin{figure*}[t]
	\centering
	\subfloat[3m]{\includegraphics[width=0.33\textwidth]{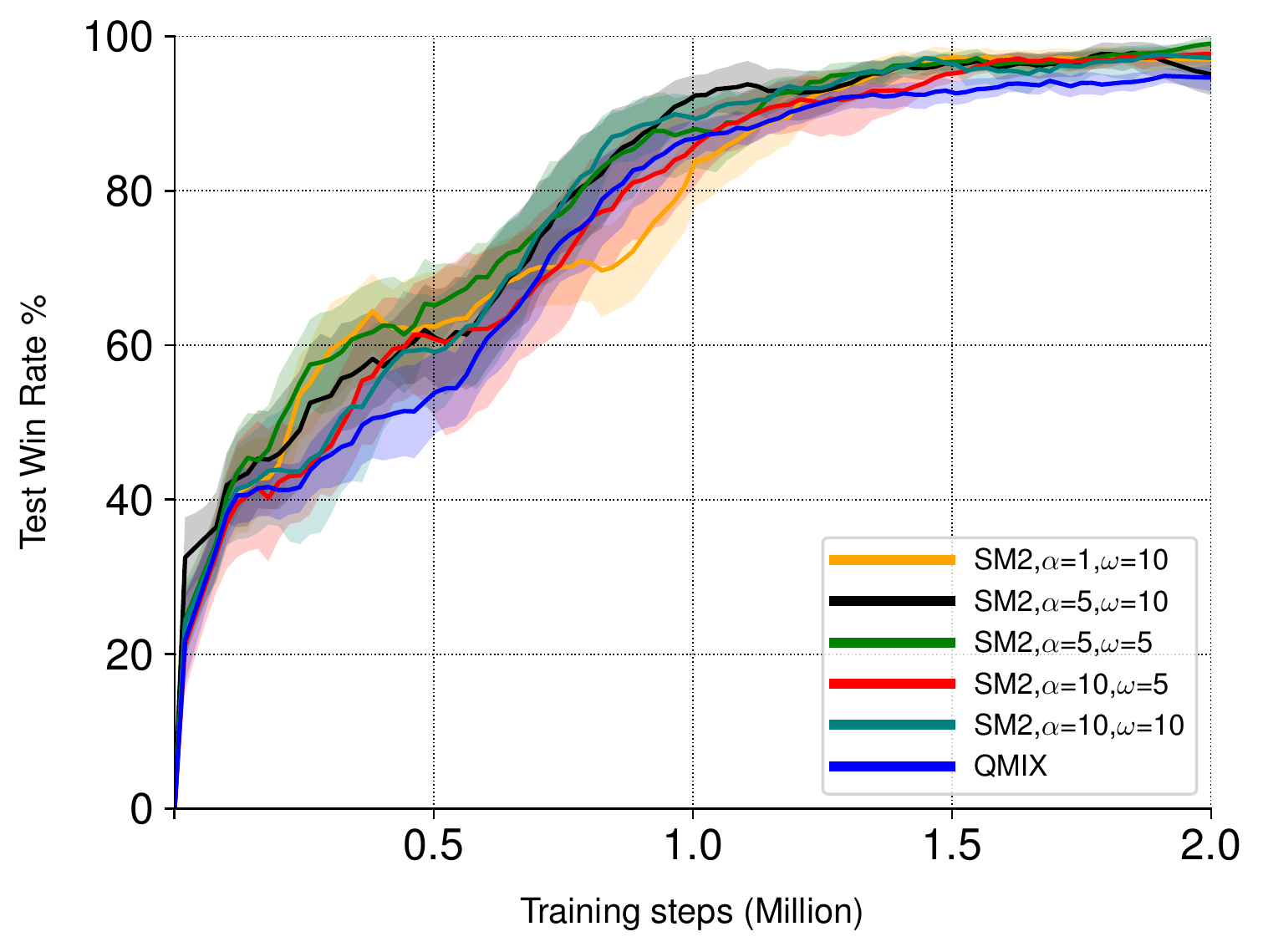}}
	\subfloat[8m]{\includegraphics[width=0.33\textwidth]{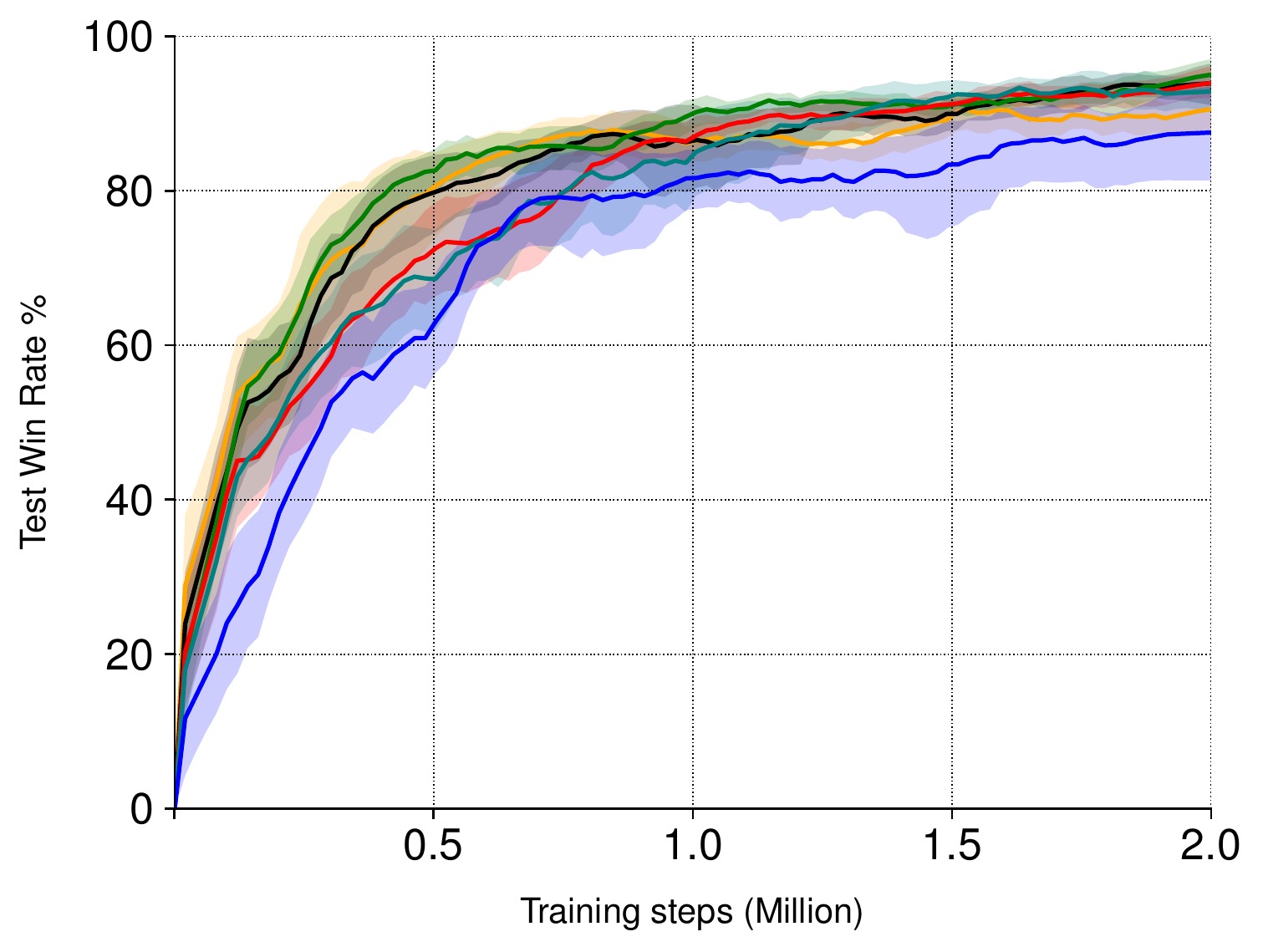}}
	\subfloat[3s5z]{\includegraphics[width=0.33\textwidth]{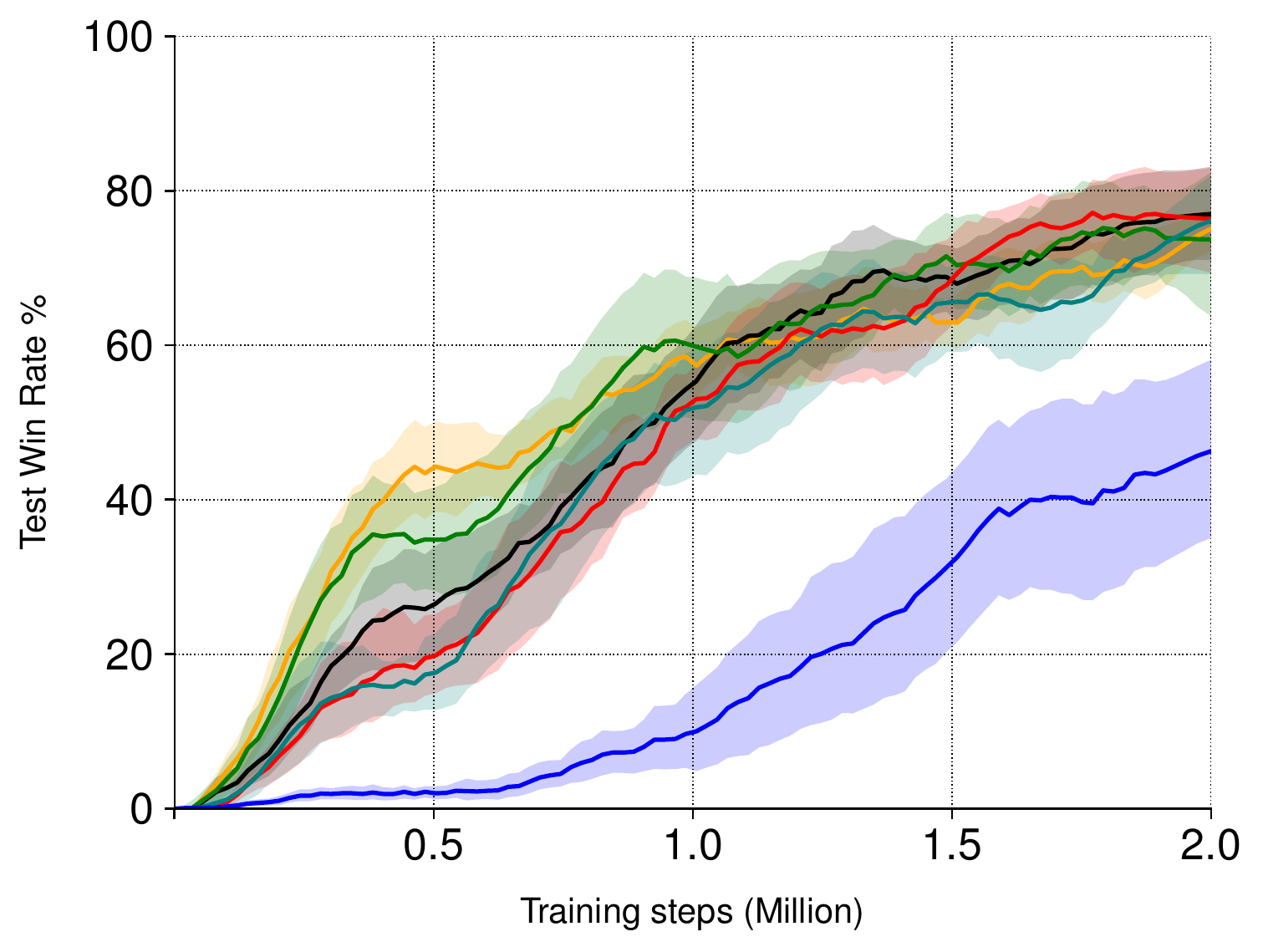}}\\
	\subfloat[2s\_vs\_1sc]{\includegraphics[width=0.33\textwidth]{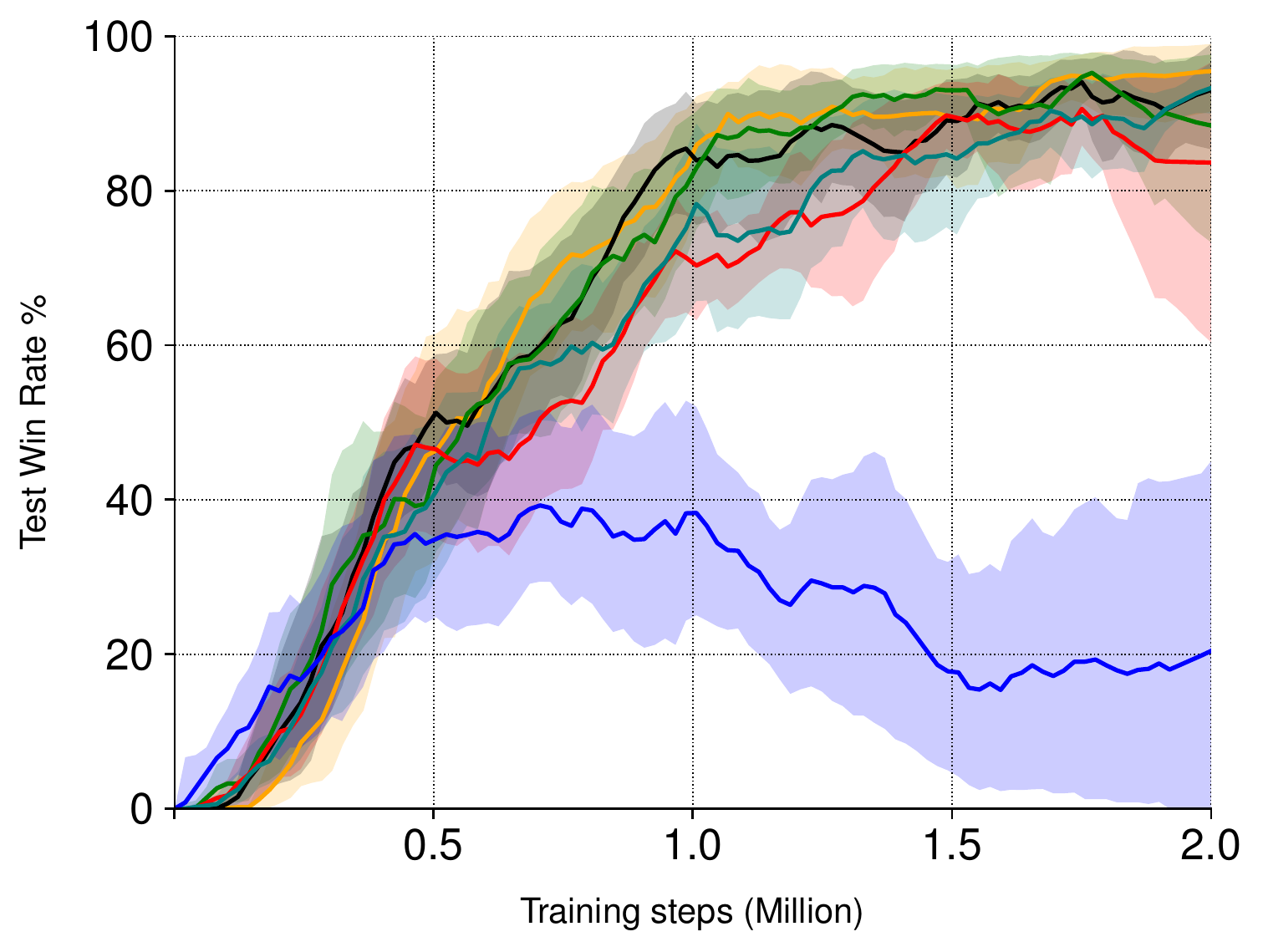}}
	\subfloat[3s6z]{\includegraphics[width=0.33\textwidth]{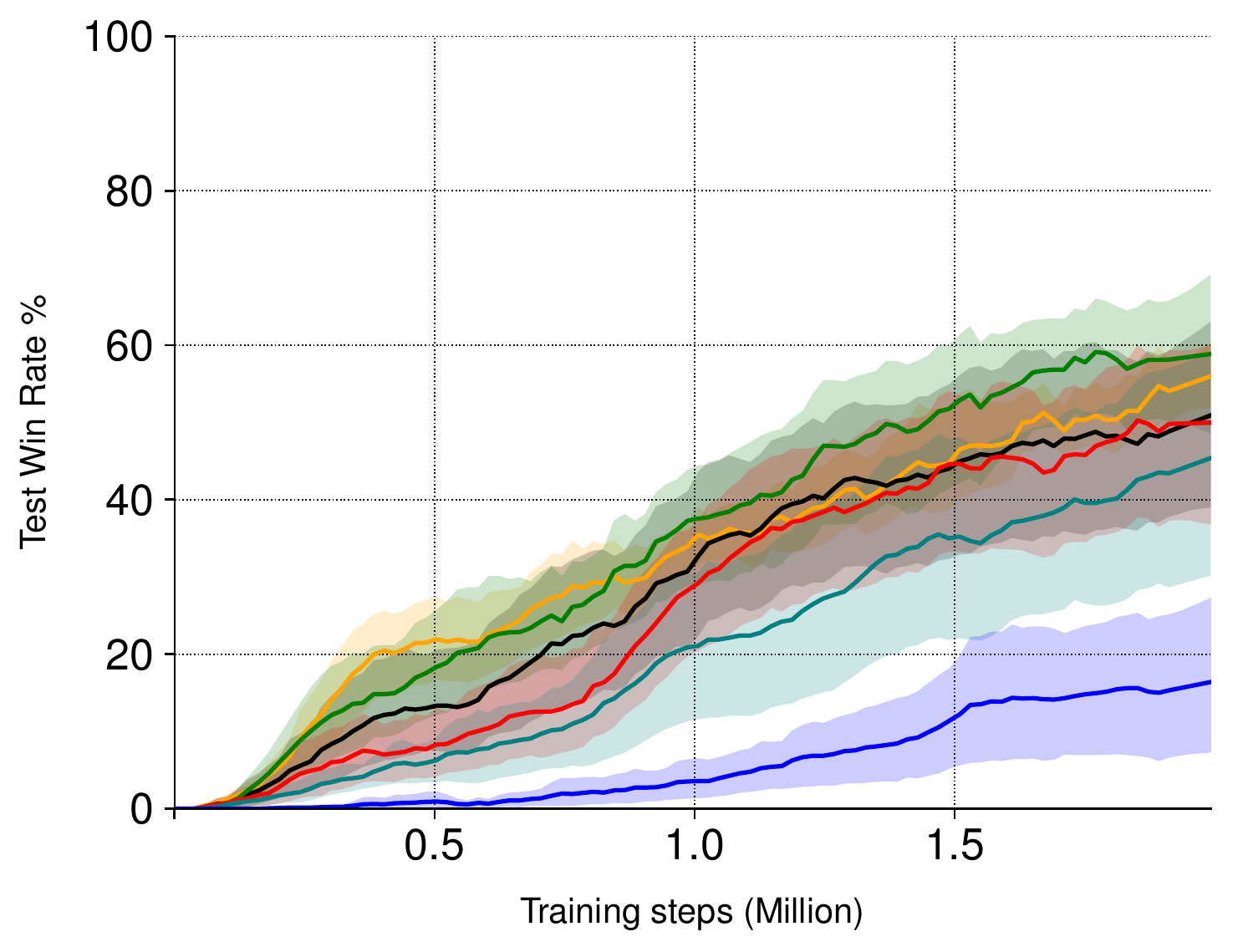}}
	\subfloat[1c3s5z]{\includegraphics[width=0.33\textwidth]{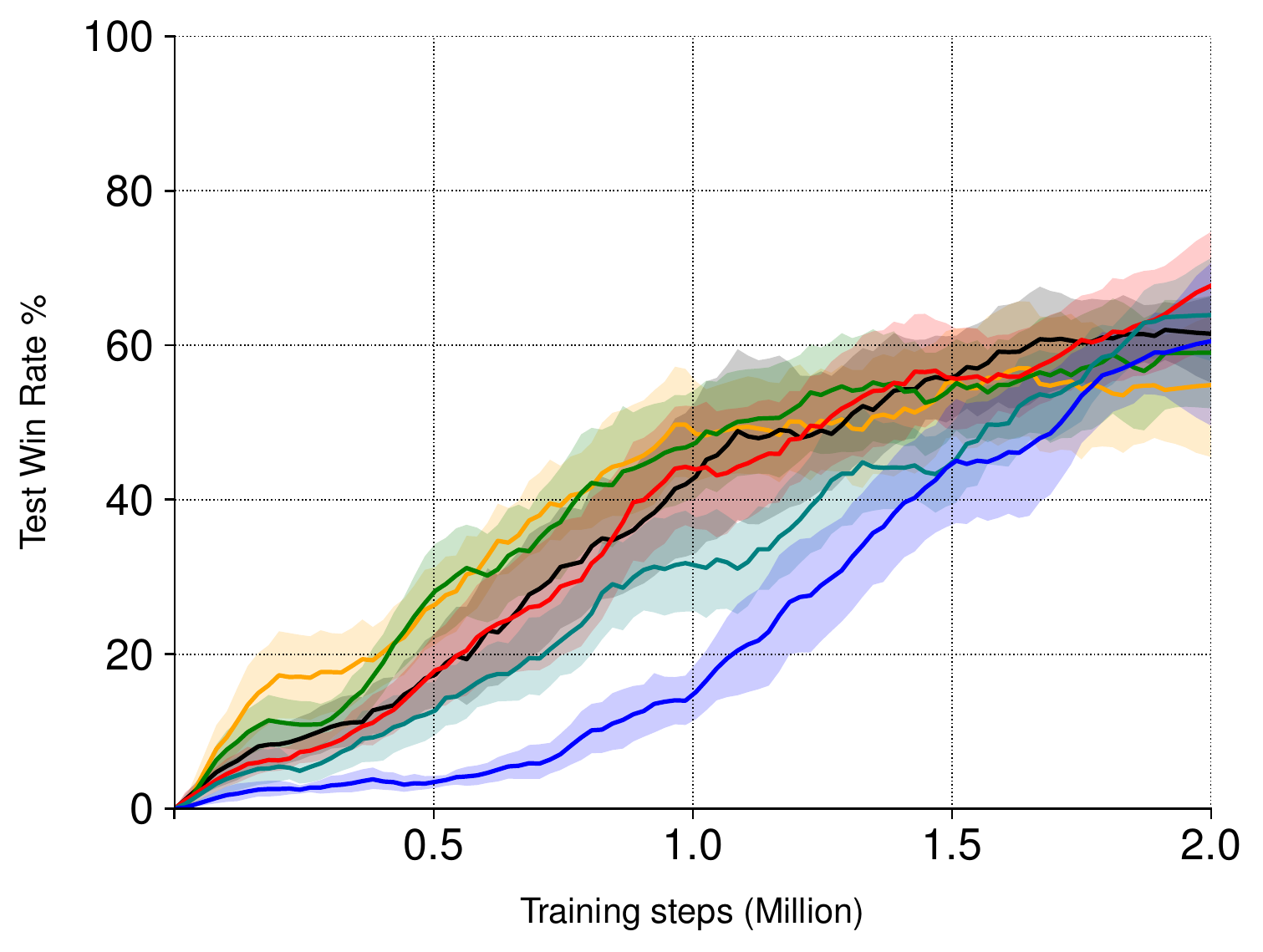}}
	\caption{Test win rates for our methods SM2 for different values of $\alpha$ and $\omega$, and comparison methods QMIX in six different scenarios.
		The mean and 95\% confidence interval are shown across 10 independent runs. The legend in (a) applies across all plots.
	}
	\label{fig:aoperformance}
\end{figure*}
\subsection*{A.5 Additional Experiments}
\subsubsection{Experimental Setup.}
For PLE game environments, the neural network was a multi-layer perceptron with hidden layer fixed to [64, 64].

The discount factor was 0.99.
The size of the replay buffer was 10000.
The weights of neural networks were optimized by RMSprop with gradient clip 5.
The batch size was 32.
The target network was updated every 200 frames.
$\epsilon$-greedy was applied as the exploration policy with $\epsilon$ decreasing linearly from 1.0 to 0.01 in 1, 000 steps.
After 1, 000 steps, $\epsilon$ was fixed to 0.01.

\begin{figure}[t!]  
	\centering
	\subfloat[Catcher-PLE]{\includegraphics[width=0.23\textwidth]{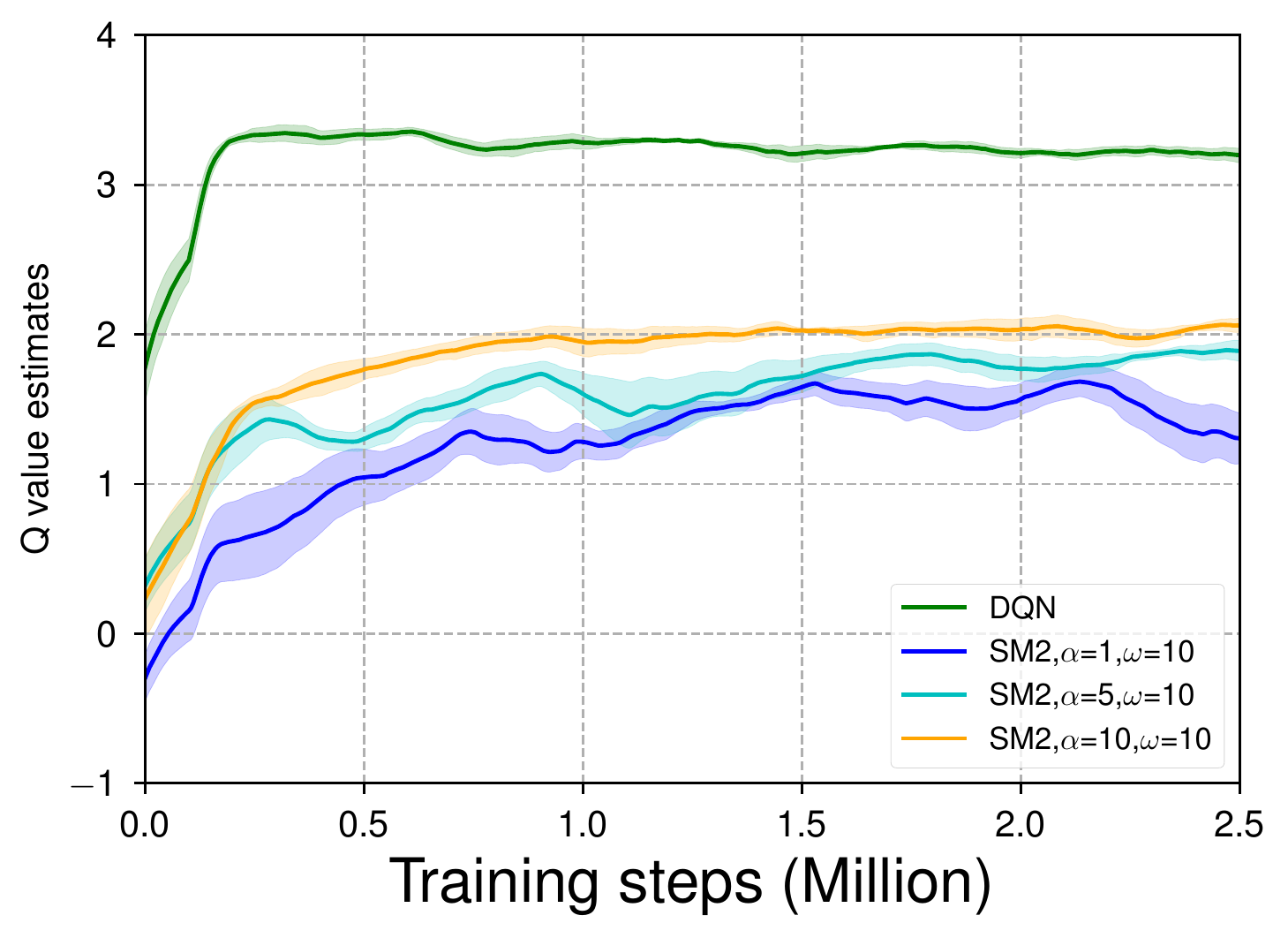}}\hfill
	\subfloat[Asterix-MinAtar]{\includegraphics[width=0.23\textwidth]{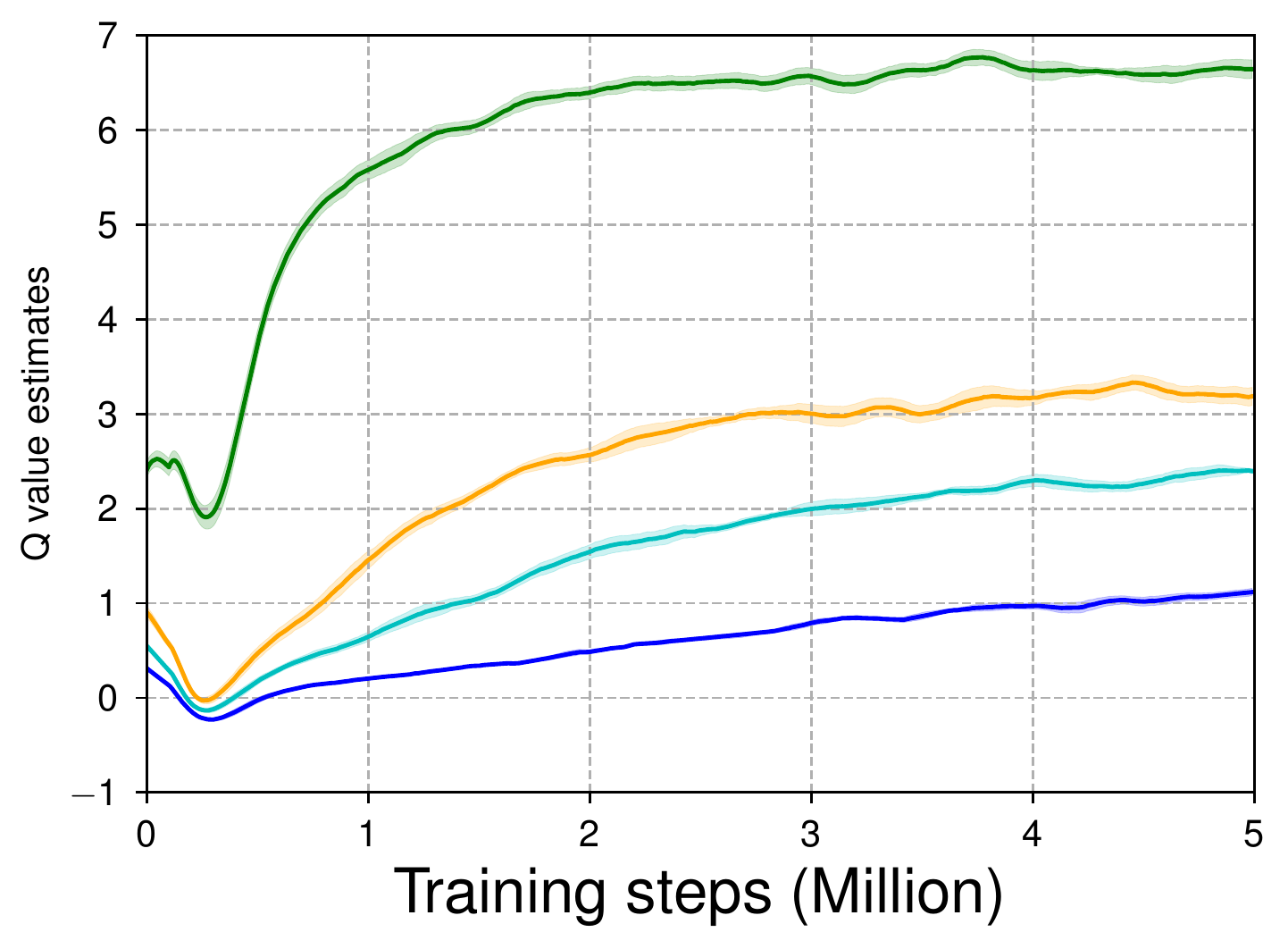}}
	\caption{The estimated Q values of SM2 under condition $ \alpha<\omega $. The mean and a standard deviation are shown across 5 independent runs.
	}
	\label{fig:single_a_leq_w}
\end{figure}
\begin{figure*}[t!]
	\centering
	\subfloat[3m]{\includegraphics[width=0.33\textwidth]{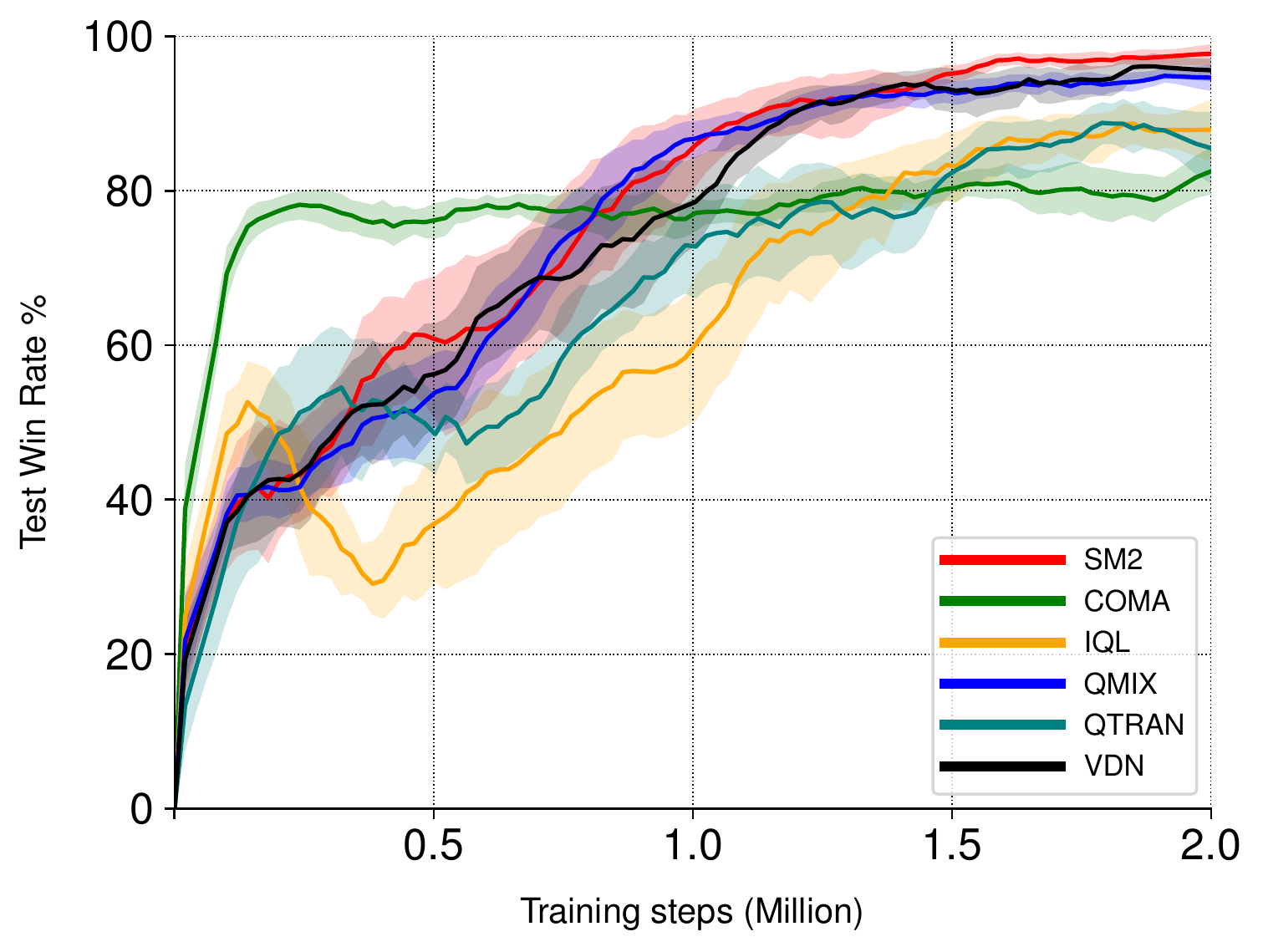}}
	\subfloat[8m]{\includegraphics[width=0.33\textwidth]{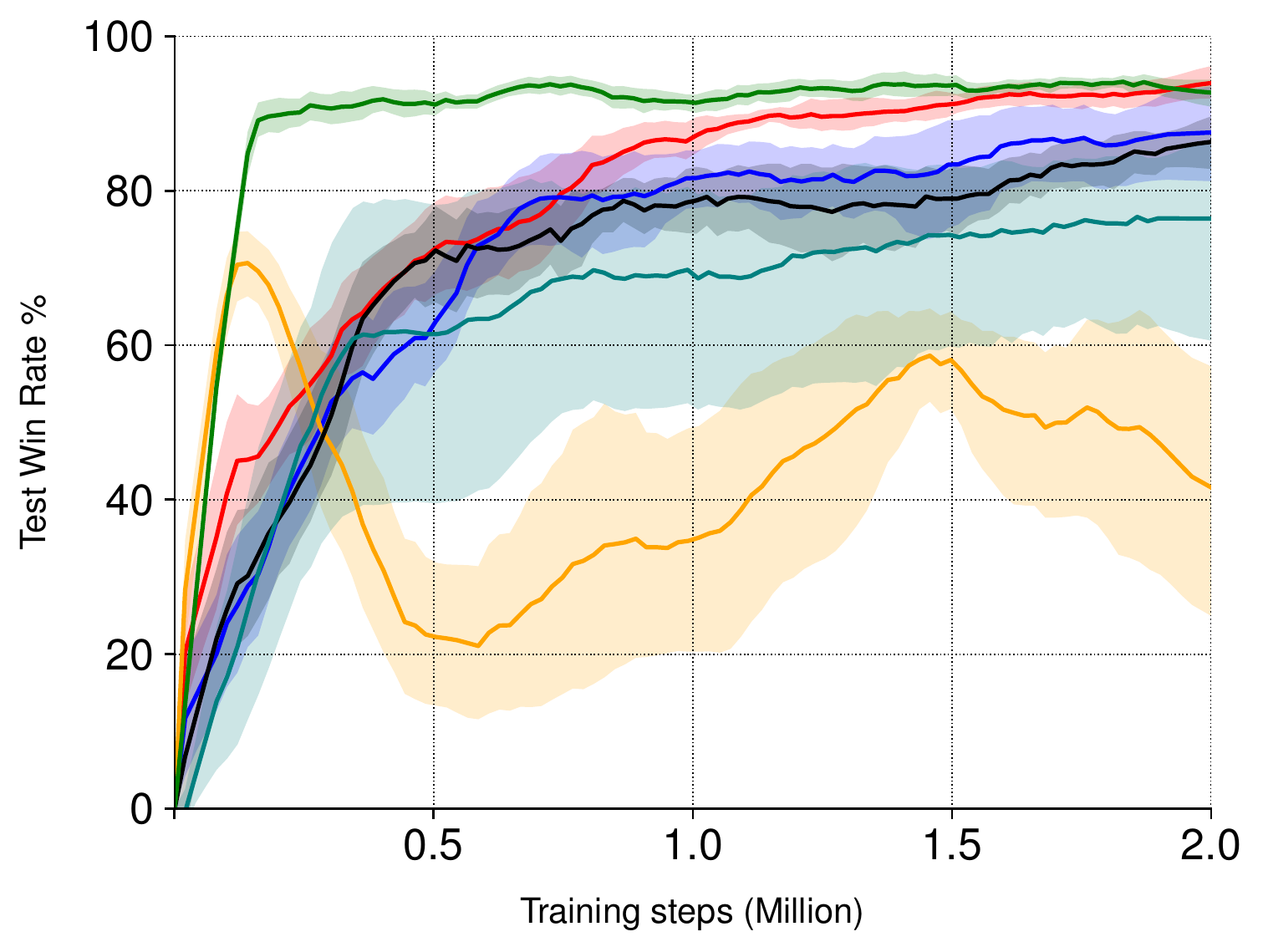}}
	\subfloat[3s5z]{\includegraphics[width=0.33\textwidth]{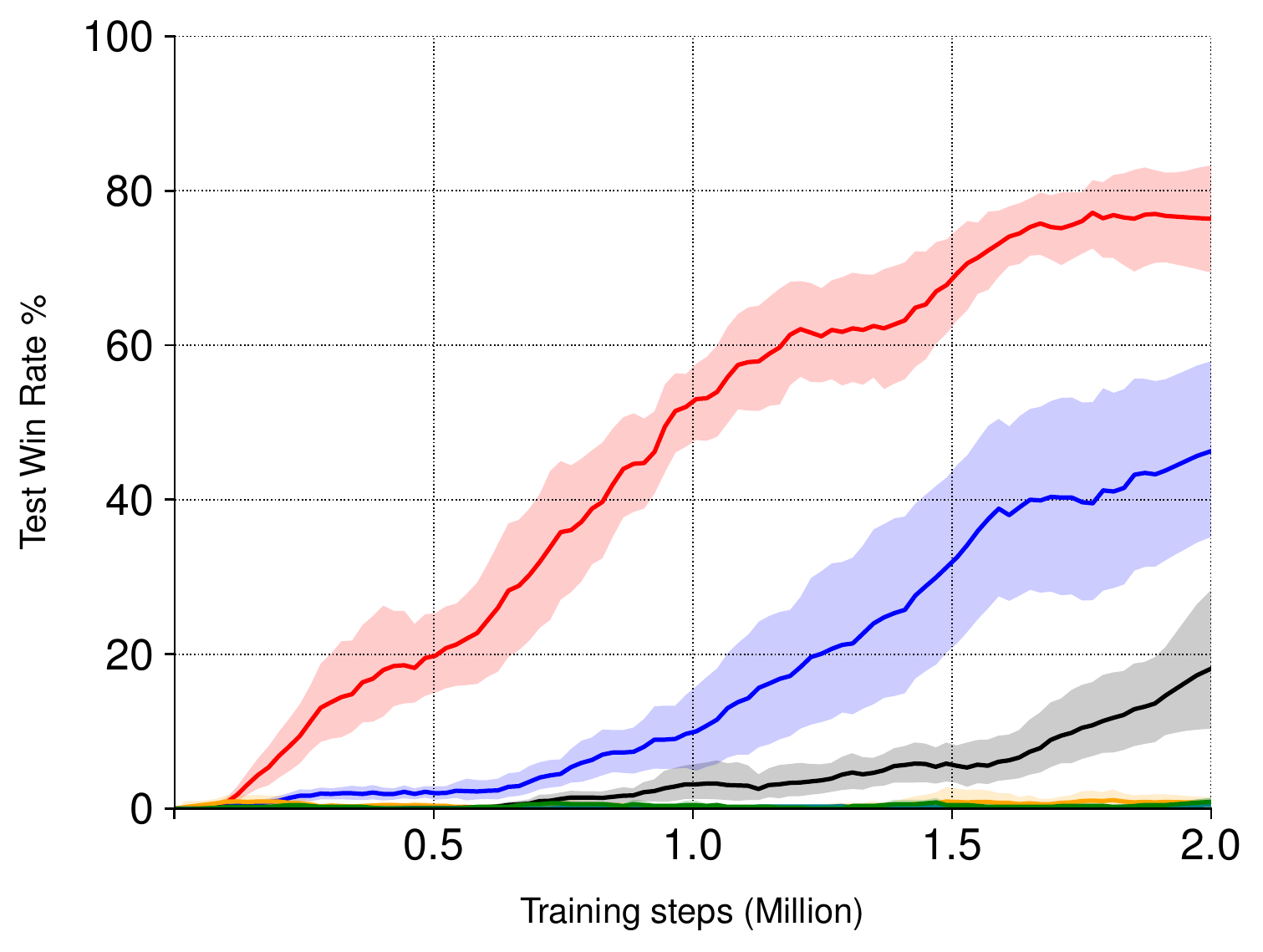}}\\ 
	\subfloat[2s\_vs\_1sc]{\includegraphics[width=0.33\textwidth]{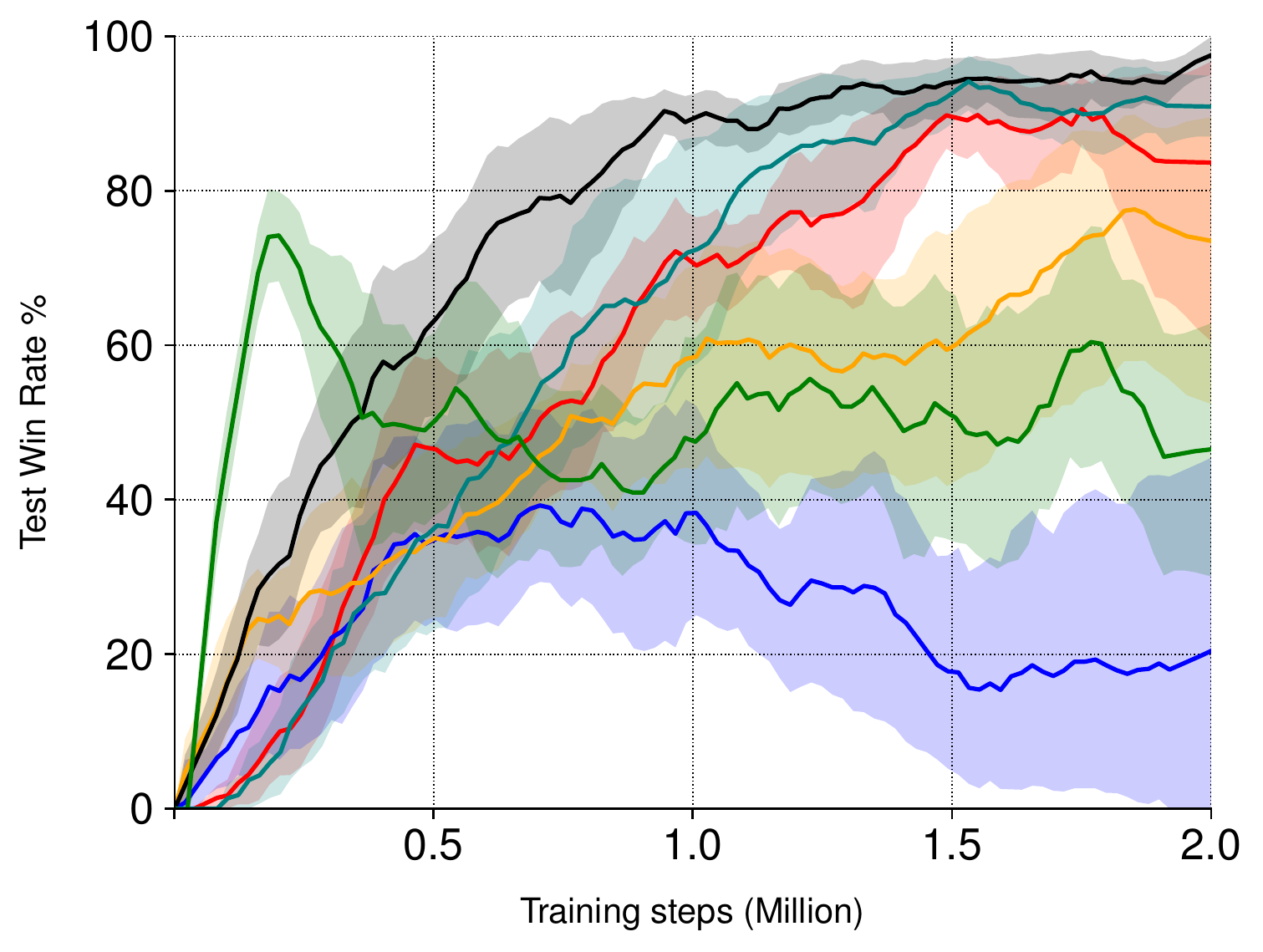}}
	\subfloat[3s6z]{\includegraphics[width=0.33\textwidth]{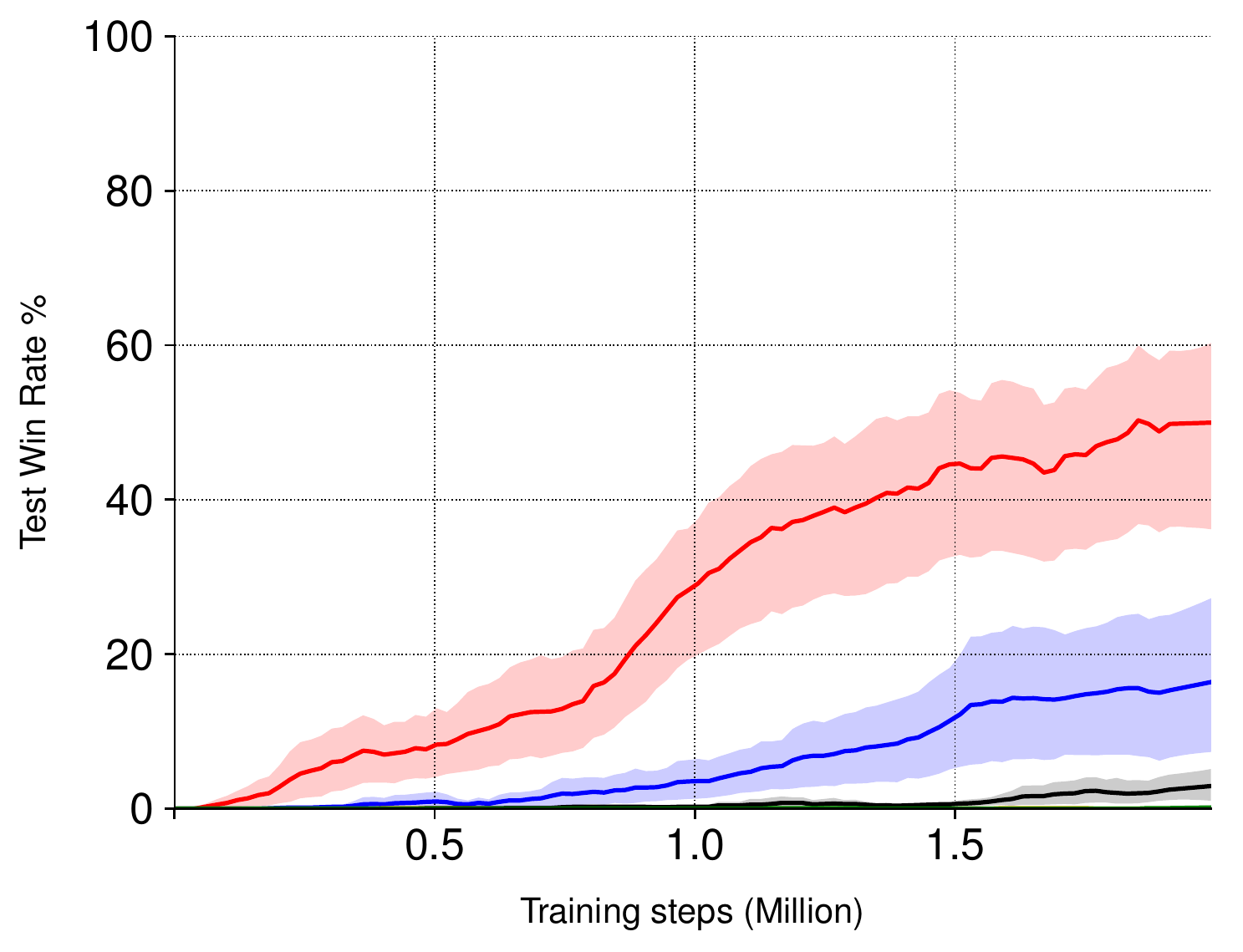}}
	\subfloat[1c3s5z]{\includegraphics[width=0.33\textwidth]{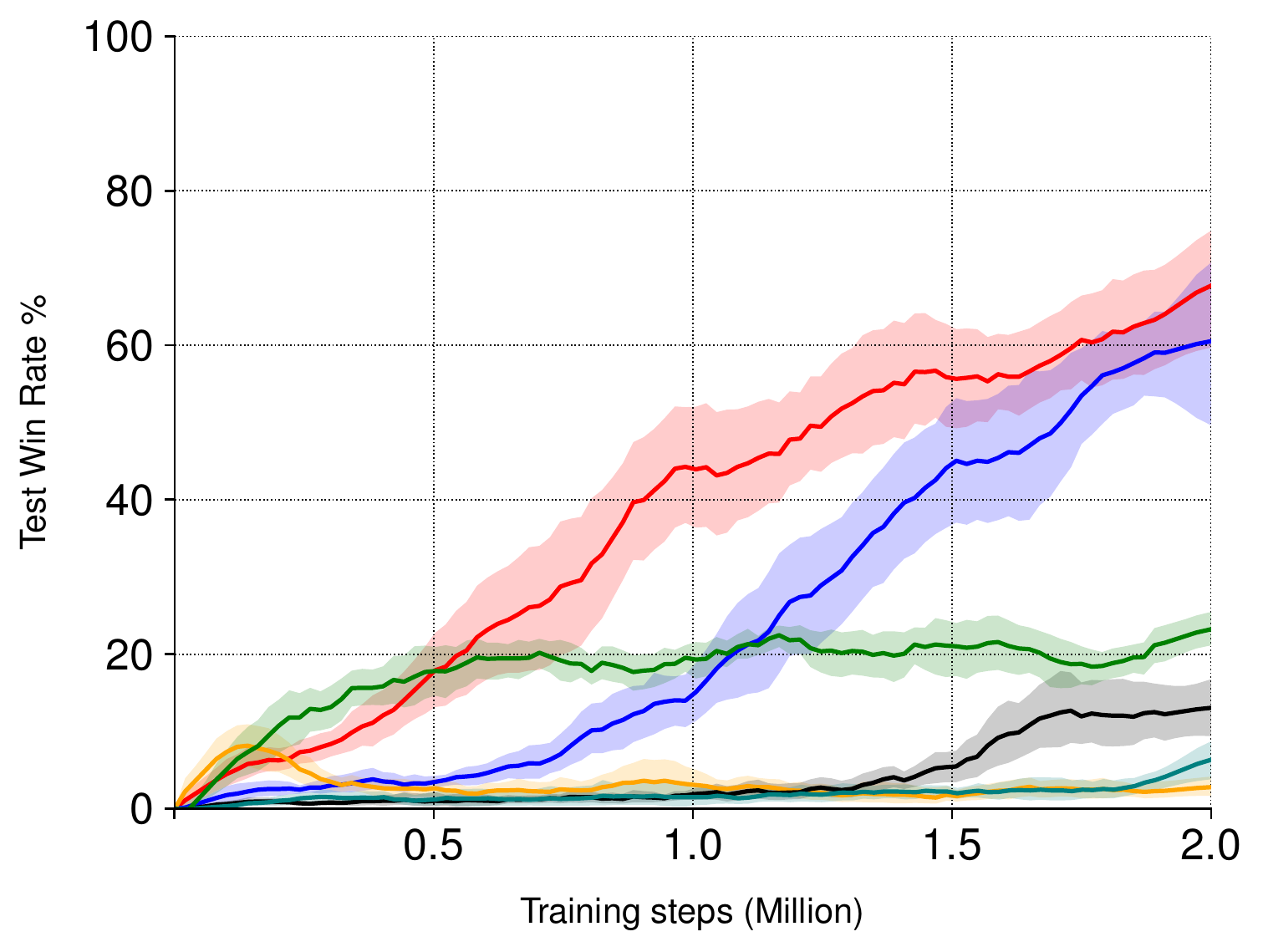}}\\
	\caption{Test win rates for our methods SM2, and comparison methods COMA, IQL, QMIX, QTRAN, and VDN in six different scenarios.
		The mean and 95\% confidence interval are shown across 10 independent runs.
	}
	\label{fig:performance}
\end{figure*}
\begin{figure}[t!]
	\centering
	\subfloat[Catcher-PLE]{\includegraphics[width=0.22\textwidth]{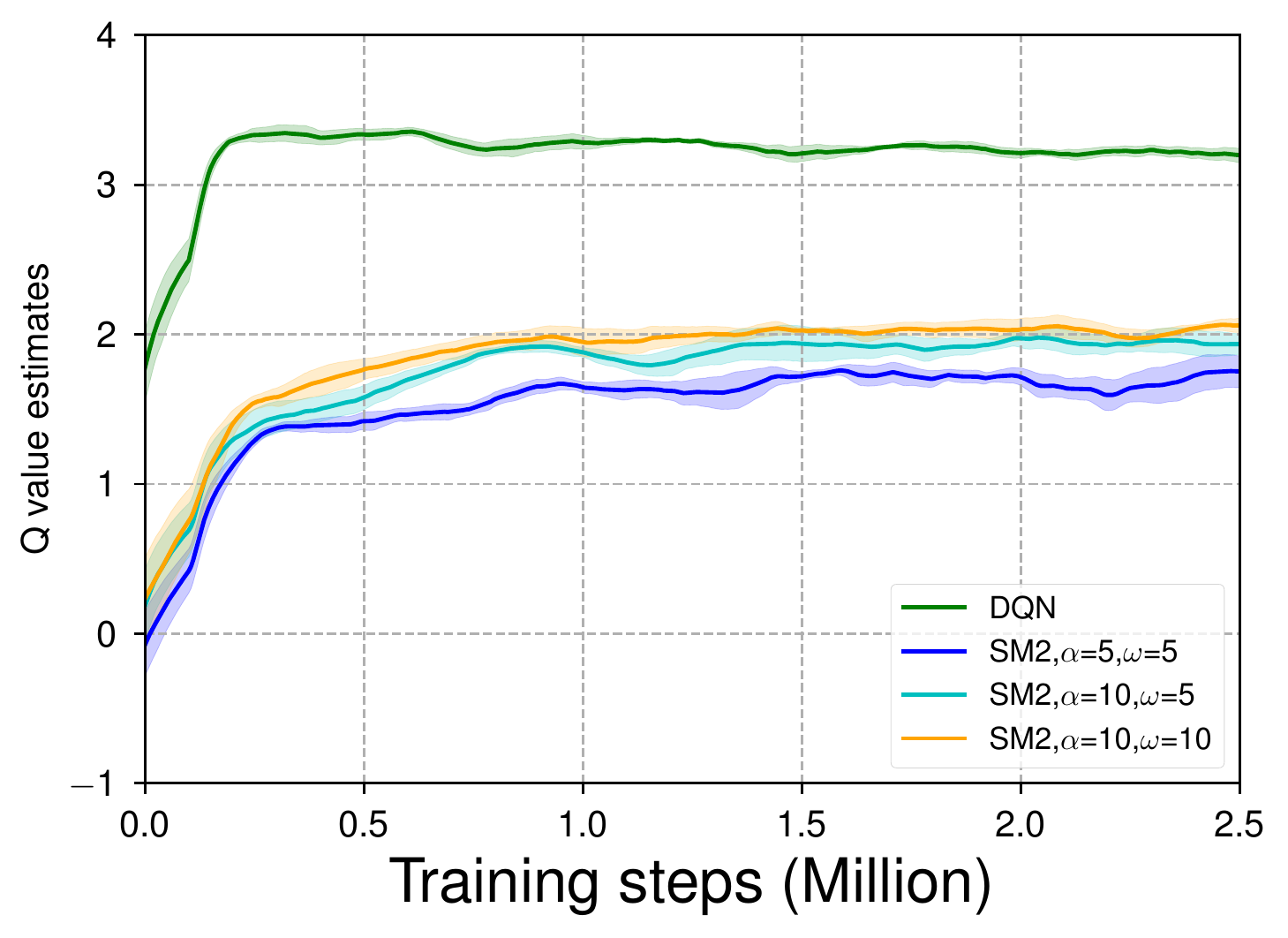}}\hfill
	\subfloat[Asterix-MinAtar]{\includegraphics[width=0.22\textwidth]{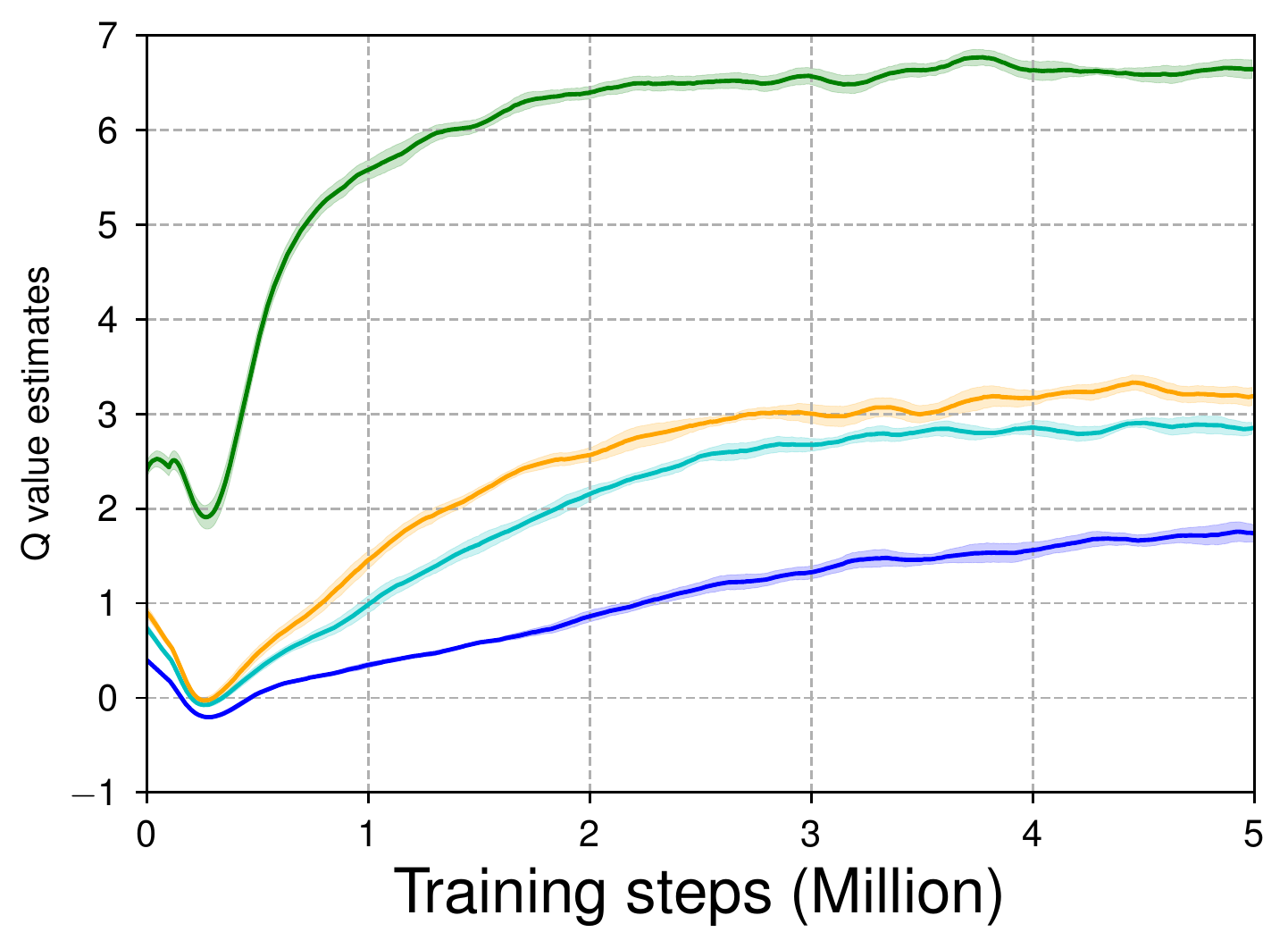}}
	\caption{The estimated Q values of SM2 under condition $ \alpha\geq\omega $. The mean and a standard deviation are shown across 5 independent runs.}
	\label{fig:single_w_leq_a}
\end{figure}

The StarCraft Multi-agent Challenge (SMAC) environment (Samvelyan et al. 2019) has become a benchmark for evaluating state-of-the-art MARL approaches such as VDN (Sunehag et al. 2018), COMA (Foerster et al. 2018), QMIX (Rashid et al. 2018) and QTRAN (Son et al. 2019).
The StarCraft is a rich environment with stochastic dynamics that cannot be easily emulated.
It focuses solely on micromanagement in SMAC.
SMAC is based on the popular real-time strategy (RTS) games StarCraft II.
SMAC provides several challenging micro scenarios that aim to evaluate different aspects of cooperative behaviors of a group of agents.
Each unit by an independent agent that conditions only on local observations restricted to a limited field of view centred on that unit.
The discrete set of actions that agents are allowed to take consists of move[direction], attack[enemy\_id], stop and no-op.
Agents can only move in four directions: north, south, east, or west.
The maximum number of actions an agent can take ranges between 7 and 70, depending on the scenario.

At each time step, the agents receive a joint reward equal to the total damage dealt on the enemy units. In addition, agents receive a bonus of 10 points after killing each opponent, and 200 points after killing all opponents.
These rewards are all normalized to ensure the maximum cumulative reward achievable in an episode is 20.
We use the default setting for the reward.
Refer to (Samvelyan et al. 2019) for more details.

We use the same network architecture with QMIX.
The architecture of all agent networks is a deep recurrent network (DRQN) with a recurrent layer comprised of a GRU with a 64-dimensional hidden state, with a fully-connected layer before and after.
DRQN learns policies that it is both robust enough to handle to missing game screens, and scalable enough to improve performance as more data becomes available (Hausknecht and Stone 2015).
Exploration is performed during training using independent $\epsilon$-greedy action selection, where each agent $i$ performs $\epsilon$-greedy action selection over its own $Q_{i}$ (Rashid et al. 2018).
Throughout the training, $\epsilon$ is linear from 1.0 to 0.05 over 50k time steps.
And $\gamma$ = 0.99 for all experiments.
The replay buffer contains the most recent 5000 episodes.
We train all network by sampling batches of 32 episodes uniformly from the replay buffer.
The target networks are updated after every 200 training episodes.
We use the default evaluation metric, which is a test win rate.
The test win rate is evaluated in the following procedure: we run 24 independent episodes with each agent performing greedy decentralized action selection after every 20000 timesteps.
The win rate is the percentage of episodes where the agents defeat all enemy units within the time limit.

\end{document}